\documentclass{article} 
\usepackage{iclr2027_conference,times}

\usepackage{amsmath,amsfonts,bm}

\def\eqref#1{equation~\ref{#1}}

\def\1{\bm{1}}

\DeclareMathAlphabet{\mathsfit}{\encodingdefault}{\sfdefault}{m}{sl}
\SetMathAlphabet{\mathsfit}{bold}{\encodingdefault}{\sfdefault}{bx}{n}

\usepackage{hyperref}
\usepackage{url}
\usepackage{booktabs}
\usepackage{float}
\usepackage{tabularx}
\usepackage{amsmath}
\usepackage{amssymb}
\usepackage{threeparttable}
\usepackage{tablefootnote}
\usepackage{graphicx}
\usepackage{wrapfig}
\usepackage{tikz}

\usepackage{amsthm}
\newtheorem{assumption}{Assumption}
\newtheorem{theorem}{Theorem}

\usetikzlibrary{
  positioning,
  fit,
  calc,
  arrows.meta,
  backgrounds,
  matrix
}

\title{Let the Carrier Carry the Attack:
Preserving the Subject in Adversarial Image Generation}

\author{%
  Linfeng Jiang$^{1}$ \quad Steven McDonagh$^{2}$ \quad Yuhang Chen$^{3}$ \quad Xingyu Zhao$^{3,4}$ \\ \textbf{Siddartha Khastgir}$^{3}$ \quad \textbf{Andi Zhang}$^{3}$\thanks{Corresponding Author.} \\
  $^1$University of Maryland, College Park \quad $^2$University of Edinburgh \\ $^3$WMG, University of Warwick \quad
  $^4$Wuhan University\\
  \texttt{andi.zhang@warwick.ac.uk} \\
}

\iclrfinalcopy 
\begin{document}

\maketitle

\begin{abstract}

Strong unrestricted adversarial attacks can distort the primary object of an image, hereafter referred to as the \emph{subject}. To preserve subject integrity without compromising attack magnitude, we introduce the \textbf{carrier}: a secondary visual element that provides an auxiliary region to facilitate the attack under global classifier guidance. 
%
We demonstrate three key findings: 1.~A carrier mitigates subject distortion by absorbing a larger share of globally normalized attack updates. 2.~A carrier improves cross-model transferability, governed by the strength of target-related features that balance semantic separation and transfer performance. 3.~Successful targeted attacks retain the personalized subject as the primary content perceived by humans while successfully misleading the classifier. 
Our results demonstrate that a visually secondary carrier offers an auxiliary spatial pathway for adversarial changes, enabling strong and transferable attacks while improving subject preservation. Code is available at \url{https://github.com/DavidJlf/carrier-attack}.

\end{abstract}

\section{Introduction}
\label{sec:introduction}

Adversarial attacks aim to mislead classifiers while preserving human recognition of the image content~\citep{goodfellow2015,szegedy2014}. Traditional attacks constrain perturbations within a small budget, whereas unrestricted attacks allow larger variations in pose, viewpoint, composition, and background~\citep{conceptbased,brown2018,song2018}. By operating on an entire object category rather than a single image, unrestricted attacks can generate diverse adversarial examples with higher attack efficiency while preserving the underlying semantics. 
Yet, as these attacks grow stronger to ensure misclassification, they tend to sacrifice the fidelity of the original subject. 
Moreover, targeted attacks impose an additional requirement: the classifier must predict a specified target class rather than merely deviate from the correct label. This motivates our central question: 
Can we achieve strong targeted attacks without sacrificing the integrity of the primary subject?

To address this challenge, we introduce a \textbf{carrier}: a visually secondary object, constructed separate from the primary subject, that provides an additional region for attack-related changes. We define an initial ``No-Carrier'' baseline which implements concept-based attack by~\citet{conceptbased}; the approach already achieves strong WhiteBox attack performance as shown in Table~\ref{tab:main_attack_analysis}. Our goal is to improve cross-model transferability and maintain high attack strength while preserving the subject. By adding a secondary region for the attack, the carrier reduces the relative update applied to the primary subject under normalization while its target-relevant features simultaneously boost transferability across different models.

To ensure the carrier remains visually secondary, we construct it separate from the primary subject by dividing the source image into subject and background layers. The carrier can then be realized in the background region through direct scene compositing or mask-guided inpainting conditioned on the source image.


Constructing a background with the subject and carrier alone only produces an altered image rather than an attack. To demonstrate that the carrier is not tied to a single construction or attack-integration strategy, we explore three distinct attack realizations: Composite Reconstruction Attack (CRA), Clean Inpainting Reconstruction Attack (CIRA), and Joint Inpainting Attack (JIA).
CRA starts from a clean composite,
whereas CIRA starts from a completed clean inpainting. Inspired by inversion-based
generative attacks~\citep{aca,diffattack}, both routes invert the constructed image
to an intermediate FLUX state and apply global normalized guidance during selected
return steps, with LoRA conditioning supporting subject preservation. JIA instead couples carrier construction with classifier guidance within the same
conditional inpainting trajectory, where each guided update is followed by native mask blending. All three routes use global normalized guidance,
with the visually secondary carrier providing an additional semantic region outside
the primary subject for expressing attack-related changes.

To evaluate how semantic alignment with the target class influences attack
effectiveness and transferability, we compare three regimes: \textbf{Non-Target Carrier}, \textbf{Hybrid Carrier}, and \textbf{Target Carrier}. These conditions offer flexibility in trading semantic separation from the target against transferability. We also implement the original concept-based attack on FLUX as our \textbf{{No-Carrier}} baseline.

Our contributions can be summarized as follows:
\begin{itemize}

\item We identify a conditional update-allocation mechanism under global RMS normalization: for matched subject and remaining-region gradients, a carrier component reduces the normalized update applied to the primary-subject region. We further demonstrate improved subject preservation under matched strong attacks
(Sections~\ref{sec:carrier_preservation_principle}
and~\ref{sec:preservation}).

\item We empirically show that a carrier improves cross-model transferability, with stronger target-related carrier characteristics yielding progressively larger gains.
We further derive a conditional bound showing how a larger ideal target margin produces a non-decreasing guaranteed lower bound on targeted transferability
(Sections~\ref{sec:carrier_transferability_principle}
and~\ref{transferability}).

\item We develop a carrier-guided spatial attack framework across three realizations and show that a visually secondary carrier can support successful targeted attacks while retaining the primary subject as the primary image content
(Sections~\ref{sec:method}
and~\ref{sec:human_test_evluation}).

\end{itemize}

\begin{figure}[t]
  \centering
  \newlength{\carrierfigcell}
  \setlength{\carrierfigcell}{0.116\linewidth}
  \begin{tikzpicture}
    \tikzset{
      carrierimage/.style={draw=black!82,line width=0.35pt,inner sep=0pt,outer sep=0pt},
      sourcecase/.style={text width=\carrierfigcell,align=center,font=\scriptsize\bfseries,
        text=black!88,minimum height=2.8mm,inner sep=0pt,outer sep=0pt},
      targetcase/.style={text width=\carrierfigcell,align=center,font=\tiny\itshape,
        text=black!55,minimum height=2.5mm,inner sep=0pt,outer sep=0pt},
      carrierrow/.style={align=right,font=\scriptsize\bfseries,text=black!76,inner sep=0pt,outer sep=0pt},
      grouptitle/.style={font=\scriptsize\bfseries,inner xsep=1.1mm,inner ysep=0.25mm},
      flowarrow/.style={-{Latex[length=1.25mm,width=0.9mm]},line width=0.35pt,draw=black!45},
      imagematrix/.style={matrix of nodes,nodes={anchor=center},column sep=0.85mm,row sep=3.4mm,
        row 1/.style={nodes={sourcecase}},
        row 2/.style={nodes={targetcase}},
        row 3/.style={nodes={carrierimage}},
        row 4/.style={nodes={carrierimage}},
        row 5/.style={nodes={carrierimage}},
        row 6/.style={nodes={carrierimage}}}
    }
    \matrix (m) [imagematrix] {
      {Bear plushie} & {Cat} & {Dog 5} & {Dog 2} \\[-2.3mm]
      {castle} & {hare} & {cellphone} & {ambulance} \\[1.0mm]
      {\includegraphics[width=\carrierfigcell,height=\carrierfigcell]{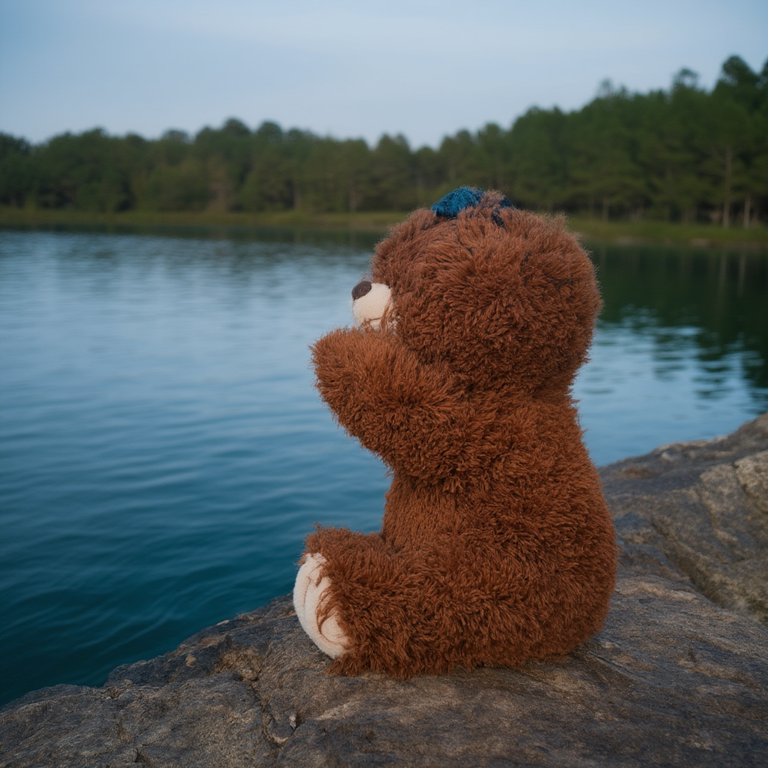}}
      & {\includegraphics[width=\carrierfigcell,height=\carrierfigcell]{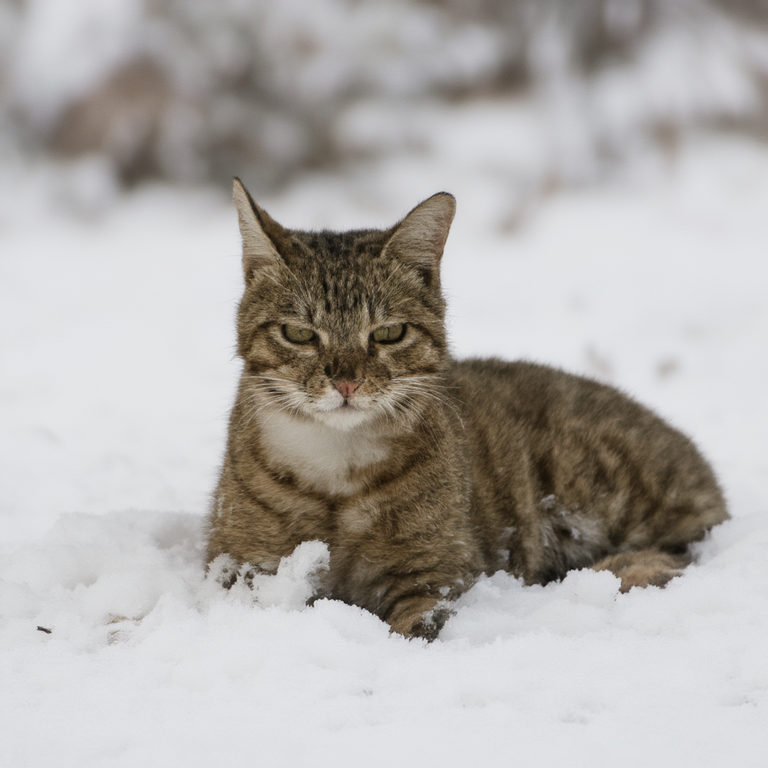}}
      & {\includegraphics[width=\carrierfigcell,height=\carrierfigcell]{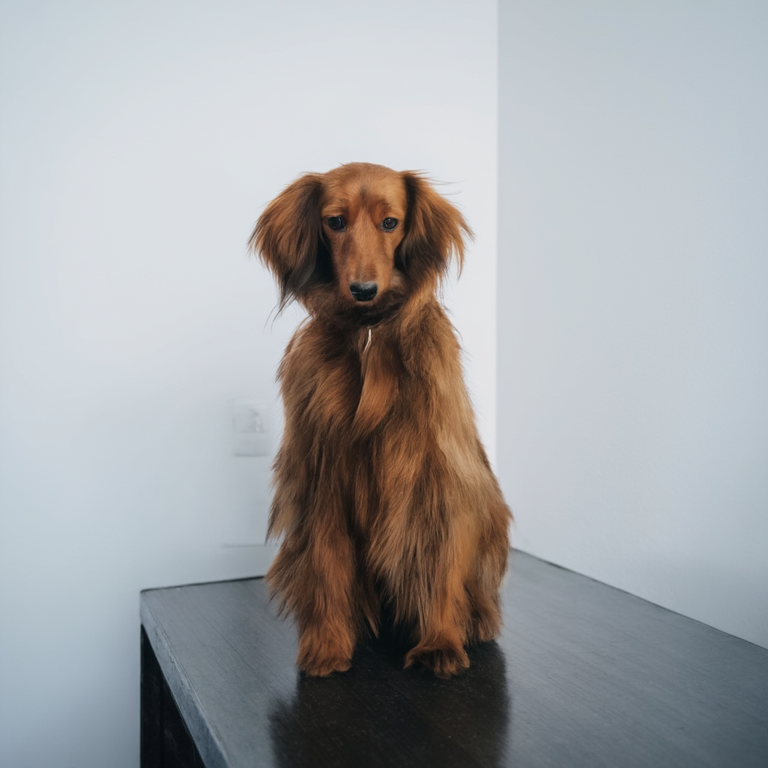}}
      & {\includegraphics[width=\carrierfigcell,height=\carrierfigcell]{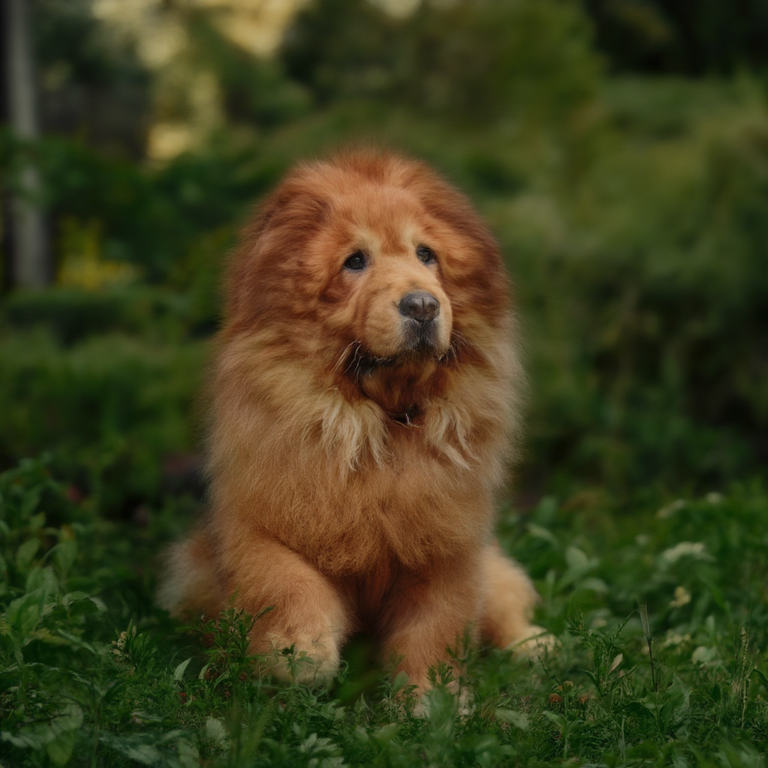}} \\
      {\includegraphics[width=\carrierfigcell,height=\carrierfigcell]{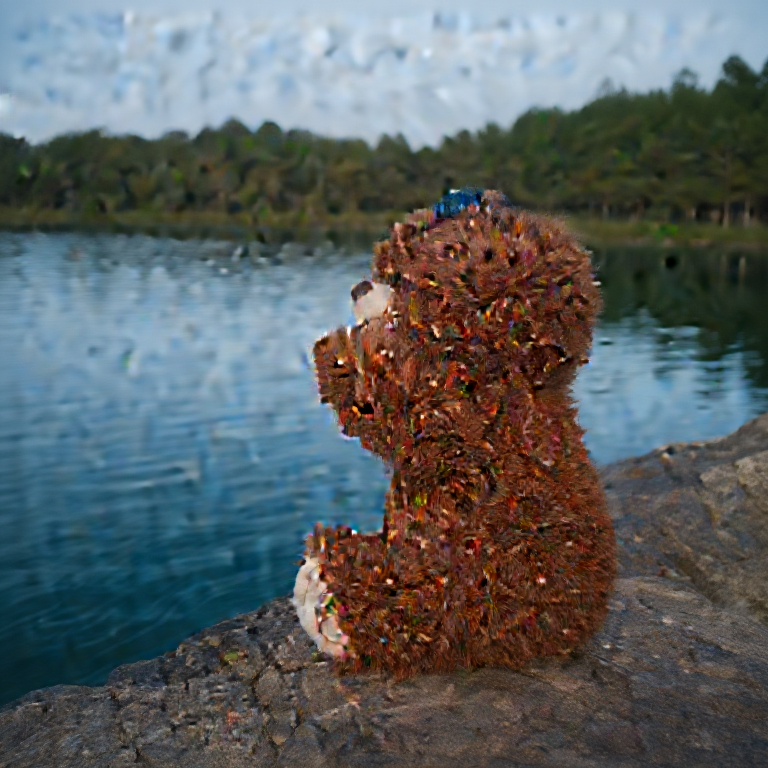}}
      & {\includegraphics[width=\carrierfigcell,height=\carrierfigcell]{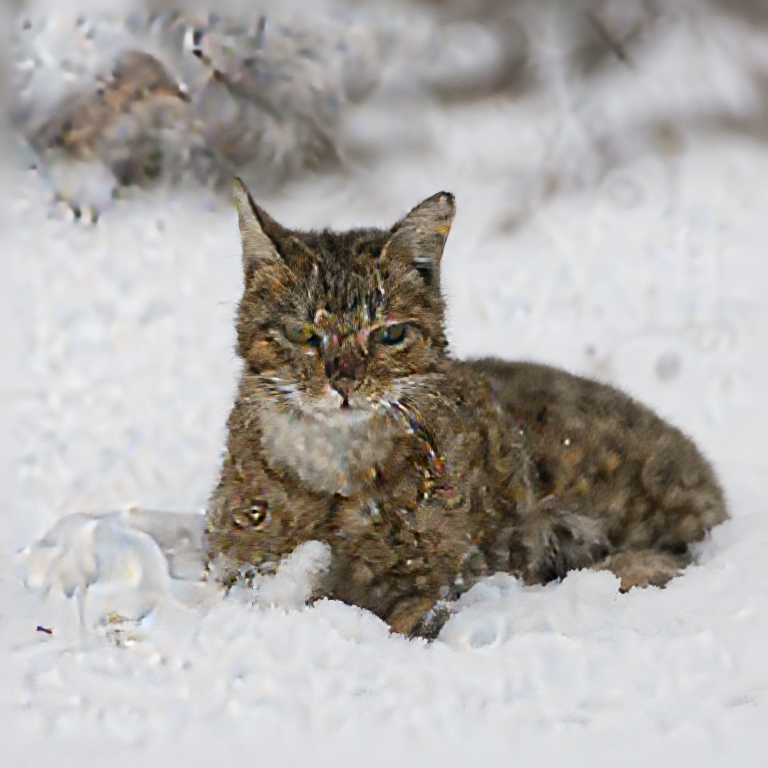}}
      & {\includegraphics[width=\carrierfigcell,height=\carrierfigcell]{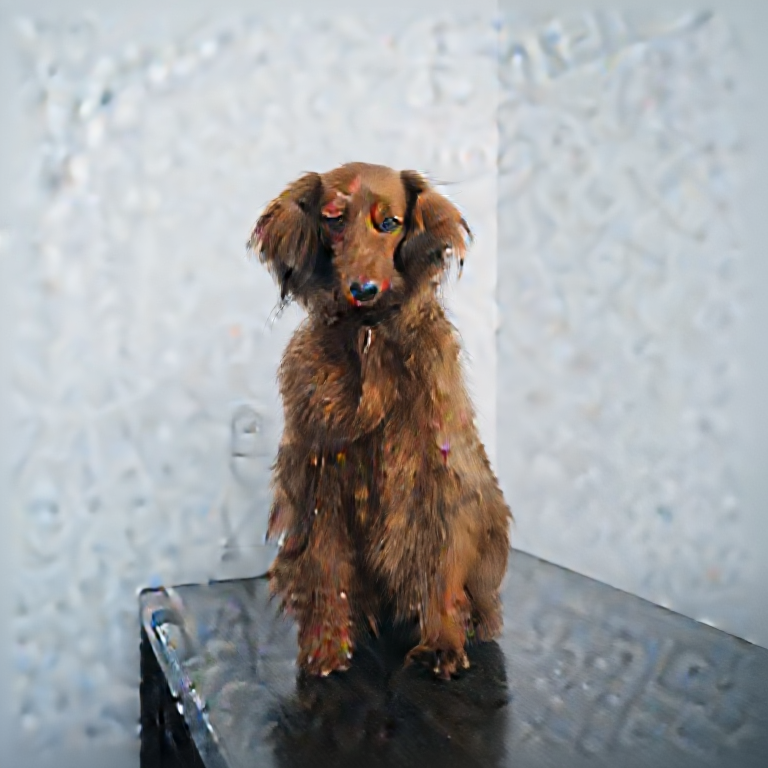}}
      & {\includegraphics[width=\carrierfigcell,height=\carrierfigcell]{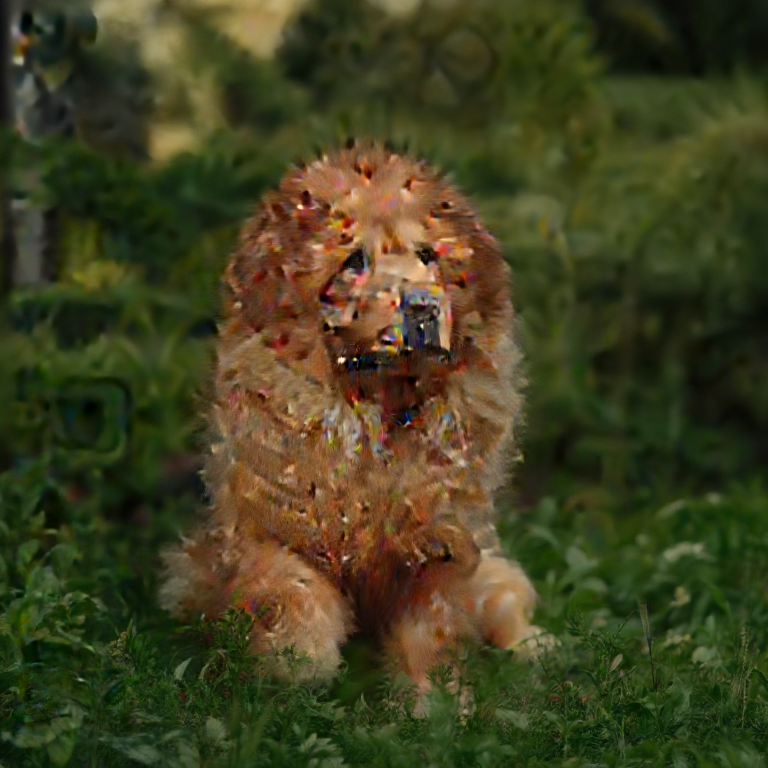}} \\[5.2mm]
      {\includegraphics[width=\carrierfigcell,height=\carrierfigcell]{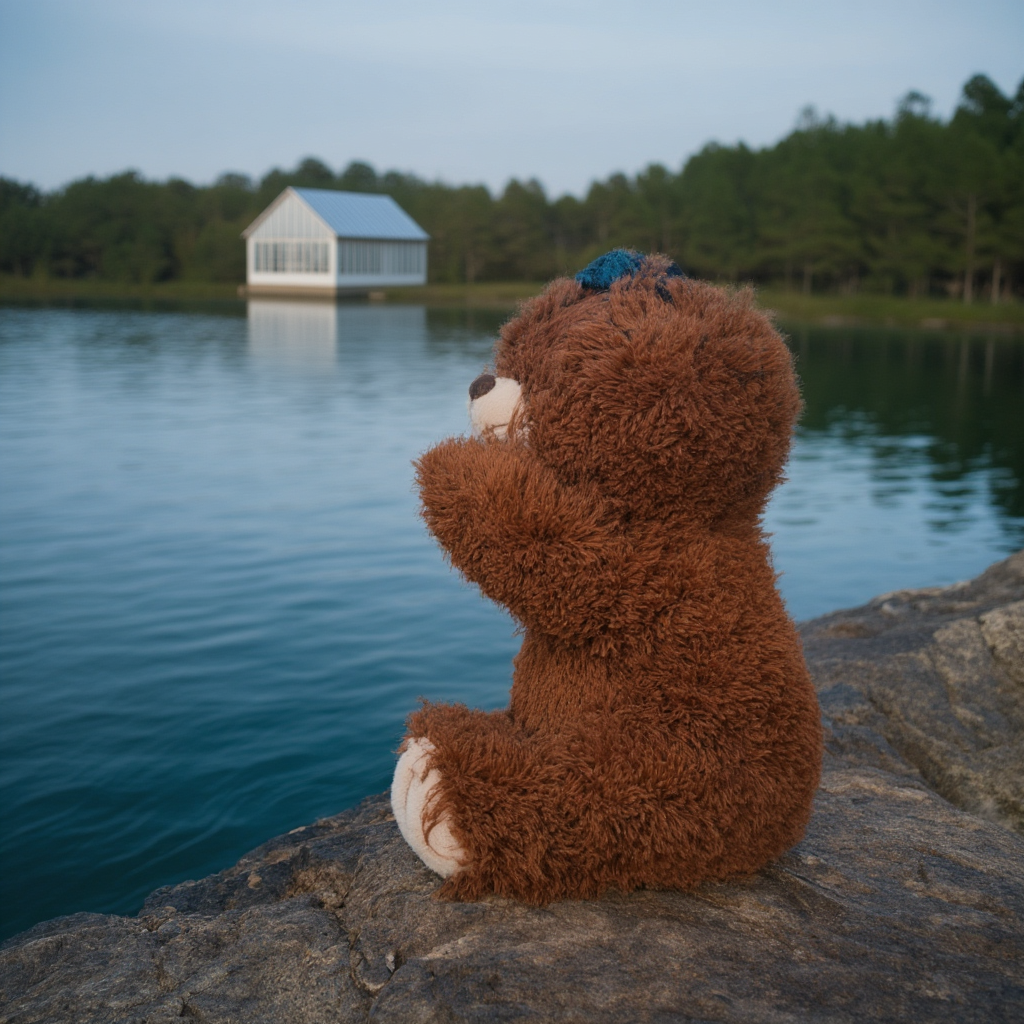}}
      & {\includegraphics[width=\carrierfigcell,height=\carrierfigcell]{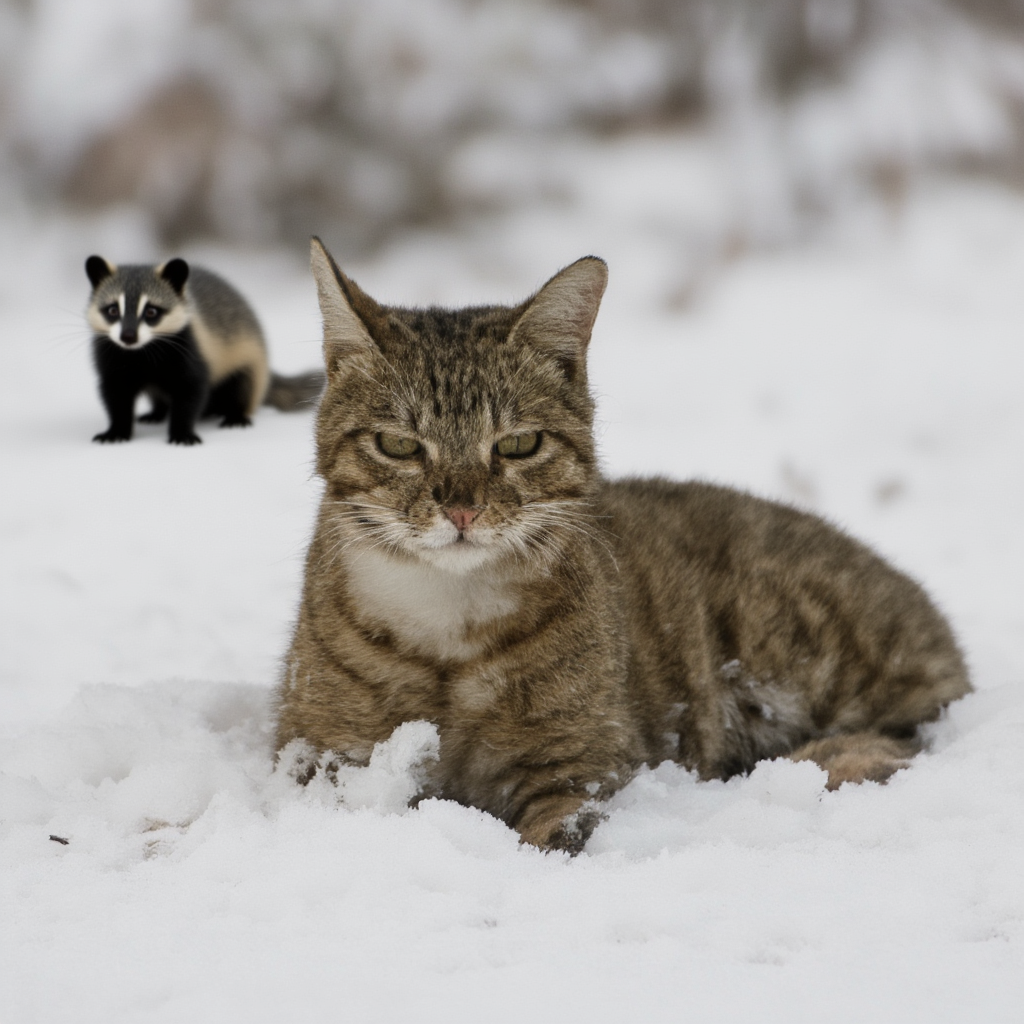}}
      & {\includegraphics[width=\carrierfigcell,height=\carrierfigcell]{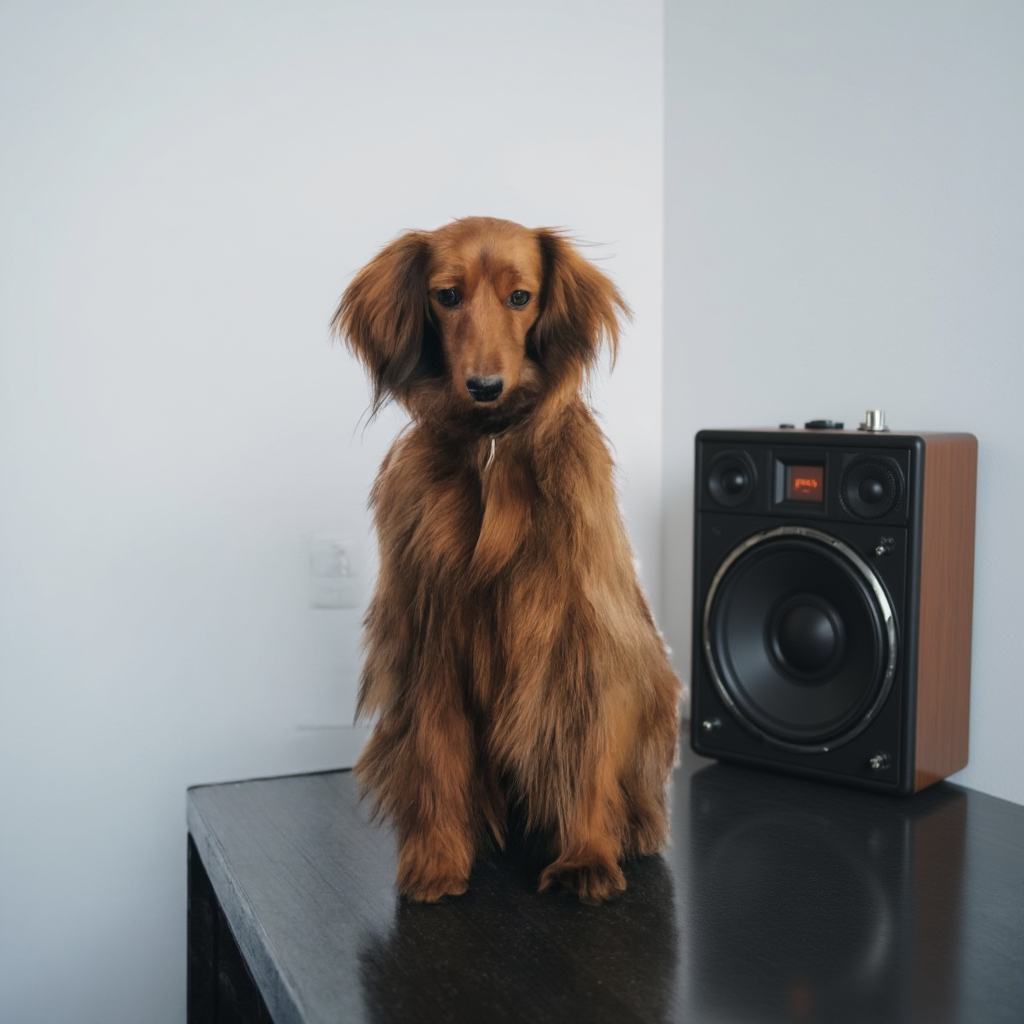}}
      & {\includegraphics[width=\carrierfigcell,height=\carrierfigcell]{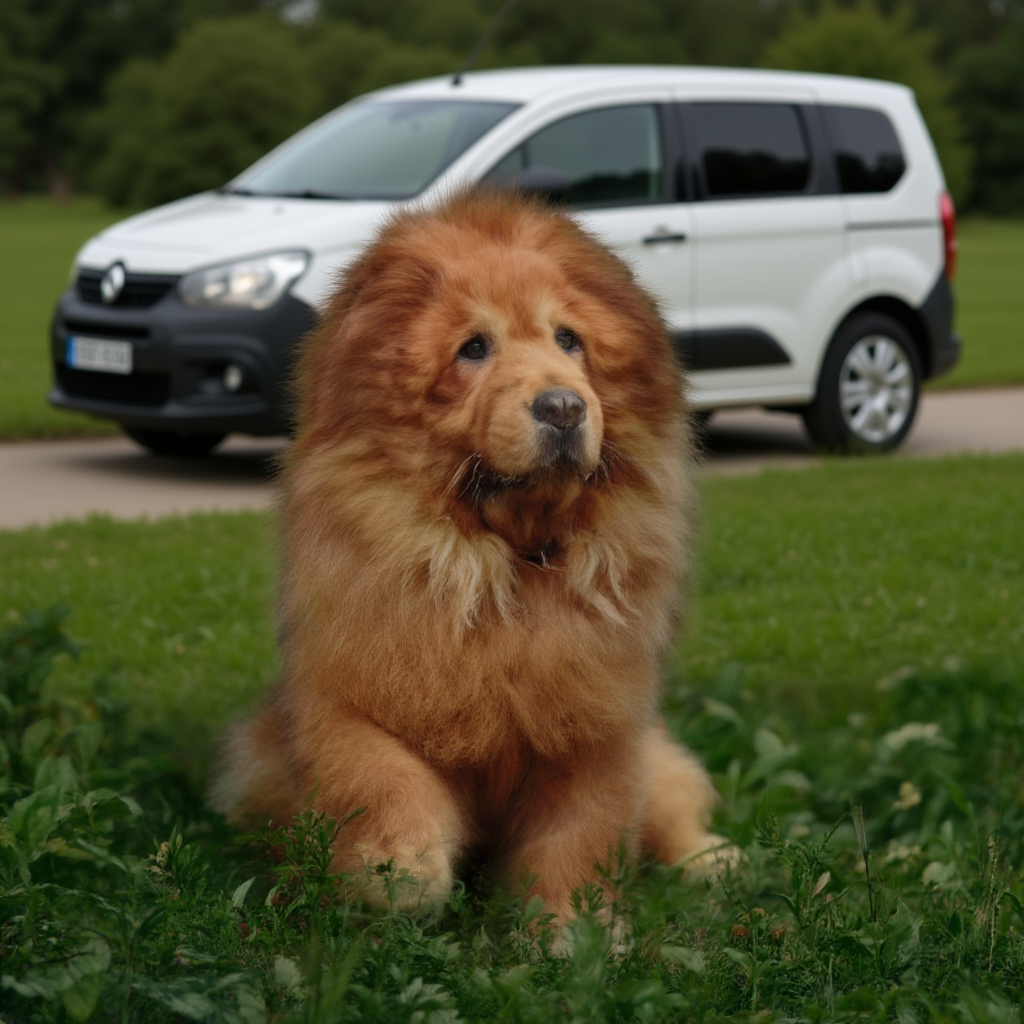}} \\
      {\includegraphics[width=\carrierfigcell,height=\carrierfigcell]{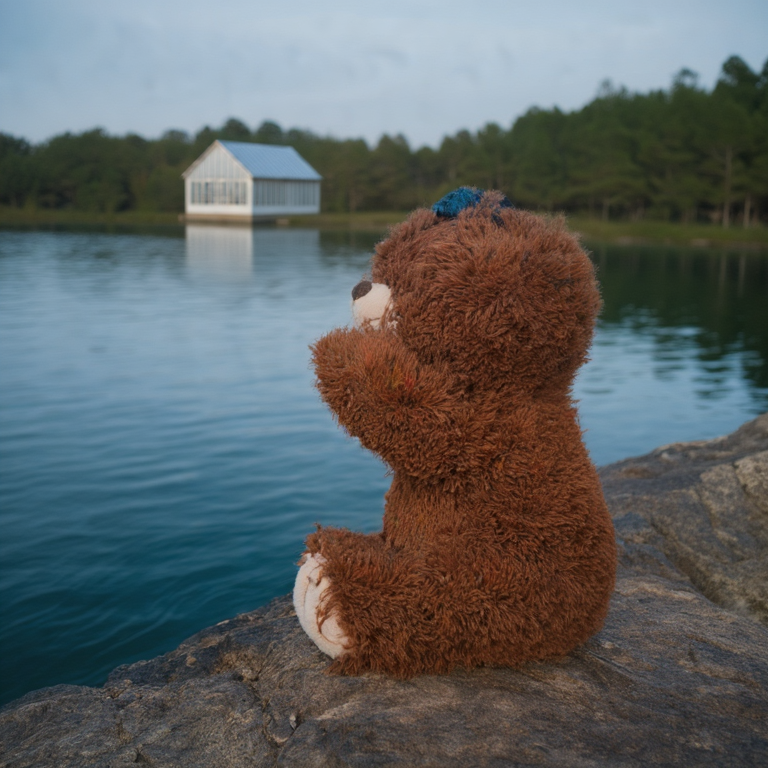}}
      & {\includegraphics[width=\carrierfigcell,height=\carrierfigcell]{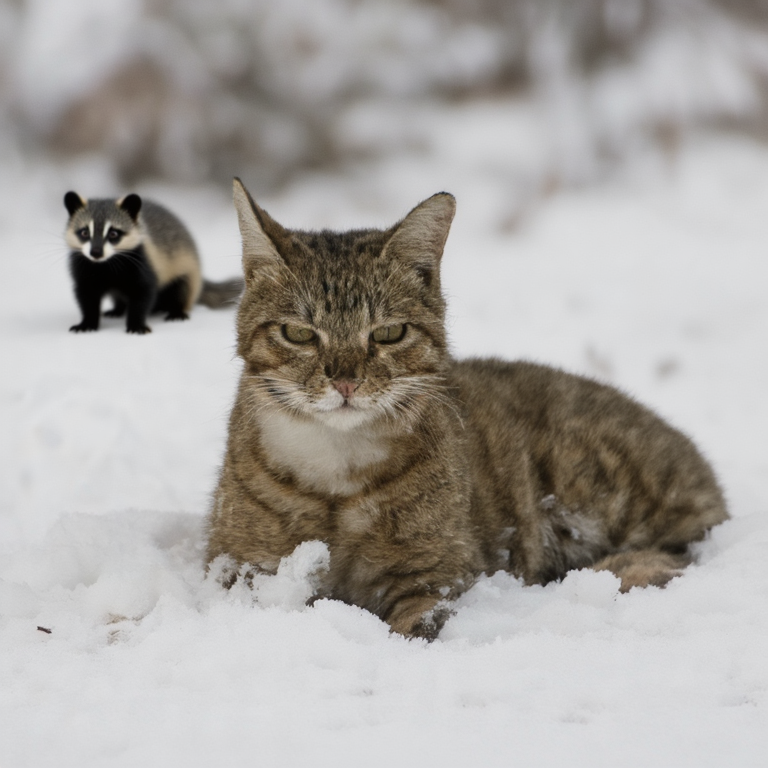}}
      & {\includegraphics[width=\carrierfigcell,height=\carrierfigcell]{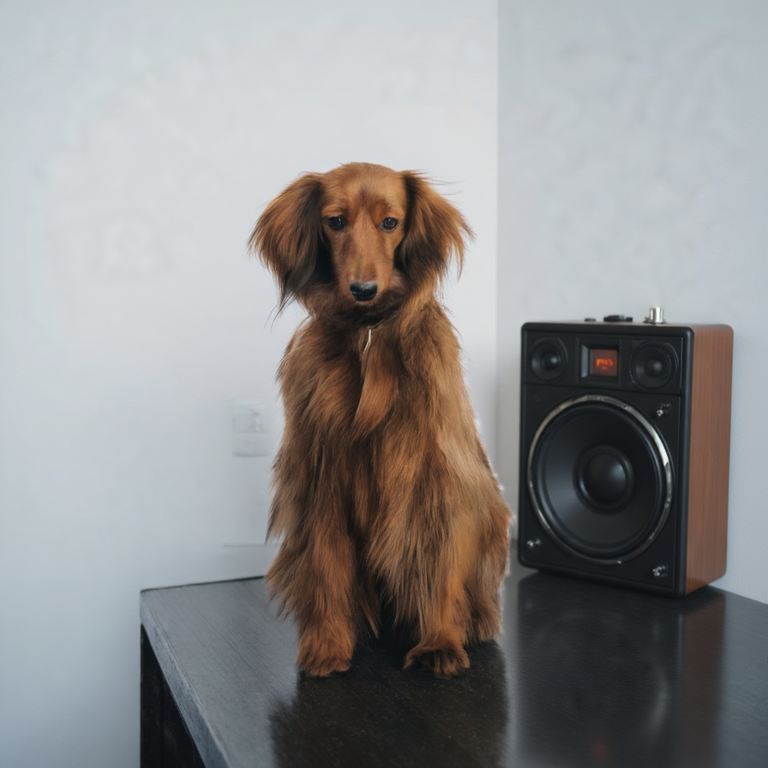}}
      & {\includegraphics[width=\carrierfigcell,height=\carrierfigcell]{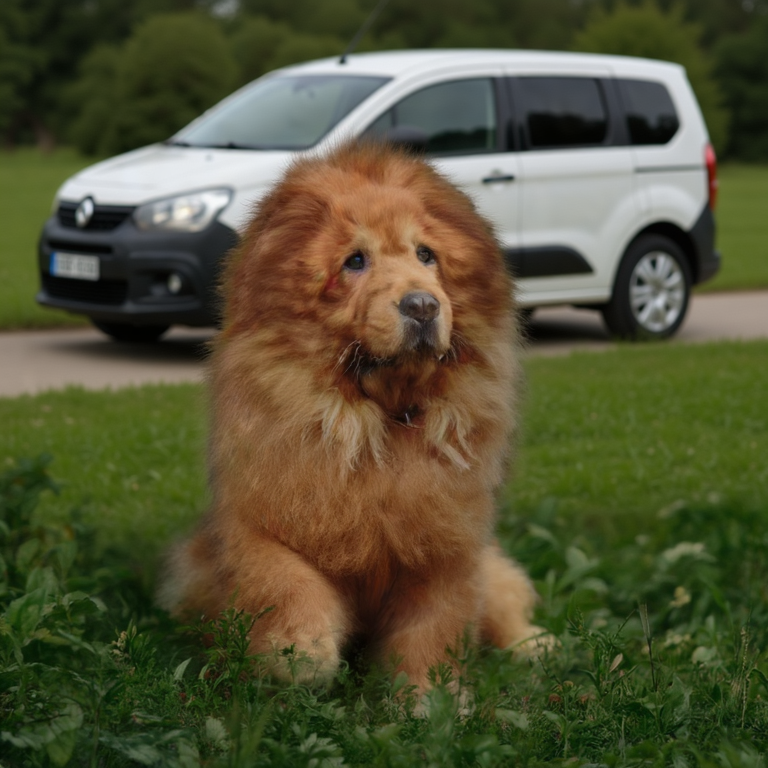}} \\
    };
    \matrix (t) [imagematrix,anchor=north west]
      at ($(m.north east)+(3.2mm,0)$) {
      {Dog} & {Teapot} \\[-2.3mm]
      {fountain} & {Arabian camel} \\[1.0mm]
      {\includegraphics[width=\carrierfigcell,height=\carrierfigcell]{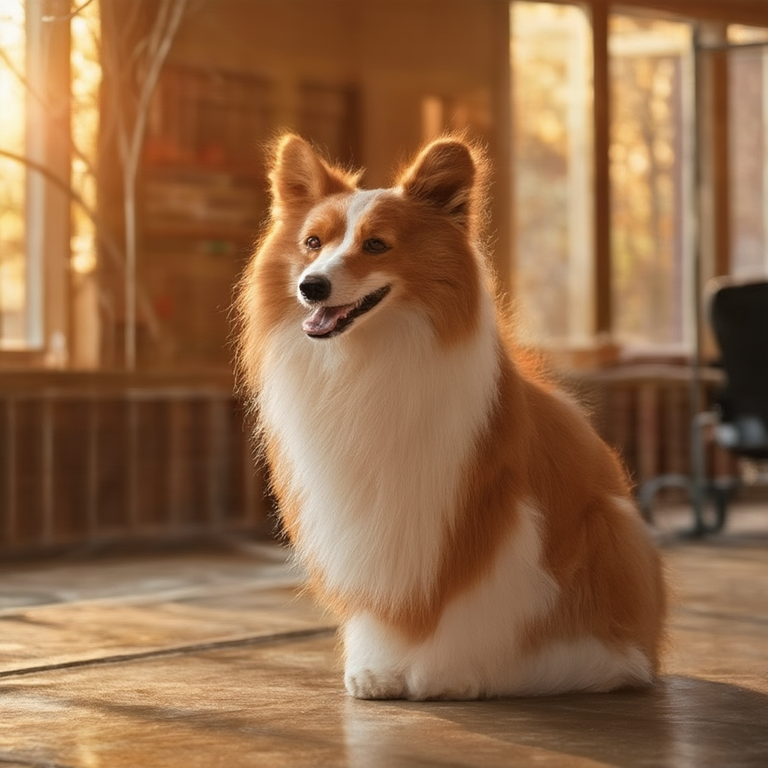}}
      & {\includegraphics[width=\carrierfigcell,height=\carrierfigcell]{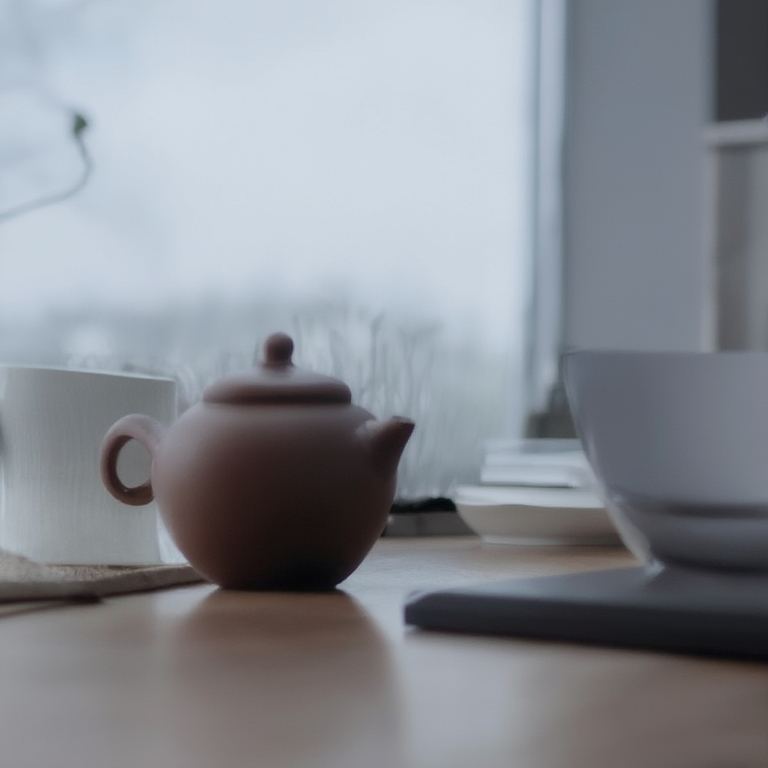}} \\
      {\includegraphics[width=\carrierfigcell,height=\carrierfigcell]{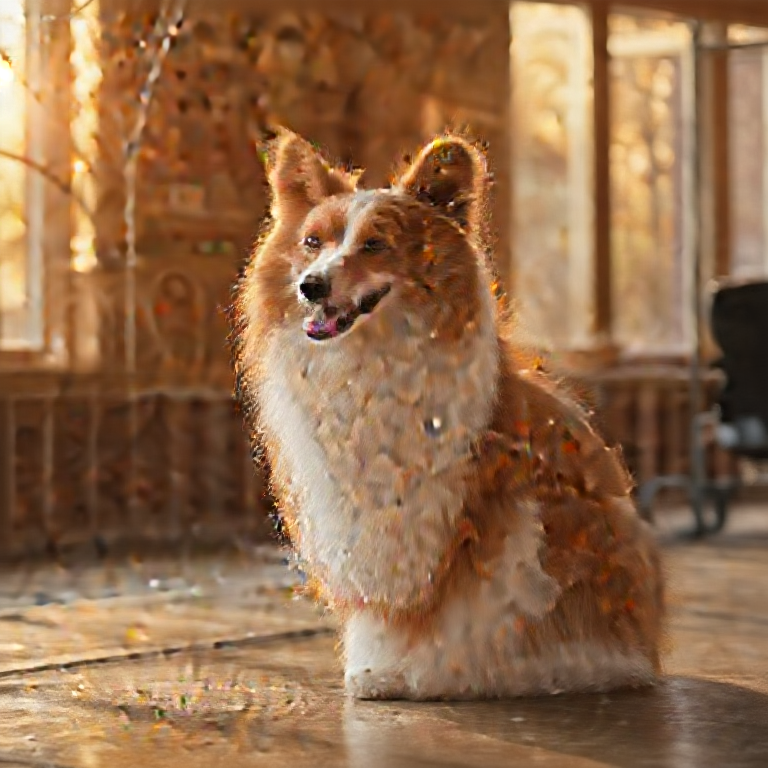}}
      & {\includegraphics[width=\carrierfigcell,height=\carrierfigcell]{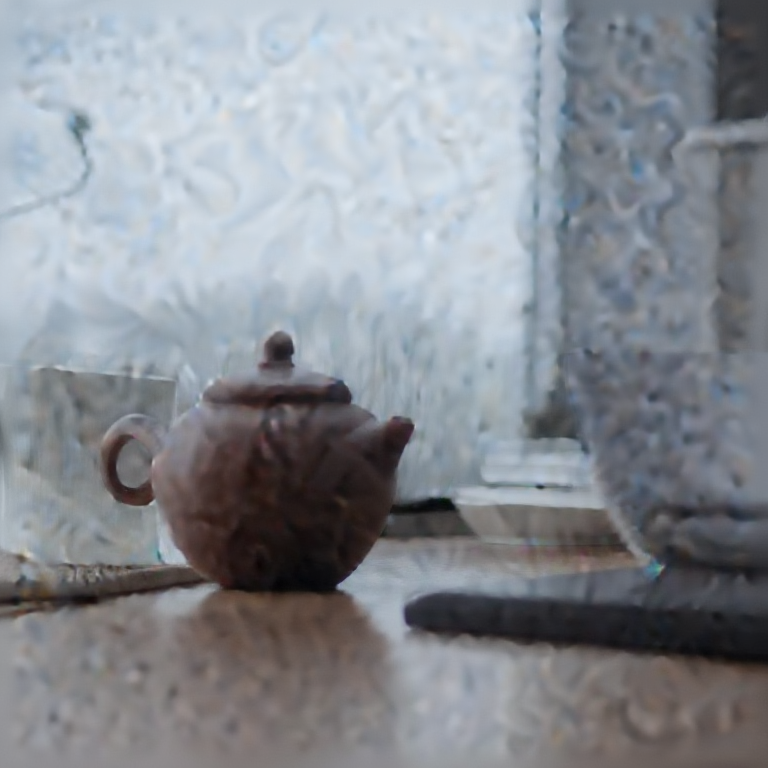}} \\[5.2mm]
      {\includegraphics[width=\carrierfigcell,height=\carrierfigcell]{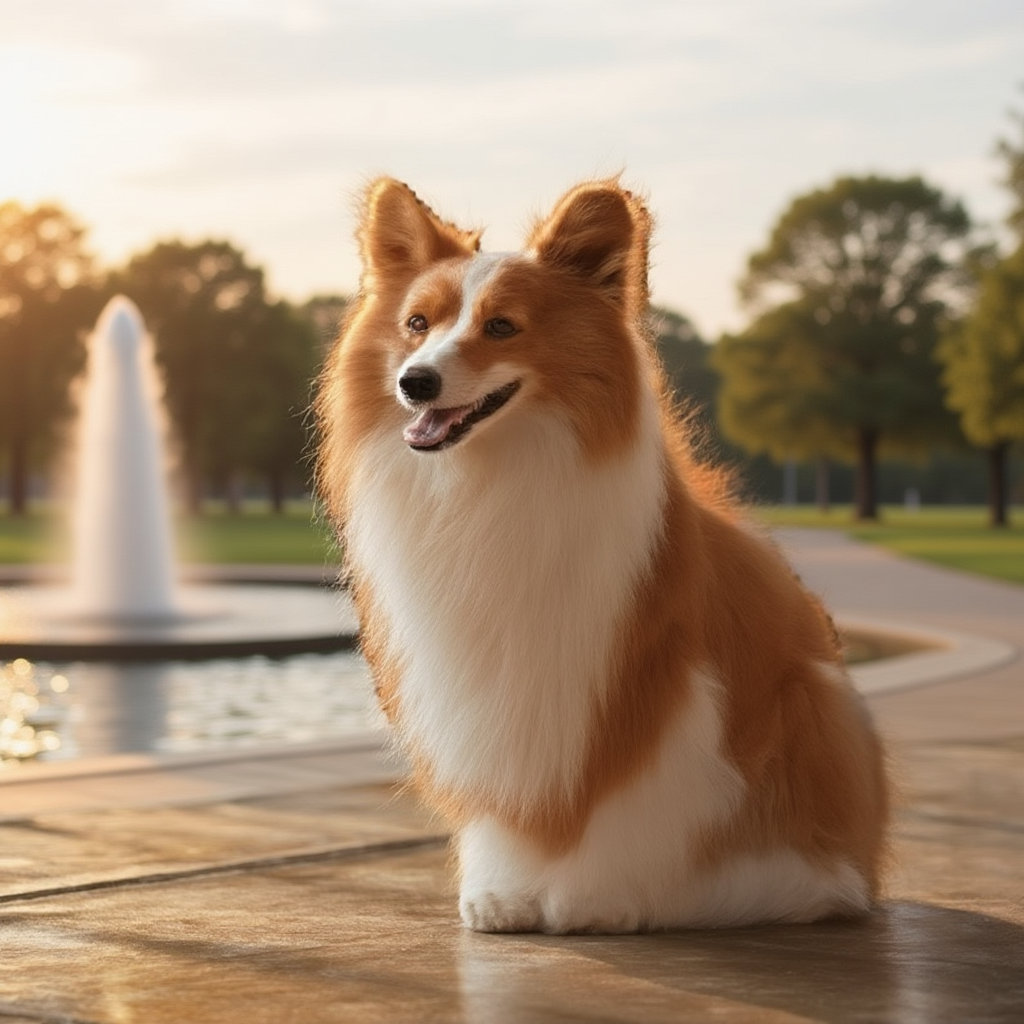}}
      & {\includegraphics[width=\carrierfigcell,height=\carrierfigcell]{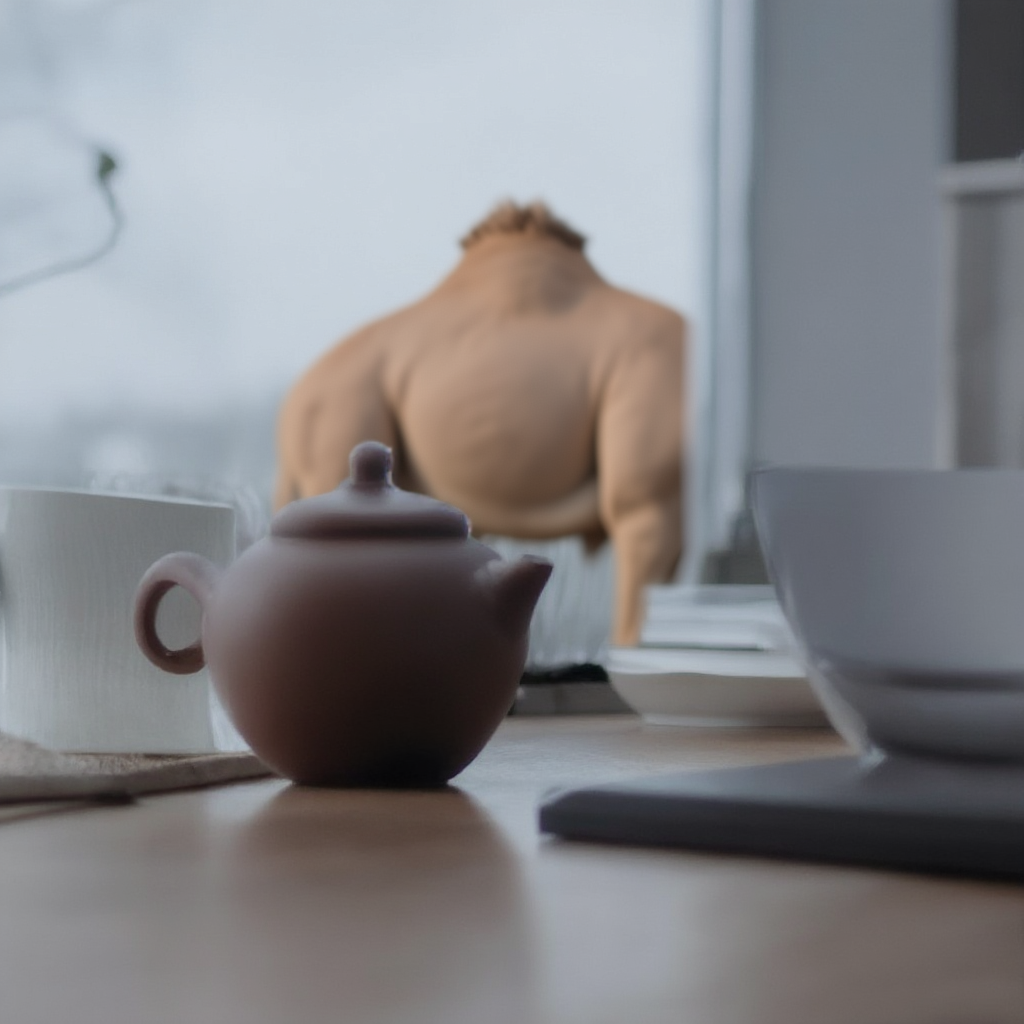}} \\
      {\includegraphics[width=\carrierfigcell,height=\carrierfigcell]{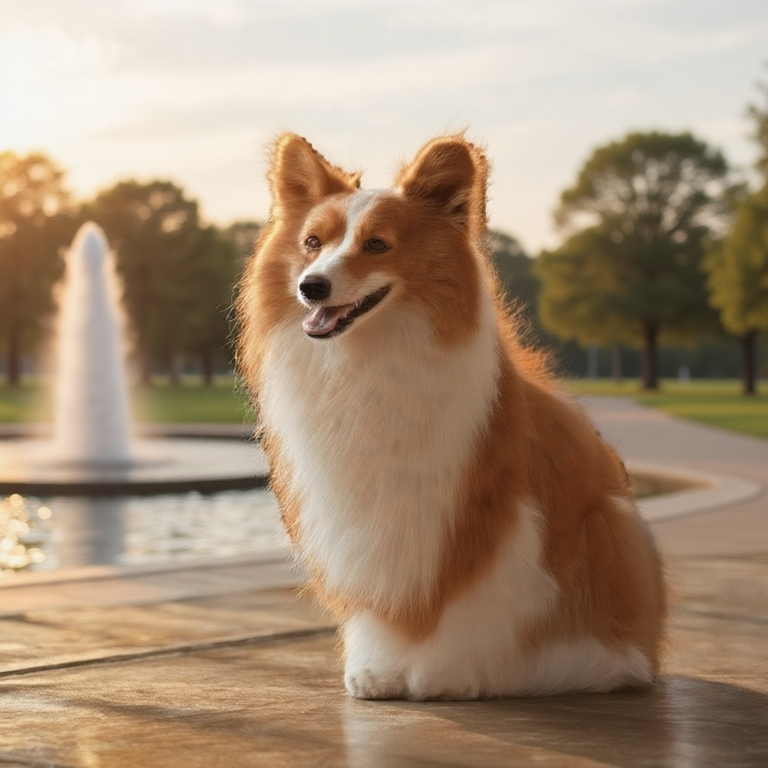}}
      & {\includegraphics[width=\carrierfigcell,height=\carrierfigcell]{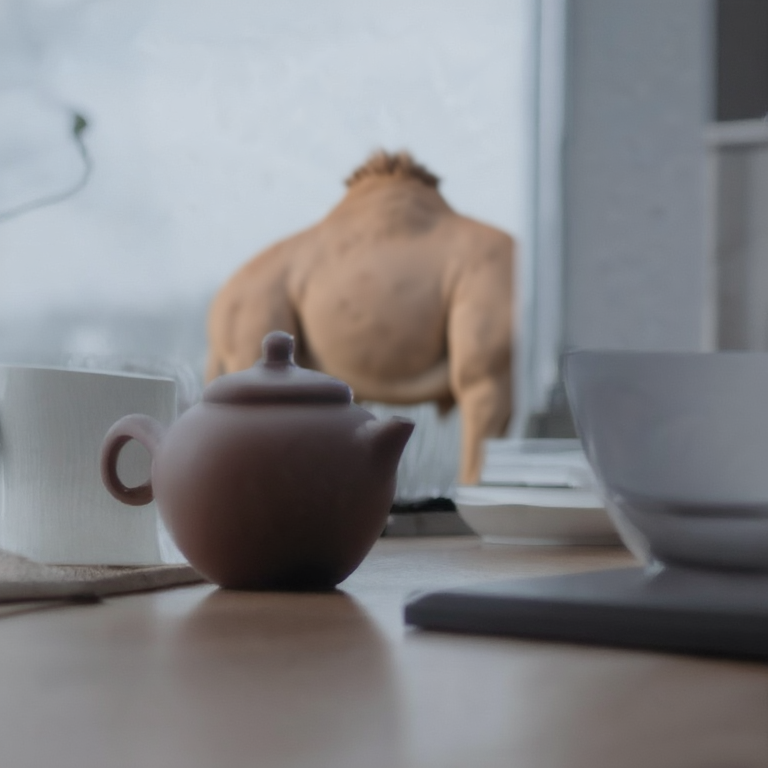}} \\
    };
    \node[carrierrow,anchor=east] (subject-label) at ($(m-1-1.west)+(-1.5mm,0)$) {Subject};
    \node[carrierrow,anchor=east] (target-label) at ($(m-2-1.west)+(-1.5mm,0)$) {Target};
    \node[carrierrow,anchor=east] (nc-clean-label) at ($(m-3-1.west)+(-1.5mm,0)$) {Clean};
    \node[carrierrow,anchor=east] (nc-attacked-label) at ($(m-4-1.west)+(-1.5mm,0)$) {Attacked};
    \node[carrierrow,anchor=east] (c-clean-label) at ($(m-5-1.west)+(-1.5mm,0)$) {Clean};
    \node[carrierrow,anchor=east] (c-attacked-label) at ($(m-6-1.west)+(-1.5mm,0)$) {Attacked};
    \coordinate (common-left-gray) at ($(m-3-1.west)+(-12.5mm,0)$);
    \coordinate (common-left-blue) at ($(m-5-1.west)+(-12.5mm,0)$);
    \foreach \c in {1,...,4} {
      \draw[flowarrow] ($(m-3-\c.south)+(0,-0.35mm)$) -- ($(m-4-\c.north)+(0,0.35mm)$);
      \draw[flowarrow] ($(m-5-\c.south)+(0,-0.35mm)$) -- ($(m-6-\c.north)+(0,0.35mm)$);
    }
    \foreach \c in {1,2} {
      \draw[flowarrow] ($(t-3-\c.south)+(0,-0.35mm)$) -- ($(t-4-\c.north)+(0,0.35mm)$);
      \draw[flowarrow] ($(t-5-\c.south)+(0,-0.35mm)$) -- ($(t-6-\c.north)+(0,0.35mm)$);
    }
    \begin{scope}[on background layer]
      \node[fit=(common-left-gray)(subject-label)(target-label)(nc-clean-label)(nc-attacked-label)(m-1-1)(t-1-2)(t-4-2),fill=black!2,
        draw=black!22,line width=0.45pt,rounded corners=1.1mm,
        inner xsep=1.6mm,inner ysep=1.7mm] (nocblock) {};
      \node[fit=(common-left-blue)(c-clean-label)(c-attacked-label)(m-5-1)(m-6-4),fill=blue!5,
        draw=blue!48!black,line width=0.55pt,rounded corners=1.1mm,
        inner xsep=1.6mm,inner ysep=1.7mm] (ntblock) {};
      \node[fit=(t-5-1)(t-6-2),fill=orange!8,draw=orange!65!black,
        line width=0.55pt,rounded corners=1.1mm,
        inner xsep=1.45mm,inner ysep=1.7mm] (tblock) {};
    \end{scope}
    \node[grouptitle,anchor=south west,text=black!82,fill=white]
      at ($(nocblock.north west)+(1.0mm,-0.65mm)$) {No-Carrier baseline};
    \node[grouptitle,anchor=south west,text=blue!48!black,fill=white]
      at ($(ntblock.north west)+(1.0mm,-0.65mm)$) {Non-Target Carrier};
    \node[grouptitle,anchor=south west,text=orange!65!black,fill=white]
      at ($(tblock.north west)+(1.0mm,-0.65mm)$) {Target Carrier};
  \end{tikzpicture}
  \caption{Qualitative comparison of primary subject preservation with and without a visually secondary carrier. Each column
shows one source-target case. No Carrier attacks visibly degrade the subject, whereas
the shown carrier-based CIRA outputs preserve the personalized subject more faithfully. The first four carrier cases use a Non-Target Carrier, while the final two use a Target Carrier. Both the No Carrier and Carrier-based settings are evaluated under the same strong attack strength and the same attack steps.}
  \label{fig:carrier-motivation}
\end{figure}

\section{related work}
\label{sec:related_work}
\subsection{Unrestricted and Generative Adversarial Attacks}
Traditional adversarial attacks mislead classifiers through small, often imperceptible perturbations around a fixed source image~\citep{szegedy2014,goodfellow2015}. Targeted attacks impose a stricter success criterion than untargeted attacks: the classifier must predict a designated
target class rather than merely misclassify the image.

Unrestricted attacks relax these norm constraints and permit larger changes by diffusion model while retaining human-recognizable semantic content\citep{brown2018,song2018}.
Our work follows this broader setting and further requires the original subject to remain the primary image content.

Concept-based attacks further extend the adversarial generation from a fixed image to a personalized subject concept $C_s$, allowing variations in pose, viewpoint, composition, and background while preserving subject recognizability~\citep{zhang2024constructing, conceptbased}. However, strong global guidance can distort the primary subject’s identity-defining visual characteristics. Building upon this setting, we study primary subject preservation through spatially structured adversarial generation without limiting attack strength.

Generative attacks search beyond small perturbations around a fixed image.~\citet{aca} optimize adversarial content on a generative manifold, while ~\citet{advdiff} and~\citet{diffattack} incorporate classifier objectives into diffusion sampling and latent reconstruction. Building upon  image inversion and trajectory guidance, we attack a clean image containing spatially seprated subject and carrier from its intermediate generative state.

NatADiff~\citep{natadiff} guides diffusion toward the intersection of source and target classes, using augmented classifier guidance and time-travel sampling to generate transferable adversarial images. It also introduces the target-related features during generation, making it closely related to our use of target-related content. Its original setting does not explicitly require primary subject preservation, We therefore compare with an adaptation of NatADiff using the same subject-specific LoRA, evaluating its attack performance and subject preservation.

\subsection{Object-Aware Gradients and Regional Attack Effects}

Prior works connects adversarial gradients with discriminative image regions. Empirically,~\citet{superpixel} find object-region perturbations are more effective than background perturbations, while~\citet{FIT} show that gradients correlate with objects of interest and exploit important features to improve transferability. Theoretically,~\citet{doublemainbody} relate perceptually aligned gradients to off-manifold robustness, under which input gradients lie approximately along data-manifold directions. 
Therefore, we examine whether strong global guidance concentrates updates on the primary subject, a visually secondary carrier changes this regional distribution, and introducing extra target-related features affect transferability.

\subsection{Personalized Generation and Subject Preservation}

Subject-specific generation aims to preserve a specific subject across different scenes, poses, and viewpoints with a small reference set.
DreamBooth adapts pretrained diffusion models for subject-specific generation~\citep{dreambooth}, while LoRA ~\citep{lora} provides a parameter-efficient adaptation mechanism by learning low-rank weight updates. We use DreamBooth reference images to characterize each personalized concept and train a corresponding subject-specific LoRA that governs source-candidate generation, carrier construction, and the subsequent attack process.
We evaluate primary subject preservation under strong classifier guidance jointly with attack success.

We assess primary subject preservation through the representation and spatial consistency. We use DINOv3 cosine similarity~\citep{dinov3} to measure the visual consistency between clean and attacked image, and SAM3 IoU tests spatial consistency.
\subsection{Segmentation-Guided Image Editing and Spatial Evidence}

Segmentation-guided editing provides spatial structure for controllable carrier generation. SAM3~\citep{sam3} supports prompt detection and segmentation, while existing diffusion-based editing enables localized construction and object insertion within editable regions~\citep{flux}. We use SAM3 to localize the primary subject and construct a carrier outside subject through compositing or mask-guided inpainting.

The carrier is intended to remain visually secondary while the subject remains the primary image content. Research on composition and object importance relates perceived prominence to size, position, contrast, and sharpness, while showing that saliency alone does not determine perceived importance~\citep{intrestsofobject2008,objectimpotance2011}. Semantic information also guides human attention beyond low-level saliency~\citep{meaningguidance2017}. We therefore evaluate primary-subject judgments directly through a blinded human study, DINOv3 similarity and spatial consistency.

Grad-CAM can localize image regions associated with a classifier
prediction~\citep{gradcam}. We use it to visualize prediction-related regions instead of guidance, not as a direct measurement of attack gradients. Inspired by the prediction sensitivity measurement~\citep{whitening}, we additionally remove the visible carrier region to measure the resulting changes in target predictions.

\section{Theoretical Analysis of Carrier Effects}
\label{sec:carrier_principles}
A visually secondary carrier can affect adversarial generation in two ways: First, under global RMS normalization, the carrier gradient component changes the allocation
of the applied attack update across spatial regions. Second, when increasingly target-related carrier characteristics produce a larger ideal target margin, this yields a higher conditional lower bound on targeted transferability.  We formalize these two effects below, with complete proofs provided in
Appendics~\ref{app:attack_update_allocation}
and~\ref{app:transferability_improvement}.

\subsection{Carrier-Induced Reduction of the Normalized Subject Update}
\label{sec:carrier_preservation_principle}
Consider the targeted classifier gradient $g_t$ used in the global guidance update
defined in Section~\ref{sec:attack}. We decompose it as \[
g_t=g_{S,t}+g_{C,t}+g_{R,t}.
\]
Here, $g_{S,t}$, $g_{C,t}$, and $g_{R,t}$ denote the gradient components
associated with the primary-subject, carrier, and remaining regions,
respectively. A region is considered class-related if a change within it can shift the classifier's prediction toward or away from the target class.

\begin{assumption}[Matched regional components]
\label{ass:matched_components}
The No-Carrier and carrier-based condition are compared while holding $g_{S,t}$,
$g_{R,t}$, the number of gradient elements $n$, and the guidance coefficient
$\lambda\sigma_t$ fixed. 
\[
\begin{aligned}
\text{No-Carrier:}\qquad
&g_t^{\mathrm{NC}}
=
g_{S,t}+g_{R,t},\\
\text{Carrier:}\qquad
&g_t^{\mathrm{C}}
=
g_{S,t}+g_{C,t}+g_{R,t}.
\end{aligned}
\]
This matched comparison isolates the effect of introducing the carrier
component $g_{C,t}$. It is an analytical comparison and does not require
independently constructed carrier-based and No-Carrier images to produce
identical regional gradients.

\end{assumption}

The superscripts $\mathrm{NC}$ and $\mathrm{C}$ denote the No-Carrier and Carrier
conditions.

Let $\delta_{S,t}$ denote the subject-region component of the classifier-guided
velocity increment at return step $t$. Under global RMS normalization, the
corresponding subject updates are
\[
\begin{aligned}
\text{No-Carrier:}\qquad
\delta_{S,t}^{\mathrm{NC}}
&=
\lambda\sigma_t\sqrt{n}\,
\frac{g_{S,t}}
{\sqrt{
\|g_{S,t}\|_2^2+
\|g_{R,t}\|_2^2
}},
\\
\text{Carrier:}\qquad
\delta_{S,t}^{\mathrm{C}}
&=
\lambda\sigma_t\sqrt{n}\,
\frac{g_{S,t}}
{\sqrt{
\|g_{S,t}\|_2^2+
\|g_{C,t}\|_2^2+
\|g_{R,t}\|_2^2
}}.
\end{aligned}
\]

\begin{theorem}[Carrier-induced reduction of the normalized subject update]
\label{thm:subject_update}
Under Assumption~\ref{ass:matched_components}, when the carrier region contributes a class-related gradient component, we have $\|g_{C,t}\|_2>0$, and then the subject-region component of the globally
normalized update satisfies
\[
\left\|
\delta_{S,t}^{\mathrm{C}}
\right\|_2
<
\left\|
\delta_{S,t}^{\mathrm{NC}}
\right\|_2.
\]
\end{theorem}

Thus, at the same guidance strength, the carrier provides an additional region to accommodate the attack update and reduces the normalized update applied to the subject. This result does not imply a reduction in the raw subject-gradient magnitude or directly guarantee final-image preservation.
The complete proof is provided in Appendix~\ref{app:attack_update_allocation}, and  the corresponding empirical preservation results are reported in Section~\ref{sec:preservation}.

\subsection{Targeted Transferability with Increasingly Target-Related Carriers}
\label{sec:carrier_transferability_principle}

To analyze the effect of increasingly target-related carrier characteristics
on transferability, let $\mathcal{D}=\{\mathrm{NC},\mathrm{NT},\mathrm{H},\mathrm{T}\}$ denote the
No-Carrier, Non-Target Carrier, Hybrid Carrier, and Target Carrier conditions. For a fixed source, target, seed, and attack configuration, let
$x_d^{\mathrm{adv}}$ denote the final adversarial image under condition
$d\in\mathcal{D}$.

We assume the existence of an ideal semantic classifier $p^*$ whose class
probabilities reflect the underlying semantic content of an image,
independently of the errors or biases of any particular practical
classifier. Although $p^*$ is not directly observable, it provides a
theoretical reference for the natural semantic decision rule associated
with a given image.We treat practical BlackBox classifiers \(m\sim\mathcal Q\) as different approximations to it. The ideal classifier is a theoretical reference rather than an observable model, and its margin characterizes the target-related semantic evidence present in the final image. Their discrepancy under condition $d$ is measured by
\[
D_{m,d}
=
D_{\mathrm{KL}}
\left(
p^*(\cdot\mid x_d^{\mathrm{adv}})
\;\middle\|\;
p_m(\cdot\mid x_d^{\mathrm{adv}})
\right).
\]
For target class $t$, we define the ideal targeted Top-1 margin and transfer probability as
\[
\gamma_d^*
=
p_t^*\!\left(x_d^{\mathrm{adv}}\right)
-
\max_{y\neq t}
p_y^*\!\left(x_d^{\mathrm{adv}}\right),
\qquad
T_d
=
\Pr_{m\sim Q}
\left[
\operatorname{Top1}_m
\left(x_d^{\mathrm{adv}}\right)=t
\right].
\]
The ideal margin $\gamma_d^*$ measures the underlying semantic evidence
supporting the target class over its strongest competing class.

\begin{assumption}[Ideal-margin ordering]
\label{ass:margin_ordering}
When the realized carrier content becomes increasingly related to the target
while the source and attack configuration remain fixed, the underlying
semantic evidence for the target class does not decrease:
$\gamma_{\mathrm{T}}^*
\geq
\gamma_{\mathrm{H}}^*
\geq
\gamma_{\mathrm{NT}}^*
\geq
\gamma_{\mathrm{NC}}^*$.
This ordering formalizes the semantic progression of the realized carrier
content rather than an ordering based only on the nominal condition labels.
\end{assumption}

\begin{assumption}[Uniform classifier discrepancy]
\label{ass:classifier_discrepancy}
All BlackBox classifiers share the same label space
and classification task with the ideal classifier. We therefore assume
that their expected discrepancy from the ideal classifier is bounded above
by the same constant across conditions:
\[
\mathbb{E}_{m\sim\mathcal{Q}}\!\left[D_{m,d}\right]
\leq \bar{\kappa},
\qquad
\forall d\in\mathcal{D}.
\]
This assumption does not require different BlackBox classifiers or carrier
conditions to have identical discrepancies. It only rules out a
condition-specific increase in model discrepancy that would itself explain
the observed difference in transferability.
\end{assumption}

For each condition $d$ with a positive ideal target margin, we represent its certified
targeted Top-1 transferability lower bound as
\[
L_d
:=
\left[
1-
\frac{2\bar{\kappa}}
{(\gamma_d^*)^2}
\right]_+.
\]

\begin{theorem}[Conditional targeted Top-1 transferability bound]
\label{thm:transferability}
Under Assumptions~\ref{ass:margin_ordering}
and~\ref{ass:classifier_discrepancy}, the targeted Top-1 transfer probability
satisfies
\[
T_d\geq L_d.
\]
Moreover, within the positive-margin regime of the bound,
Assumption~\ref{ass:margin_ordering} yields
\[
L_{\mathrm{T}}
\geq
L_{\mathrm{H}}
\geq
L_{\mathrm{NT}}
\geq
L_{\mathrm{NC}}.
\]
\end{theorem}

The complete proof is provided in
Appendix~\ref{app:transferability_improvement}, and the corresponding empirical
cross-model transferability results are reported in Section~\ref{transferability}.
\section{Methodology}
\label{sec:method}

\subsection{Problem Formulation}
\label{sec:formulation}

Following the concept-based attack setting of~\citet{conceptbased}, we study targeted adversarial generation. Given a source image $x$ and a binary mask $m_s$ for its primary subject, we define the subject and background regions as
$x_s=m_s\odot x$ and $x_b=(1-m_s)\odot x$. After constructing the visually secondary carrier, with $m_c$ denotes its visible mask, satisfying $m_s\odot m_c=0$ to ensure spatial separation from the primary subject.

Let $c_v$ denote carrier's visual-semantic
category and $c_t$ the attack target category. Given the clean image
$x_{\mathrm{clean}}$, our objective is to generate $x_{\mathrm{adv}}$ such that
$f(x_{\mathrm{adv}})=c_t$, while preserving the subject's
appearance from $x_{\mathrm{clean}}$ to $x_{\mathrm{adv}}$ and retaining it
as the primary image content.

\subsection{Source Preparation and LoRA}\label{sec:source}
We use the subject-specific LoRA to expand the limited reference images into source candidates varying in pose, layout, and background. We randomly select a source image \(x\) and the LoRA remains active during subsequent adversarial generation to maintain the primary subject concept. 

We apply SAM3~\citep{sam3} to $x$ for the primary subject mask $m_s$ and the complementary background mask $m_b = 1 - m_s$. To avoid overlap with the primary subject, we construct the carrier and its surrounding scene within $m_b$. The masks govern image construction; they are not used to project the classifier gradient in latent space.

\subsection{Carrier-Based Background Construction}\label{sec:construction}

 We use FLUX~\citep{flux} to construct a clean carrier-based image through two mechanisms: direct compositing and mask-guided inpainting.

\begin{enumerate}
    \item \textbf{Direct compositing.}
    For CRA, FLUX first generates an independent background image containing the visually secondary carrier and composites the primary subject pixels of $x$ according to subject mask $m_s$, producing the composite image \(x_{\mathrm{clean}}^{\mathrm{comp}}\).

    \item \textbf{Mask-guided inpainting.}
   For CIRA and JIA, $x$ remains the image condition while FLUX reconstructs the region defined by background mask $m_b$ using a carrier condition prompt $p_c$ that specifies the carrier. CIRA completes and freezes the clean inpainting $x_{\mathrm{clean}}^{\mathrm{inp}}$ before the attack, whereas JIA introduces adversarial guidance during the same inpainting trajectory.
\end{enumerate}
\[
x_{\mathrm{clean}}^{\mathrm{comp}}
=
m_s \odot x
+
m_b \odot x_{\mathrm{b}}(p_c),
\qquad
x_{\mathrm{clean}}^{\mathrm{inp}}
=
\operatorname{Inpaint}(x, m_b, p_c).
\]
Here, $p_c$ denotes the construction prompt associated with the carrier condition. We construct three carrier conditions. \textbf{Non-Target Carrier}: \(c_v \neq c_t\), the carrier shares moderate visual characteristics with the target while remaining semantically distinct. \textbf{Hybrid Carrier}: The carrier retains its non-target identity while exhibiting additional visual characteristics associated with $c_t$. \textbf{Target Carrier}: the carrier directly instantiates the target category, \(c_v = c_t\). These conditions provide varying degrees of target-related content, allowing a flexible trade-off between semantic separation from the target and transferability.

\subsection{Image Inversion and Finite-Return Attack}\label{sec:attack}

CRA and CIRA apply finite-return attacks to the clean images constructed in Section~\ref{sec:construction}.
Inspired by the inversion-based attacks raised by \citet{aca} and \citet{diffattack}, we map \(x_{\mathrm{clean}}\) to an intermediate FLUX state on a reconstruction trajectory and apply global classifier guidance at selected return steps. Starting from a clean image avoids regeneration from unconstrained noise and also enables evaluation of primary subject preservation.
Specifically, clean images are inverted to an intermediate state as
$z_\tau=\operatorname{Inv}_{\mathrm{FLUX}}(x_{\mathrm{clean}};\tau)$.
Starting from $z_{\tau}$, classifier guidance is applied during the finite reconstruction toward $z_0$. We estimate the clean state and target loss and inject its gradient with respect to $z_t$ into the FLUX velocity:
\begin{align*}
\hat{z}_0^{(t)}
&= z_t-\sigma_t v_\theta(z_t,t,p),
\qquad
\hat{x}_0^{(t)}
= D\!\left(\hat{z}_0^{(t)}\right),
\\
g_t
&= \nabla_{z_t}
\operatorname{CE}\!\left(
f\!\left(P_f\!\left(\hat{x}_0^{(t)}\right)\right),c_t
\right),
\\
v_t^{\mathrm{adv}}
&= v_\theta(z_t,t,p)
+\mathbf{1}[t\in\mathcal{A}]\,\lambda\sigma_t
\operatorname{Norm}_{\mathrm{RMS}}(g_t).
\end{align*}

Here, $p$ denotes the text prompt describing the personalized subject,
carrier, and surrounding scene; $v_\theta$ is the FLUX velocity field;
$\sigma_t$ is the noise level at step $t$; $D$ is the VAE decoder; $P_f$ is the classifier preprocessing function. $\operatorname{CE}$ denotes cross-entropy loss, $\lambda$ is the guidance
scale, and $\mathcal{A}$ is the set of guided return steps.
$\operatorname{Norm}_{\mathrm{RMS}}$ applies global RMS normalization
to the classifier gradient. Its effect on the subject-region update is analyzed in
Section~\ref{sec:carrier_preservation_principle}, and detailed inversion process is provided in the Appendix~\ref{app:inversion_return}.

\subsection{Framework Overview and Realizations}
\label{sec:realizations}

Our framework consists of three stages: personalized source preparation, carrier
construction, and adversarial realization. Source preparation generates a recognizable
personalized image and obtains its subject and background masks. Carrier construction
then introduces a visually secondary carrier into the background through direct
compositing or mask-guided inpainting. Finally, global classifier guidance produces
the targeted adversarial image. CRA, CIRA, and JIA differ in how carrier construction
is coupled with this final adversarial stage. Here, $x$ denotes the source image, $m_b$ the background mask,
$p_c$ the carrier condition prompt, and $c_t$ the target class. We use $z_\tau$ for the
finite-inversion latent and $x_{\mathrm{adv}}$ for the final adversarial image.

\textbf{Composite Reconstruction Attack (CRA)} applies the finite-return attack in Section~\ref{sec:attack} to the clean composite:
\[
x_{\mathrm{clean}}^{\mathrm{comp}}
\xrightarrow{\mathrm{Inv.}}
z_{\tau}
\xrightarrow{\mathrm{Guided\ Recon.}}
x_{\mathrm{adv}}^{\mathrm{CRA}}.
\]

\textbf{Clean Inpainting Reconstruction Attack (CIRA)} completes and freezes the clean inpainting image  \(x_{\mathrm{clean}}^{\mathrm{inp}}\) before applying the finite-inversion attack defined in Section~\ref{sec:attack}:

\[
x_{\mathrm{clean}}^{\mathrm{inp}}
\xrightarrow{\mathrm{Inv.}}
z_{\tau}
\xrightarrow{\mathrm{Guided\,Recon.}}
x_{\mathrm{adv}}^{\mathrm{CIRA}}.
\]

\textbf{Joint Inpainting Attack (JIA)} injects global classifier guidance directly into the FLUX inpainting trajectory, jointly performing carrier construction and adversarial steering:
\[
(x,m_b,p_c)
\xrightarrow{\mathrm{Inpainting+Guidance}(c_t)}
x_{\mathrm{adv}}^{\mathrm{JIA}}.
\]

Detailed implementations are provided in Appendix~\ref{app:algorithm}.

\section{Experiments}
\label{sec:experiment}

\subsection{Experiment Setup}

\textbf{Data and implementation}: We evaluate 20 personalized subjects from the DreamBooth dataset and 30 ImageNet-1K target classes. Each subject is associated with a subject-specific FLUX LoRA and one selected source image. We use FLUX.1-Kontext-dev by ~\citet{flux}, 50 FLUX steps, classifier guidance scale $0.5$, and global RMS-normalized guidance. ResNet-50 works as the WhiteBox evaluator.  Implementation detail are provided in Appendix~\ref{app:experiment_set_up}.

\textbf{Baselines}: We compare against two baselines. No-Carrier implements the original concept-based attack by on FLUX, applying global adversarial guidance directly to the personalized source image without visually secondary carrier construction~\citep{conceptbased}. NatADiff is adapted to FLUX with the same LoRA to evaluate its generation-based attack under personalized conditioning~\citep{natadiff}. Its clean and attacked outputs are generated using matched seeds and conditioning, with adversarial guidance disabled and enabled, respectively. Detailed are provided in Appendix~\ref{app:baseline_realization}

\textbf{Evaluation}: We report WhiteBox targeted results on ResNet-50 and mean BlackBox targeted transferability. Primary subject preservation is measured using DINOv3 similarity on fixed subject crops of clean image and attack image pair~\citep{dinov3} and SAM3 mask IoU obtained by re-segmenting~\citep{sam3}. JIA's preservation is evaluated with clean image with classifier guidance disabled. Stronger guidance attack results are also presented to examine the preservation cost of increasing attack strength. To verify that carrier construction alone does not account for attack
success, we additionally report conditional ASR on samples for which
the target class is absent from the clean image's predictions. The resulting success rates remain close to the unconditional ASR,
showing that high attack success generally requires adversarial guidance
rather than carrier construction alone. Experiment details are provided in Appendix~\ref{app:conditional_asr}.BlackBox clean prediction provided in Appendix ~\ref{app:clean_black}.

\begin{table}[ht]
\centering
\caption{Main attack and subject-preservation results over 600
source-target pairs. Cond. Top-5 reports ASR only on samples for which the target class is absent
from the clean Top-5 predictions. For NatADiff adaptation, preservation metrics are computed on the 68.17\%
of pairs for which the designated subject is detected in both the clean and attacked
images (Appendix~\ref{app:baseline_realization}). Hybrid rows indicate the requested
construction condition; CIRA and JIA do not consistently realize its target-related
attributes (Appendix~\ref{hybrid_incomplete}).}
\label{tab:main_attack_analysis}

\setlength{\tabcolsep}{3pt}
\renewcommand{\arraystretch}{1.05}

\begin{tabular}{@{}llcccccc@{}}
\toprule
& &
\multicolumn{3}{c}{WhiteBox} &
\multicolumn{1}{c}{BlackBox} &
\multicolumn{2}{c}{Subject Preservation} \\
\cmidrule(lr){3-5}
\cmidrule(lr){6-6}
\cmidrule(l){7-8}

Method & Carrier
& Top-1 $\uparrow$
& Top-5 $\uparrow$
& Cond. Top-5 $\uparrow$
& Mean Top-5 $\uparrow$
& DINO $\uparrow$
& IoU $\uparrow$ \\
\midrule

No-Carrier & None
& 93.33 & 98.00 & -- & 2.17 & 0.9530 & 0.9940 \\

NatAdiff & N/A
& 31.83 & 45.67 & -- & 19.27 & 0.9411 & 0.9649 \\

\midrule

CRA & Non-Target
& 96.50 & 99.50 & 99.49 & 7.23 & 0.9857 & 0.9934 \\
CRA & Hybrid
& 98.33 & 99.67 & 99.64 & 15.17 & 0.9818 & 0.9898 \\
CRA & Target
& 98.00 & 99.67 & 99.56 & 34.23 & 0.9845 & 0.9924 \\

\addlinespace[2pt]

CIRA & Non-Target
& 95.00 & 98.50 & 98.48 & 3.00 & 0.9799 & 0.9919 \\
CIRA & Hybrid
& 95.00 & 98.83 & 98.82 & 3.07 & 0.9805 & 0.9928 \\
CIRA & Target
& 99.67 & 99.83 & 99.69 & 52.13 & 0.9809 & 0.9929 \\

\addlinespace[2pt]

JIA & Non-Target
& 29.67 & 55.83 & 55.39 & 5.10 & 0.9958 & 0.9930 \\
JIA & Hybrid
& 27.83 & 55.50 & 55.28 & 5.30 & 0.9960 & 0.9935 \\
JIA & Target
& 74.17 & 90.67 & 82.50 & 55.07 & 0.9959 & 0.9932 \\

\bottomrule
\end{tabular}
\end{table}

\subsection{Carrier Contribution and Subject Preservation}
\label{sec:preservation}

\begin{table}[ht]
\centering
\caption{Primary subject preservation under strong attacks. NatADiff adaptation preservation metrics are computed on the 51.00\% of pairs for which the designated subject is detected in both the clean and attacked images (Appendix~\ref{app:baseline_realization}).}
\label{tab:subject_preservation}

\small
\setlength{\tabcolsep}{7pt}
\renewcommand{\arraystretch}{1.05}

\begin{tabular}{@{}lccc@{}}
\toprule
&
\multicolumn{1}{c}{Attack} &
\multicolumn{2}{c}{Subject Preservation} \\
\cmidrule(lr){2-2}
\cmidrule(l){3-4}

Method / Condition
& Top-1 ASR $\uparrow$
& DINO $\uparrow$
& SAM3 IoU $\uparrow$ \\
\midrule

NatADiff
& 98.83
& 0.7461
& 0.9149 \\

No-Carrier
& 100.00
& 0.7928
& 0.9694 \\

Target Carrier $\cdot$ CRA
& 100.00
& 0.9345
& 0.9666 \\

Target Carrier $\cdot$ CIRA
& 100.00
& 0.9809
& 0.9695 \\

\bottomrule
\end{tabular}
\end{table}

Theorem~\ref{thm:subject_update} predicts that, under the matched comparison, a
class-related carrier component reduces the globally normalized update applied to the
primary-subject region. We therefore compare carrier-based and No-Carrier attacks under the same strong attack.

In Table~\ref{tab:subject_preservation}, No-Carrier, Target-Carrier
CRA, and Target-Carrier CIRA all achieve 100\% WhiteBox Top-1 attack success.
CRA and CIRA retain DINO similarities of 0.9345 and 0.9809,
compared with 0.7928 for No-Carrier. These results demonstrate improved subject preservation under the same
strong guidance setting and at matched attack success, consistent with
Theorem~\ref{thm:subject_update}. Together, DINO, SAM3 IoU, human evaluation, and clean subject Top-5 by classifiers assess subject-category recognizability, feature and spatial consistency, and human category dominance, respectively.

Table~\ref{tab:main_attack_analysis} shows that
CRA and CIRA achieve high WhiteBox Top-1 attack success with DINO
similarities of 0.9799-0.9857. Our NatADiff adaptation
achieves 31.83\% Top-1 attack success; stronger guidance raises this rate to 98.83\%
but reduces DINO similarity from 0.9411 to 0.7461. Its preservation metrics
are computed only on pairs in which the subject is detected in both the clean
and attacked images, covering 68.17\% and 51.00\% of pairs under the main
and stronger-guidance settings, respectively. These conditional measurements
do not account for subject-detection failures. Details
are provided in Appendix~\ref{app:preservation}, and the NatADiff adaptation
and missing-pair analysis are provided in
Appendix~\ref{app:baseline_realization}.

\subsection{Carrier Region Removal Ablation}
\label{sec:preservation_analysis}

To examine the extent to which the visually secondary carrier contributes predictive evidence to the
target class, we perform a whitening intervention on CRA outputs by
removing the visible carrier region while leaving the primary subject unchanged. Full intervention details and additional logit-level analyses are
provided in Appendix~\ref{app:whitening}.

Among the 1,130 CRA attacks that initially predict the target class as Top-1,
597 no longer predict the target after whitening, corresponding to a failure
rate of \(52.8\%\). With the qualitative examples in
Figure~\ref{fig:carrier_whitening}, this result shows that the visible carrier
region provides prediction-relevant evidence in a substantial fraction of
successful attacks.

\subsection{Carrier Target Characteristics and Transferability}
\label{transferability}

Target-related carrier characteristics provide a flexible trade-off between semantic separation and transferability. For CRA, mean BlackBox Top-5 transferability increases from 7.23\% with Non-Target Carriers to 15.17\% with Hybrid Carriers and 34.23\% with Target Carriers. CIRA and JIA show the same endpoint trend, reaching 52.13\% and 55.07\% with Target carriers. This progression aligns with the conditional targeted transferability bound in Theorem~\ref{thm:transferability}. 
Detailed analyses are provided in Appendix~\ref{app:BlackBox_Results}.

\subsection{Human Perception and Subject Prominence}
\label{sec:human_test_evluation}
The carrier is intended to remain visually secondary while the subject remains the primary image content. Because perceived object importance
depends on both visual and semantic factors, it cannot be inferred from low-level
saliency or object size alone
\citep{intrestsofobject2008,objectimpotance2011,meaningguidance2017}.

Therefore, we conduct a blinded human evaluation across three carrier conditions and three attack routes.
For each image, three independent
annotators select the primary subject from four randomly ordered labels: the
personalized subject, the attack target, and two distractors.
By majority vote, the
personalized subject is selected in 5,398 of 5,400 images
(99.96\%) selected as the primary image content, confirming that it remains the primary
in nearly all evaluated attacks. 
Evaluation details are provided in Appendix~\ref{app:human_test}.

\section{conclusion}
\label{conclusion}

We introduce the carrier, a visually secondary object that supports targeted adversarial generation while preserving the primary subject. Our analysis shows that, under matched regional gradients, carrier can reduce the globally normalized update applied to the subject.  Our experiments demonstrate
improved subject preservation without limiting attack strength, while Target Carriers substantially improve cross-model transferability over both the concept-based and NatADiff adaptation~\citep{conceptbased,natadiff}. Different degrees of realized target-related carrier content provide a flexible trade-off between semantic separation and transferability. Human evaluation confirms that the personalized subject remains the primary image content. Together, these results
establish the visible secondary carrier as a practical mechanism for improving
attack effectiveness, transferability, and primary subject
preservation.

\subsubsection*{Acknowledgments}
The work presented in this paper has been supported by UKRI Future Leaders Fellowship (Grant MR/S035176/1) and funded by the European Union under the Horizon Europe project AIGGREGATE (AI-enhanced collective intelligence for resilient, ethical and user-centric awareness and decision making in CCAM applications, Grant Agreement No. 101202457). Views and opinions expressed are those of the author(s) only and do not necessarily reflect those of the European Union. Neither the European Union nor the granting authority can be held responsible for them.

\subsection*{AI use statement}

In this work, we used generative AI tools for language polishing and literature retrieval and discovery. We have not used generative AI tools for research ideation, manuscript drafting, or proving mathematical claims. Additionally, we used Qwen2.5-VL-7B-Instruct for carrier quality checking as part of our experimental pipeline. We have reviewed all AI-assisted work. All AI-assisted language edits were reviewed by the authors, references identified with AI assistance were manually verified against the original sources, and Qwen2.5-VL-7B-Instruct outputs were used only within the specified quality gate control procedure as discussed in~\ref{app:SAM3_gates}. We take responsibility for the final content of this work, including text, claims or artifacts produced with the aid of generative AI.

\subsection*{Ethics statement}

This work studies adversarial attacks on image classifiers and therefore has potential dual-use implications. Although the proposed method could be misused to evade image-classification systems, our purpose is to expose weaknesses in current models and support the development of more robust and reliable recognition systems. We evaluate the method under a clearly specified experimental setting and do not deploy it against real-world services or systems.

Our human test evaluation involved only perceptual judgments of generated images and did not require participants to provide sensitive personal information. Participants were informed about the study procedures, the use of their responses, the voluntary nature of participation, and any applicable compensation before providing consent. No identifying information is reported in this paper or released with the study results.

\subsection*{Reproducibility statement}

We provide the information required to reproduce our construction, attack, and evaluation pipelines throughout the main paper and appendices. The method and experimental protocol are described in Sections~\ref{sec:method} and~\ref{sec:experiment}. Appendices~\ref{app:experiment_set_up},~\ref{app:SAM3_gates}, and~\ref{app:baseline_realization} provide the complete experimental settings, model and implementation details, spatial construction procedure, quality-control gates, and attack configuration. Reproducibility details include model versions, source-target pairings, prompts, random seeds, attack hyperparameters, evaluation criteria, and data-processing procedures. An anonymized implementation and the associated experimental configurations are provided at \url{https://github.com/DavidJlf/carrier-attack}.


\bibliography{iclr2027_conference}
\bibliographystyle{iclr2027_conference}

\appendix
\clearpage
\section*{Appendix}

\section{Experimental Setup and Reproducibility}
\label{app:experiment_set_up}

\subsection{Data and Experimental Design}

We use the personalized subjects from the DreamBooth reference image dataset by ~\citet{dreambooth} and 30 ImageNet-1K target classes, producing $20\times30=600$ source--target pairs to evaluate our framework. The personalized subject refers to specific subject rather than an entire semantic category, and each subject is associated with a subject-specific FLUX LoRA, and one source image is randomly selected from the LoRA-generated imgae pool for the subsequent experiments.

The subjects are backpack, bear plushie, candle, cat, cat 2, dog, dog 2, dog 3, dog 5, dog 7, duck toy, fancy boot, grey sloth plushie, pink sunglasses, poop emoji, RC car, robot toy, shiny sneaker, teapot, and vase. Numerical suffixes distinguish dataset instances.

Therefore, our experiment contains \(20\times30=600\) source–target pairs.Non-Target, Hybrid, and Target Carrier conditions using CRA, CIRA, and JIA, resulting in 1,800 carrier-controlled cases and 5,400 attacked images.  No-Carrier build upon concept-based attack~\citet{conceptbased} and NatADiff by~\citet{natadiff}
are evaluated separately on the same 600 source--target pairs. The 5,400 carrier-based
attacked images are additionally used in the human evaluation.

The target classes are listed in Table~\ref{tab:target_classes} using zero-based ImageNet-1K IDs.

\begin{table}[H]
\centering
\caption{ImageNet-1K target classes used in our experiments.}
\label{tab:target_classes}
\begin{tabular}{r l r l r l}
\toprule
\textbf{Target ID} & \textbf{Target} &
\textbf{Target ID} & \textbf{Target} &
\textbf{Target ID} & \textbf{Target} \\
\midrule
94  & hummingbird        & 113 & snail               & 272 & coyote \\
277 & red fox            & 301 & ladybug             & 331 & hare \\
340 & zebra              & 344 & hippopotamus        & 347 & bison \\
348 & ram                & 354 & Arabian camel       & 386 & African elephant \\
404 & airliner           & 407 & ambulance           & 483 & castle \\
487 & cellular telephone & 504 & coffee mug          & 555 & fire engine \\
562 & fountain           & 580 & greenhouse          & 626 & lighter \\
644 & matchstick         & 695 & padlock             & 709 & pencil box \\
717 & pickup             & 722 & ping-pong ball      & 761 & remote control \\
779 & school bus         & 852 & tennis ball         & 902 & whistle \\
\bottomrule
\end{tabular}
\end{table}

\subsection{Target–Carrier Mapping}
We realize our framework under three different carrier conditions, including the Non-Target Carrier, Hybrid, and Target Carrier condition. The carrier mapping and selection is guided by compatibility in shape, coarse structure, or function, while seeking to avoid the target itself, synonymous categories, and difficult-to-distinguish fine-grained neighbors. This aims to provide correlation for generating the visually secondary carrier and target and retaining different visible category identities. For example, ambulance is mapping to minivan, and zebra to sorrel. We avoid excessively close target–carrier pairs, especially fine-grained sibling classes such as Pembroke Welsh corgi (ID 263) versus Cardigan Welsh corgi (ID 264), Such pairs would blur the semantic distinction between a Non-Target Carrier and the target itself. The pairs we choose should share general structure, whereas target-related markings and stripes are not directly specified by the base carrier category.

The mapping is a construction heuristic and does not guarantee an optimal carrier for attack performance. Qwen VLM model by~\citet{qewen} subsequently participates in candidate verification and description refinement, adjusting the carrier’s visual presentation and generation requirements. This process is distinguished from the initial category mapping. Detailed about Qwen usage  can be found in Appendix~\ref{gate_check}.

Non-Target conction uses the mapped non-target category. Hybrid Carrier retains the same base category and requeststarget-related attributes through additional prompt descriptions. Target Carrier directly instantiates the target category itself. The attack target \(c_t\) remains unchanged across all three conditions.

Table~\ref{tab:target_carrier_mapping} lists mappings for 10 representative targets among the 30 used in the formal experiments. Multiple targets can share a carrier category; the mapping is therefore not required to be one-to-one.

\begin{table}[H]
\centering
\caption{Target-to-base-carrier mappings used in the formal experiments.}
\label{tab:target_carrier_mapping}
\begin{tabular}{r l r l}
\toprule
\textbf{Target ID} & \textbf{Target} & \textbf{Carrier ID} & \textbf{Base Carrier} \\
\midrule
113 & snail              & 390 & eel \\
340 & zebra              & 339 & sorrel \\
386 & African elephant   & 344 & hippopotamus \\
404 & airliner           & 417 & balloon \\
407 & ambulance          & 656 & minivan \\
487 & cellular telephone & 848 & tape player \\
555 & fire engine        & 656 & minivan \\
562 & fountain           & 915 & yurt \\
580 & greenhouse         & 832 & stupa \\
722 & ping-pong ball     & 417 & balloon \\
\bottomrule
\end{tabular}
\end{table}

\subsection{Personalized Sources and LoRA}

We use subject-specific LoRA to expand reference images provided by Dreambooth into personalized candidates with variations in pose, viewpoint, and scene layout~\citep{lora}. Our LoRA training configuration uses BF16 computation, a resolution of \(1024\times1024\), LoRA rank and alpha of 16, a learning rate of \(10^{-4}\), and 1,250 optimization steps. The recorded pipeline uses four or five available reference images from Dreambooth per subject and generates an initial pool of 200 candidates using distinct prompts and seeds. Quality filtering can reduce the retained pool. These candidates are source-preparation assets rather than independent attack trials.

Among the 200 candidates, we randomly select the images that preserve the primary subject but we exclude images with visible artifacts such as ghosting, duplication, or other structural distortions. The corresponding LoRA scale is 1.0 for carrier-background generation and finite return. Clean background inpainting is performed without LoRA, whereas the subsequent attack stage loads the corresponding subject-specific adapter. Specifically, CIRA loads LoRA for finite return after completing clean inpainting, while JIA loads LoRA for its attacked inpainting trajectory. The route-specific clean and
attacked constructions are detailed in Appendix~\ref{app:algorithm}.

\subsection{Models and Attack Configuration}

We use FLUX.1-Kontext-dev by~\citet{flux} for image generation, ResNet-50 as the WhiteBox classifier, SAM3 by~\citet{sam3} for primary subject segmentation, and Qwen2.5-VL-7B-Instruct by~\citet{qewen} for carrier-candidate verification and feedback-based prompt refinement. Qwen operates during construction and quality check but not participate in the gradient computation of the classifier loss.

The standard classifier scale is \(\lambda=0.5\) in main attack setting. The attack minimizes targeted cross-entropy and applies global RMS normalization to the gradient~\citep{RMS}. The source-class suppression coefficient is zero, and no identity or DINO loss is included in the attack objective.

For all three attack routes, classifier guidance is applied over the final 15 intervals of a 50-step scheduler. 

For CRA, we first generate the visually secondary carrier background using 50 generation steps and FLUX guidance 3.5. Then the primary subject is composited onto the visually secondary carrier background. Starting from scheduler state 50, CRA performs partial inversion over the final 15 intervals to state 35, followed by guided return over the same intervals from state 35 back to state 50. We therefore denote its attack trajectory as \(50\rightarrow35\rightarrow50.\) During the guided return, RMS-normalized classifier gradient is weighted by \(\lambda\sigma_k\), where \(\sigma_k\) is the scheduler-dependent noise level.

For CIRA, we first perform a complete 50-step clean conditional inpainting without classifier guidance to generate the clean image containing the visually secondary carrier. Then, CIRA subsequently follows the same \(50\rightarrow35\rightarrow50\) schedule as CRA. Its classifier gradient contribution is also weighted by \(\lambda\sigma_k\). The preceding 50-step clean inpainting is a separate construction stage and is not part of the finite-return inversion.

JIA does not perform the finite-return inversion. The classifier guidance is directly introduced into the 50-step inpainting trajectory. The first 35 intervals proceed without classifier guidance, after which guidance is activated for the final 15 intervals. 

JIA uses the dimensions determined by the Kontext image-processing pipeline, and the exact width and height recorded for each output. Unlike CRA and CIRA, JIA uses a constant velocity attack coefficient \(\lambda\) over the guided intervals. The globally guided update is subsequently followed by the native inpainting mask-blending operation.

Moreover, in the primary subject preservation experiment under strong adversarial attacks, we increase the classifier scale from \(\lambda=0.5\) to \(\lambda=5\) while retaining the same 15-interval finite-return schedule.  The corresponding
NatADiff and No-Carrier settings and its method-specific time-travel budget are reported separately
in Appendix~\ref{app:baseline_realization}.

\subsection{Conditional Attack Success}
\label{app:conditional_asr}

To avoid the carrier influence the target prediction. we compute
conditional Top-5 ASR only on samples for which the target class is absent
from the clean ResNet-50 Top-5 predictions. We use Top-5 rather than Top-1
because some personalized DreamBooth subjects do not have an exact
corresponding class in ImageNet-1K, making a single-class prediction less
representative of the image content. Specifically,

\[
\mathrm{ASR}_{\mathrm{cond}}^{@5}
=
\frac{
\#\{i:\operatorname{rank}_{\mathrm{clean}}(c_t)>5
\land
\operatorname{rank}_{\mathrm{attack}}(c_t)\leq5\}
}{
\#\{i:\operatorname{rank}_{\mathrm{clean}}(c_t)>5\}
}.
\]

We additionally report whether the predefined subject category remains in the clean ResNet-50 Top-5 predictions.

Table~\ref{tab:conditional_asr} reports the number of eligible samples
after filtering and the corresponding conditional Top-5 ASR.

\begin{table}[H]
\centering
\caption{Clean-image WhiteBox prediction audit and conditional Top-5 ASR.
Clean Subject Top-5 reports whether the predefined subject category appears
in the clean ResNet-50 Top-5 predictions. Eligible denotes the number of
samples, out of 600 per setting, for which the target is absent from the
clean Top-5 predictions.}
\label{tab:conditional_asr}

\small
\setlength{\tabcolsep}{4pt}
\renewcommand{\arraystretch}{1.05}

\begin{tabular}{@{}llccc@{}}
\toprule
Method & Carrier & Clean Subject Top-5 (\%) & Eligible & Cond. Top-5 ASR $\uparrow$ \\
\midrule
CRA  & Non-Target & 82.35 & 589/600 & 99.49 \\
CRA  & Hybrid     & 76.86 & 562/600 & 99.64 \\
CRA  & Target     & 82.35 & 450/600 & 99.56 \\
\addlinespace[2pt]
CIRA & Non-Target & 89.41 & 593/600 & 98.48 \\
CIRA & Hybrid     & 90.78 & 595/600 & 98.82 \\
CIRA & Target     & 89.22 & 322/600 & 99.69 \\
\addlinespace[2pt]
JIA  & Non-Target & 89.80 & 594/600 & 55.39 \\
JIA  & Hybrid     & 90.20 & 597/600 & 55.28 \\
JIA  & Target     & 89.61 & 320/600 & 82.50 \\
\bottomrule
\end{tabular}
\end{table}

The consistently high Clean Subject Top-5 rates show that the clean carrier constructions generally retain WhiteBox recognizable evidence for the original subject category before adversarial guidance.

\section{Spatial Construction, SAM3, and Quality Gates}
\label{app:SAM3_gates}

\subsection{SAM3 Segmentation and Spatial Compositing}

SAM3 can predict the primary subject mask with the subject-specific text prompt that saved during LoRA construction, with a confidence threshold of 0.5~\citep{sam3}.During construction, a single subject is the ideal situation, while failure to detect a mask will be recorded as a construction error and multiple subjects appearing will also be marked as errors, and new images need to be selected. The SAM3 usage during the construction is distinct from the preservation evaluation, where multiple detected instances are merged to form the evaluation mask.

Composite construction resizes the source RGB image using Lanczos interpolation and the subject mask using nearest neighbors, followed by thresholding at 128 in its 8-bit representation. A Gaussian feather with radius 4 pixels produces the blending mask \(\widetilde m_s\). The soft blending mask and binary subject mask are stored separately.

\subsection{Clean Background Inpainting}
Inpainting construction uses the source image as its image condition and \(m_b\) that satisfy $m_b=1-m_s$  as the editable region, without loading LoRA. The prompt requests one visually secondary carrier with visible edges and a slightly blurred and defocused appearance, while retaining the foreground primary subject and avoiding a duplicate foreground instance.
A representative template is:

\begin{quote}
\small
Edit only the masked background. Preserve the existing foreground [subject] unchanged.
In the distant background, exactly one [carrier description], slightly blurred by shallow depth of field, with visible edges. Do not add or modify any [subject]. 
\end{quote}

This template specifies the construction request while the generated result should still be verified by subsequent quality gate check.

\subsection{Qwen-aided Carrier quality gate check}
\label{gate_check}

Carrier prompts only are not always faithfully realized. Because CRA directly composites the primary subject onto the generated background, while the inpainting relatively limits generation flexibility, the prompt constraints may not always be fully realized. As a result, carriers produced by the inpainting-based methods may occasionally deviate from the intended design. Therefore, we combine classifier check and Qwen visual feedback to form carrier quality gate to verify whether the requested carrier is generated correctly and to refine its description when needed. Qwen is only used for gate check during the construction. 

We follows the classifier-first policy. If the ResNet-50 ranks the carrier category within Top 10 list, the candidate is accepted without the Qwen quality check. Otherwise, the Qwen is run to reevaluate the acceptance of the candidates, failure explanations, and a revised carrier phrase or background prompt. The inspection considers whether the visually secondary carrier is present, whether it aligned with the desired category, and whether severe structural defects are apparent. The Qwen feedback only refines the next generation prompt. 

For CRA, we use the independent carrier background before compositing to check the carrier quality, reducing interference from the inserted subject. For clean inpainting construction, SAM3 localize the primary subject mask again with dilation to get background-only classifier view. We then apply the gate check the quality of the carrier construction. And the background-only-view is only for construction diagnostics and does not replace the clean image passed to the attack.

\subsection{Bounded Retries and Best-Candidate Selection}

Retries after the failure judgment result in a significant construction time increase, and For some potentially ambiguous categories, the classifier and the FLUX generator exhibit semantic mismatches. For example, for the class *teddy*, the classifier interprets it as a teddy bear plushie, whereas the generator may interpret it as a brown bear or a teddy dog. Therefore, we bounded retries times to avoid infinite regeneration.

Each round first keeps the current prompt fixed and generates multiple candidates by varying the seed for three times. Only if all candidates failed to pass the quality check gate, prompt is revised according to Qwen feedback. By default, at most three prompt rounds are allowed. Each construction attempt is evaluated using the background-only diagnostic view.
A failed candidate triggers a new seed, while Qwen feedback may additionally revise
the prompt within a bounded feedback budget. Once the Qwen budget is exhausted,
remaining attempts vary only the seed. If no candidate passes the gate, the candidate
with the best classifier rank for the requested carrier category is retained as a
forced fallback, and its failure status is recorded rather than treated as a successful
construction.

\begin{figure}[t]
    \centering
    \includegraphics[width=\linewidth]{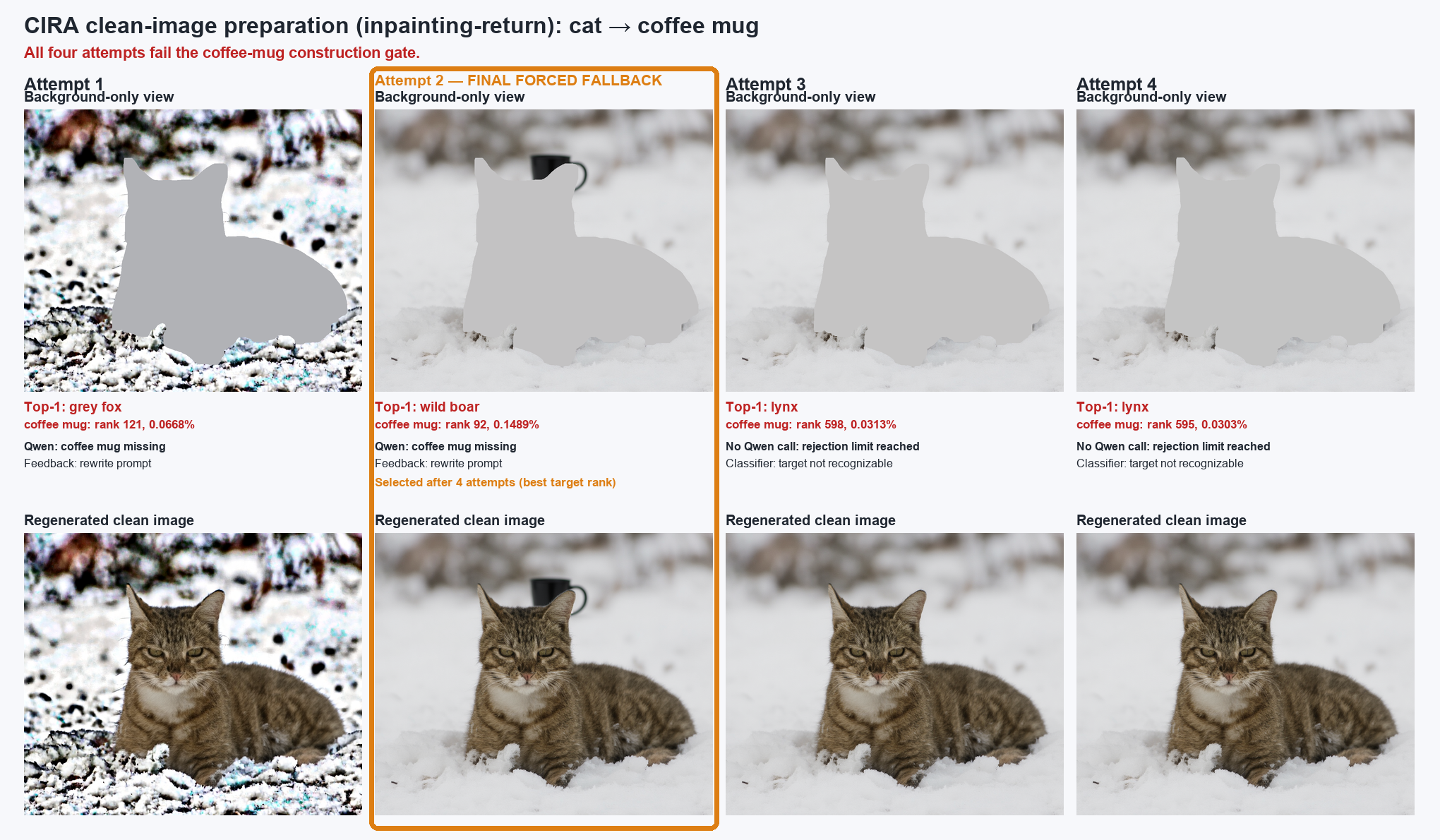}
    \caption{Bounded carrier-construction retries for CIRA. Classifier and Qwen feedback
    identify the missing coffee-mug carrier, after which the prompt and generation seed
    are varied. None of the four attempts passes the gate; Attempt 2 is retained as the
    forced fallback based on its best carrier-category rank.}
    \label{fig:gate_retry}
\end{figure}

Figure~\ref{fig:gate_retry} illustrates this process for cat $\rightarrow$ coffee mug.
All four candidates fail to produce a reliably recognizable coffee mug. Qwen identifies
the missing carrier and revises the prompt during the first two attempts; the remaining
attempts vary the seed after the feedback limit is reached. Attempt 2 is finally retained
as the forced fallback because it obtains the best coffee-mug rank, although it still
fails the construction gate.

\subsection{Incomplete Realization of Hybrid Attributes}
\label{hybrid_incomplete}
Hybrid Carrier construction should retain the majority of the Non-Target Carrier category while partially expressing the target-related characteristics. The former Non-Target mainly controls the category and overall structure, and the target-related characteristics mainly control finer details such as color, material, stripes, ornaments, or surface patterns. Thus, only being classified as the correct category does not mean a good realization of Hybrid Carrier.

\begin{figure}[H]
    \centering
    \includegraphics[width=\linewidth]{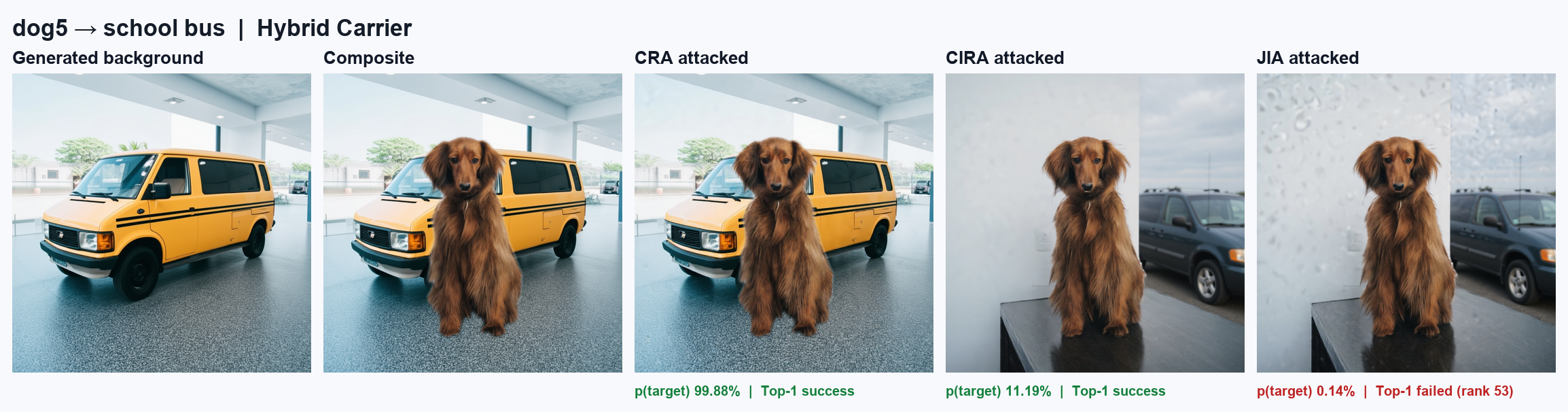}
    \caption{
    Qualitative examples under the Hybrid Carrier condition for
    \textit{dog5} targeting ''school bus''. The first two columns show
    the generated carrier background and the composite input, followed
    by the outputs of CRA, CIRA, and JIA. Green indicates Top-1
    attack success, while red indicates failure.
    }
    \label{fig:hybrid_dog5_schoolbus}
\end{figure}

In the composite construction, this issue is less common in the composite construction, construction method in CRA, because the carrier background is generated as a complete image, providing the generative model with substantially greater spatial and semantic flexibility. As a result, FLUX can usually satisfy the prompt constraints more faithfully. For example, in the Figure~\ref{fig:hybrid_dog5_schoolbus}, the generated Hybrid Carrier retains the overall identity of a non-target vehicle while incorporating target-related attributes of a school bus, such as its yellow appearance and bus-like structure.

\begin{figure}[H]
    \centering
    \includegraphics[width=\linewidth]{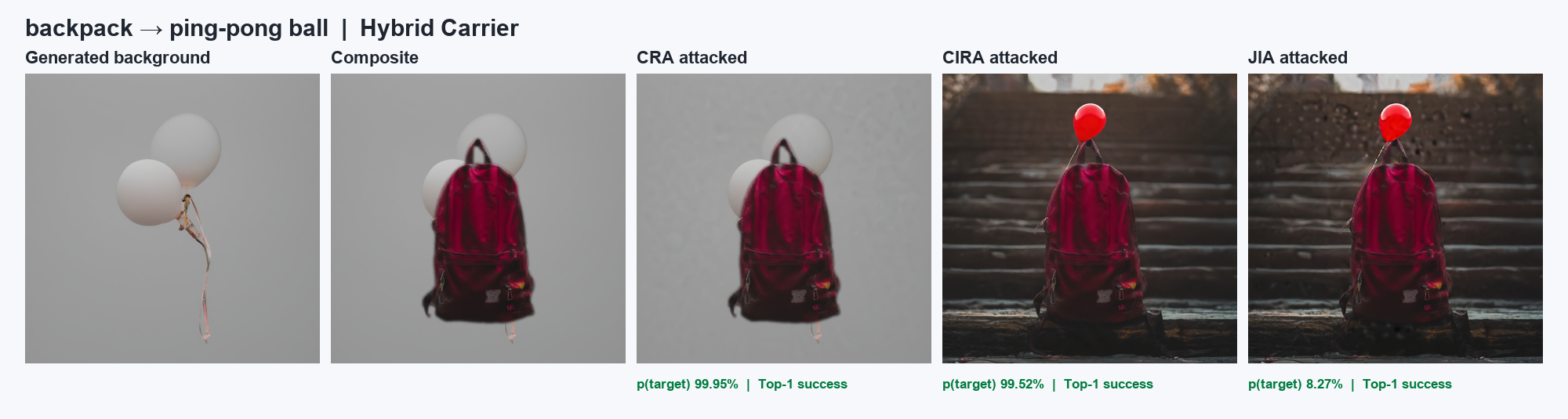}
    \caption{
    Qualitative examples under the Hybrid Carrier condition for
    \textit{backpack} targeting ''ping-pong ball''. The first two columns show
    the generated carrier background and the composite input, followed
    by the outputs of CRA, CIRA, and JIA. Green indicates Top-1
    attack success.
    }
    \label{fig:hybrid_backpack_pingpong}
\end{figure}

While in inpainting construction, FLUX retain the base carrier category and silhouette but insufficiently expresses the requested target attributes. As shown in Figure~\ref{fig:hybrid_dog5_schoolbus}, the characteristics of the target are failed to expressed to the visually secondary carrier under the prompt. And in Figure~\ref{fig:hybrid_backpack_pingpong}, the visually secondary carrier expresses the target ping-pong-ball characteristics by replacing the balloon body, while retaining the original balloon strings and keeping the personalized backpack as the primary image content.

These situations can arise because the characteristics are not visibly generated, or because a small, distant, or blurred carrier makes generated attributes difficult to recognize, and we collectively describe them as insufficient visible realization of target characteristics. Originally the Hybrid carrier ideally have higher target characteristics strength than the Non-Target Carrier, causing better transferability compare to the Non-Target. In practice, however, we observe the opposite trend: the improvement in transferability from Non-Target to Hybrid is relatively limited. This does not contradict our claim regarding transferability in Section~\ref{transferability}, as the Hybrid condition often exhibits insufficient visible realization of target-related characteristics, thereby limiting the expected transferability gain.

\section{Detailed Algorithms for CRA, CIRA, and JIA}
\label{app:algorithm}

Let \(\mathcal B(x,m_b,p_c;\xi)\) denote clean background inpainting with carrier prompt \(p_c\) and seed \(\xi\), and let \(\mathcal R(x_0,c_t;h)\) denote the classifier-guided finite-return operator defined in Appendix~\ref{app:inversion_return}. The condition \(h\) contains the image and text conditioning used by each route. We use \(\mathcal V\) to denote the bounded carrier-verification and refinement procedure in Appendix~\ref{app:SAM3_gates}.

\subsection{Composite Reconstruction Attack(CRA)}

CRA first generates a carrier background independently and apply carrier quality gate check, and then composites the segmented source primary subject onto it to get the clean image. The partial inversion and guided return should be applied to the clean image.

\[
x_{\mathrm{adv}}^{\mathrm{CRA}}
=
\mathcal{R}(x_{\mathrm{clean}}^{\mathrm{comp}}, c_t; h_{\mathrm{comp}}).
\]

Here, $\mathcal{R}$ denotes the partial-inversion and finite-return operator with
globally RMS-normalized classifier guidance.

\begin{enumerate}
    \item Load the source image and its subject-specific LoRA, and obtain the primary subject mask.

    \item Generate the initial carrier request and independent background from the target--carrier mapping.

    \item Apply $\mathcal{V}$: verify the carrier, refine the description using Qwen feedback when needed, and retain a candidate upon acceptance or budget exhaustion.

    \item Use the composite associated with the selected background and retain its prompt, seed, and gate status.

    \item Encode the composite and invert it along the low-noise suffix of the FLUX grid.

    \item Apply global targeted guidance along the same suffix, then decode and evaluate the final image.
\end{enumerate}

\begin{figure}[t]
    \centering
    \includegraphics[width=\linewidth]{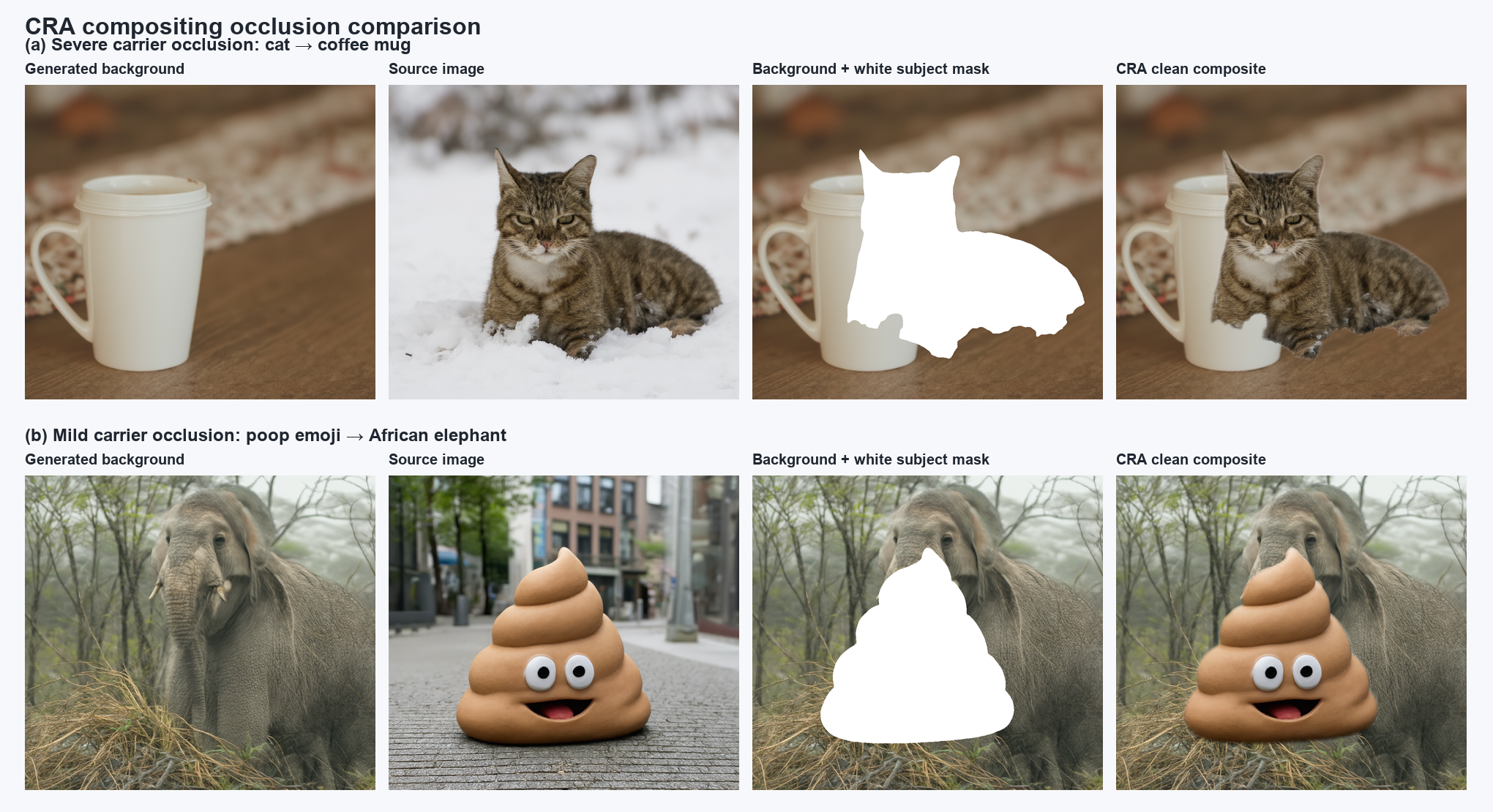}
    \caption{
    Carrier occlusion during CRA compositing. Each row shows the generated carrier
    background, source image, subject insertion mask, and resulting clean composite.
    Top: severe occlusion, where the cat substantially overlaps the coffee-mug carrier.
    Bottom: mild occlusion, where most of the elephant carrier remains visible after
    insertion of the poop-emoji subject.
    }
    \label{fig:cra_occlusion}
\end{figure}

Because subject compositing occurs after background generation, the inserted subject
may partially or completely occlude the visible carrier. Figure~\ref{fig:cra_occlusion} illustrates two representative cases.
In the severe case, the cat overlaps substantially with the coffee-mug carrier,
whereas the smaller poop-emoji subject leaves most of the elephant carrier visible.
Thus, passing the background-level carrier check does not necessarily guarantee
that the carrier remains equally visible in the final clean composite.

\subsection{Clean Inpainting Reconstruction Attack(CIRA)}

CIRA finishes the clean inpainting before adversarial reconstruction. It first constructs the visually secondary carrier in the background without LoRA, and then the image is fixed and the LoRA is loaded for the later image finite-return attack.
\[
x_{\mathrm{clean}}^{\mathrm{inp}}
=
\mathcal{B}(x,m_b,p_c;\xi),
\qquad
x_{\mathrm{adv}}^{\mathrm{CIRA}}
=
\mathcal{R}\!\left(
x_{\mathrm{clean}}^{\mathrm{inp}},
c_t;
h_{\mathrm{inp}}
\right).
\]

where $\mathcal{R}_{\mathrm{RMS}}$ denotes the partial-inversion and
finite-return operator with globally RMS-normalized classifier guidance.

\begin{enumerate}
    \item Prepare the source image, primary subject mask, and carrier request.
    \item Generate a 50-step clean inpainting without LoRA.
    \item Apply $\mathcal{V}$ to verify the primary subject and visually secondary carrier and perform bounded feedback-based retries.
    \item Fix the selected clean image, prompt, seed, and gate status.
    \item Load the corresponding subject-specific LoRA and perform partial inversion and global guided return.
    \item Decode and evaluate the final attacked image.
\end{enumerate}

\subsection{Joint Inpainting Attack(JIA)}

JIA directly introduce the attack during the background inpainting trajectory. A separate clean candidate is first generated with LoRA, and \(\mathcal V\) selects the prompt and seed. The LoRA is also loaded for attacked inpainting from the source image condition, while the clean image is  not inverted as in CIRA.  And unlike CRA and CIRA, whose RMS-normalized guidance is encapsulated in the finite-return operator \(\mathcal R\), JIA injects the normalized classifier gradient directly into the inpainting velocity:

\[
v_k^{\mathrm{JIA}}
=
v_k
+
\mathbf{1}[k\in\mathcal A]\lambda
\operatorname{Norm}_{\mathrm{RMS}}(g_k),
\qquad
\mathcal A=\{35,\ldots,49\}.
\]

\[
\begin{aligned}
z'_{k+1}
&=
\operatorname{Step}(z_k,v_k^{\mathrm{JIA}},k),\\
z_{k+1}
&=
(1-M_b)\odot q_{k+1}^{\mathrm{src}}
+
M_b\odot z'_{k+1}.
\end{aligned}
\]

Here, \(q_{k+1}^{\mathrm{src}}\) is the source latent re-noised to the next noise level using the same noise realization; the final step uses the clean source latent. Global gradient injection and subsequent inpainting blending are distinct operations, and this blend does not impose an exact identity constraint on the final decoded subject pixels.
JIA performs 50 attacked inpainting steps, activates classifier guidance in the final 15 intervals, and decodes the terminal latent.
The native inpainting blend tends to constrain visible modifications to the editable
background, although it does not impose exact foreground invariance.

\subsection{No Ca Baseline}
Build upon Concept-based attack by~\citet{conceptbased}, the No Carrier condition applies \(\mathcal R\) directly to the personalized source image using its LoRA. It does not generate a carrier background or evaluated by the carrier gate check. Classifier guidance is global and uses the same targeted objective and RMS convention as CRA and CIRA. The complete baseline configuration and its relation to
NatADiff are provided in Appendix~\ref{app:baseline_realization}.

Qualitative analysis of CRA, CIRA, and JIA are provided in Appendix~\ref{qualitative_analysis}.
\section{FLUX Inversion and Classifier-Guided Finite Return}
\label{app:inversion_return}
\subsection{Clean Image Prediction from Intermediate State}
\label{D.1}
For CRA and CIRA, the attack start from a constructed clean image \(x_{\mathrm{clean}}\). Let \(E_{\mathrm{VAE}}\) and \(D_{\mathrm{VAE}}\) represent the VAE encoder and decoder, and \(z_0=E(x_{\mathrm{clean}})\). The linear FLUX path can be expressed as: 
\[
z_\sigma = (1-\sigma)z_0 + \sigma \epsilon,
\qquad
\epsilon \sim \mathcal{N}(0,I),
\]

\(\sigma\) is the noise level and zero refers to the clean endpoint. Path velocity is \(u=\epsilon-z_0\), giving  \(z_0=z_\sigma-\sigma u\). Let \(v_\theta\) represents the predicted velocity yields the clean-image estimate \(\widehat x_0\), where \(c\) denotes fixed generation conditions, \(y_{\mathrm{tar}}\) is the target class, and \(p_f\) denotes classifier probabilities including differentiable preprocessing. Gradients propagate through the decoder to the latent, but not through the FLUX velocity network. Therefore, we can get the targeted classification loss:
\[
\begin{aligned}
\hat{x}_0
&=
D\!\left(
z_\sigma
-
\sigma v_\theta(z_\sigma,\sigma,c)
\right),\\[4pt]
\mathcal{L}_{\mathrm{tar}}
&=
-\log p_f\!\left(
y_{\mathrm{tar}}
\mid
\hat{x}_0
\right),
\qquad
g_\sigma
=
\nabla_{z_\sigma}\mathcal{L}_{\mathrm{tar}}.
\end{aligned}
\]

\subsection{Score–Velocity Correspondence}
The linear path in Appendix~\ref{D.1} defines the conditional
distribution of the noisy latent $z_\sigma$ given the clean latent $z_0$:
\[
q_\sigma(z_\sigma \mid z_0)
=
\mathcal{N}\!\left(
(1-\sigma)z_0,\,
\sigma^2 I
\right).
\]

Let \(p_\sigma(z_\sigma\mid c)\) denote the corresponding marginal density and the score is the log-density gradient with respect to \(z_\sigma\), we can get:

\[
\begin{aligned}
s_\sigma
&:=
\nabla_{z_\sigma}\log p_\sigma(z_\sigma \mid c)\\
&=
-\frac{
z_\sigma-(1-\sigma)\mathbb{E}[z_0 \mid z_\sigma,c]
}{
\sigma^2
}.
\end{aligned}
\]

therefore, we can have conditional mean velocity:
\[
s_\sigma
=
-\frac{
z_\sigma + (1-\sigma)v_\sigma
}{
\sigma
},
\qquad
v_\sigma
=
-\frac{
z_\sigma + \sigma s_\sigma
}{
1-\sigma
},
\qquad
0<\sigma<1.
\]

\subsection{Classifier-Induced Velocity Modification}

we reweight the generative density at a fixed noise level to favor states assigned higher target-class probabilities by the classifier:

\[
\tilde{p}_\sigma(z_\sigma \mid c, y_{\mathrm{tar}})
\propto
p_\sigma(z_\sigma \mid c)
\exp\!\left[
-\gamma \mathcal{L}_{\mathrm{tar}}(z_\sigma)
\right],
\qquad
\gamma \ge 0.
\]
Here, \(\gamma\) controls guidance strength. Taking the log-density gradient and using \(g_\sigma=\nabla_{z_\sigma}\mathcal L_{\mathrm{tar}}\) gives
\[
\tilde{s}_\sigma
=
s_\sigma
-
\gamma g_\sigma.
\]

According to the score-velocity relationship, we get the corresponding classifier-guided velocity:
\[
\begin{aligned}
\tilde{v}_\sigma
&=
-\frac{
z_\sigma + \sigma \tilde{s}_\sigma
}{
1-\sigma
}\\
&=
v_\sigma
+
\gamma
\frac{\sigma}{1-\sigma}
g_\sigma.
\end{aligned}
\]

Thus, classifier feedback can be introduced as an additive gradient term in the FLUX velocity. This relation provides a fixed-noise-level interpretation of guidance. While the practical return uses the normalized update discussed in Appendix~\ref{normalize_return}.

\subsection{Partial Inversion and Pivot Correction}
\label{flux_inversion}
To attack the clean input image \(x_{\mathrm{clean}}\), inspired by the DDIM inversion from~\citet{aca,diffattack}, we first partially invert its latent toward increasing noise and then return along the same intervals. Let the return noise schedule be \(\sigma_0>\cdots>\sigma_K=0\). The saved inversion states, or pivots refereed by~\citet{ddim_inversion}, ordered for return, are denoted by \(p^{(0)},\ldots,p^{(K)}\) and called pivots. Starting from \(p^{(K)}=z_0\), the remaining states are computed for \(k=K-1,\ldots,0\):
\[
p^{(k)}
=
p^{(k+1)}
+
\left(\sigma_k-\sigma_{k+1}\right)
v_\theta\!\left(
p^{(k+1)},
\sigma_{k+1},
c
\right).
\]

Inversion does not include classifier attack guidance. Return starts from \(z^{(0)}=p^{(0)}\), where \(z^{(k)}\) denotes the actual latent at step \(k\). To make the base update follow the saved reconstruction trajectory, we correct the model velocity using the finite-difference velocity between adjacent pivots:
\[
v_{\mathrm{pivot},k}
=
\frac{
p^{(k+1)}-p^{(k)}
}{
\sigma_{k+1}-\sigma_k
},
\]

\[
v_{\mathrm{base},k}
=
(1-\beta)\,
v_\theta\!\left(
z^{(k)},
\sigma_k,
c
\right)
+
\beta\,v_{\mathrm{pivot},k}.
\]

\subsection{Practical Gradient and Return Update}
\label{normalize_return}
Our return uses the clean image prediction and classification loss in Appendix~\ref{D.1}, but replacing the velocity with the base velocity in Appendix~\ref{flux_inversion}.
\[
\hat{x}_0^{(k)}
=
D\!\left(
z^{(k)}
-
\sigma_k\,\operatorname{stopgrad}\!\left(v_{\mathrm{base},k}\right)
\right),
\]

\[
g_k
=
\nabla_{z^{(k)}}
\left[
-\log p_f\!\left(
y_{\mathrm{tar}}
\mid
\hat{x}_0^{(k)}
\right)
\right].
\]

For controlling gradient scales, we apply RMS normalization to the complete latent gradient~\citep{RMS}. Here, \(n\) is the number of gradient elements and \(\delta>0\), and let \(\mathcal A\) denote the attack window and \(\lambda\) the guidance strength.
\[
\operatorname{Norm}_{\mathrm{RMS}}(g)
=
\frac{g}{
\max\!\left(
\sqrt{\frac{1}{n}\sum_{i=1}^{n} g_i^2},
\delta
\right)
}.
\]
The resulting velocity and Euler return update are:
\[
v_k^{\mathrm{adv}}
=
v_{\mathrm{base},k}
+
\mathbf{1}[k\in\mathcal{A}]\,\lambda\sigma_k
\operatorname{Norm}_{\mathrm{RMS}}(g_k),
\]

\[
z^{(k+1)}
=
z^{(k)}
+
(\sigma_{k+1}-\sigma_k)\,v_k^{\mathrm{adv}}.
\]

The final output is \(x_{\mathrm{adv}}=D(z^{(K)})\). This attack stage uses the subject-specific LoRA and global classifier gradients; subject and background masks from construction do not restrict return updates.

\section{Carrier-based Normalized Update Allocation}
\label{app:attack_update_allocation}

This appendix provides the complete derivation of
Theorem~\ref{thm:subject_update}.
Inspired by the regional evidence decomposition of~\citet{doublemainbody}, we
partition the classifier gradient according to spatial image regions. Their analysis
motivates the regional decomposition; the comparison under global RMS normalization
developed below is specific to our carrier-based setting. We first contrast a
No-Carrier image containing primary subject and remaining regions with a carrier-based image
containing primary subject, visually secondary carrier, and remaining regions. We then derive a conditional
comparison of their normalized subject updates. This comparison isolates the normalization effect and should not be interpreted as asserting that independently constructed Carrier and No-Carrier images share identical raw regional gradients.

\subsection{Regional Decomposition}

Following Section~\ref{sec:formulation}, let $m_S$ and $m_C$ denote the binary
primary-subject and carrier masks, respectively, and define the remaining-region
mask as
\[
m_R=1-m_S-m_C.
\]
Because the primary-subject and carrier regions are spatially separated,
\[
m_S\odot m_C=0.
\]
The three corresponding image-region components are
\[
x_S=m_S\odot x,
\qquad
x_C=m_C\odot x,
\qquad
x_R=m_R\odot x.
\]

A No-Carrier image contains the primary-subject and remaining-region components:
\[
x^{\mathrm{NC}}
=
x_S+x_R.
\]
A carrier-based image additionally contains the visible carrier component:
\[
x^{\mathrm C}
=
x_S+x_C+x_R.
\]
These expressions define a spatial partition of the image and do not assume that
the semantic contents of the three regions are statistically independent.

For the targeted loss
\[
\mathcal L_t(x)
=
-\log p(t\mid x),
\]
consider the latent classifier gradient $g_t$ defined in
Section~\ref{sec:attack}. Let $\Pi_S$, $\Pi_C$, and $\Pi_R$ denote the disjoint
coordinate projections associated with the primary-subject, carrier, and remaining
regions. We define the corresponding regional components as
\[
g_{S,t}=\Pi_S(g_t),
\qquad
g_{C,t}=\Pi_C(g_t),
\qquad
g_{R,t}=\Pi_R(g_t).
\]
Therefore,
\[
\begin{aligned}
\text{No-Carrier:}\qquad
&g_t^{\mathrm{NC}}
=
g_{S,t}+g_{R,t},\\
\text{Carrier:}\qquad
&g_t^{\mathrm C}
=
g_{S,t}+g_{C,t}+g_{R,t}.
\end{aligned}
\]
The projections are introduced only to analyze the regional allocation of the
gradient. The implemented attack applies the complete global gradient $g_t$ without
regional masking.

\subsection{Regional Target Gradients}
We first consider that the decomposition ideally: Both the subject and carrier are class-related. A region is considered class-related if a change within the region can shift the classifier's prediction toward or away from the target class. While the remaining region are class-independent. 

\[
p(x_R\mid x_S,x_C,y)
=
p(x_R\mid x_S,x_C),
\qquad
\forall y.
\]
By Bayes' rule,
\[
\begin{aligned}
p(t\mid x_S,x_C,x_R)
&=
\frac{
p(x_R\mid x_S,x_C,t)\,
p(x_S,x_C\mid t)\,
p(t)
}{
\sum_{j=1}^{K}
p(x_R\mid x_S,x_C,j)\,
p(x_S,x_C\mid j)\,
p(j)
}
\\
&=
\frac{
p(x_S,x_C\mid t)p(t)
}{
\sum_{j=1}^{K}
p(x_S,x_C\mid j)p(j)
}
\\
&=
p(t\mid x_S,x_C).
\end{aligned}
\]
Therefore, for the targeted loss
\[
\mathcal L_t(x)
=
-\log p(t\mid x),
\]
the ideal model gives
\[
\nabla_{x_R}\mathcal L_t(x)
=
0.
\]

This idealization is only used to illustrate how class-related support can be
attributed to the primary-subject and carrier regions. The implemented classifier
is not required to satisfy this factorization. Therefore, the normalized-update
analysis retains an arbitrary remaining-region
component $g_{R,t}$.

The primary-subject and carrier regions contain class-related visual content. For
$a\in\{S,C\}$, differentiating the target posterior with respect to the corresponding
image-region component gives
\[
\begin{aligned}
\nabla_{x_a}\mathcal L_t
&=
-\nabla_{x_a}\log p(t\mid x_S,x_C)
\\
&=
-\nabla_{x_a}\log p(x_S,x_C\mid t)
+
\sum_{j=1}^{K}
p(j\mid x_S,x_C)
\nabla_{x_a}\log p(x_S,x_C\mid j).
\end{aligned}
\]
A region is class-related when changes to its content can alter the classifier's
target prediction. Accordingly, the carrier region contributes a target-sensitive
component to the classifier gradient:
\[
\|g_{C,t}\|_2>0.
\]
Together with the regional decomposition,
the implemented gradient is represented as
\[
g_t
=
g_{S,t}+g_{C,t}+g_{R,t}.
\]

\subsection{From Two-Region to Three-Region Allocation}
\label{sec:regional_allocation}

For the No-Carrier gradient
\[
g_t^{\mathrm{NC}}
=
g_{S,t}+g_{R,t},
\]
we define the squared regional shares as
\[
\rho_{S,t}^{\mathrm{NC}}
=
\frac{\|g_{S,t}\|_2^2}
{\|g_{S,t}\|_2^2+\|g_{R,t}\|_2^2},
\qquad
\rho_{R,t}^{\mathrm{NC}}
=
\frac{\|g_{R,t}\|_2^2}
{\|g_{S,t}\|_2^2+\|g_{R,t}\|_2^2}.
\]
These shares satisfy
\[
\rho_{S,t}^{\mathrm{NC}}
+
\rho_{R,t}^{\mathrm{NC}}
=
1.
\]

For the Carrier gradient
\[
g_t^{\mathrm C}
=
g_{S,t}+g_{C,t}+g_{R,t},
\]
the corresponding shares are
\[
\rho_{S,t}^{\mathrm C}
=
\frac{\|g_{S,t}\|_2^2}
{\|g_{S,t}\|_2^2+\|g_{C,t}\|_2^2+\|g_{R,t}\|_2^2},
\]
\[
\rho_{C,t}^{\mathrm C}
=
\frac{\|g_{C,t}\|_2^2}
{\|g_{S,t}\|_2^2+\|g_{C,t}\|_2^2+\|g_{R,t}\|_2^2},
\qquad
\rho_{R,t}^{\mathrm C}
=
\frac{\|g_{R,t}\|_2^2}
{\|g_{S,t}\|_2^2+\|g_{C,t}\|_2^2+\|g_{R,t}\|_2^2},
\]
with
\[
\rho_{S,t}^{\mathrm C}
+
\rho_{C,t}^{\mathrm C}
+
\rho_{R,t}^{\mathrm C}
=
1.
\]

Under the matched comparison in
Assumption~\ref{ass:matched_components}, $g_{S,t}$ and $g_{R,t}$ are fixed.
Because the class-related carrier contributes $\|g_{C,t}\|_2>0$, we obtain
\[
\rho_{S,t}^{\mathrm C}
<
\rho_{S,t}^{\mathrm{NC}},
\qquad
\rho_{R,t}^{\mathrm C}
<
\rho_{R,t}^{\mathrm{NC}},
\qquad
\rho_{C,t}^{\mathrm C}>0.
\]
The carrier therefore introduces an additional class-related share of the complete
gradient, reducing the relative shares assigned to the primary-subject and remaining
regions under the matched comparison.

\subsection{Carrier-Induced Reduction under Global RMS Normalization}

\subsection{Carrier-Induced Reduction under Global RMS Normalization}
\label{sec:normalized_subject_update}

For a gradient $g_t$ with $n$ elements, global RMS normalization gives
\[
\operatorname{Norm}_{\mathrm{RMS}}(g_t)
=
\sqrt{n}\,
\frac{g_t}{\|g_t\|_2}.
\]
Using the orthogonal regional decomposition, its squared norm is
\[
\|g_t\|_2^2
=
\|g_{S,t}\|_2^2
+
\|g_{C,t}\|_2^2
+
\|g_{R,t}\|_2^2.
\]

Under Assumption~\ref{ass:matched_components}, the No-Carrier and Carrier gradients
are
\[
\begin{aligned}
\text{No-Carrier:}\qquad
&g_t^{\mathrm{NC}}
=
g_{S,t}+g_{R,t},\\
\text{Carrier:}\qquad
&g_t^{\mathrm C}
=
g_{S,t}+g_{C,t}+g_{R,t},
\end{aligned}
\]
while $g_{S,t}$, $g_{R,t}$, $n$, and the guidance coefficient
$\lambda\sigma_t$ are held fixed.

The corresponding subject-region components of the classifier-guided velocity
increment are
\[
\begin{aligned}
\text{No-Carrier:}\qquad
\delta_{S,t}^{\mathrm{NC}}
&=
\lambda\sigma_t\sqrt{n}\,
\frac{g_{S,t}}
{\sqrt{
\|g_{S,t}\|_2^2+
\|g_{R,t}\|_2^2
}},
\\[6pt]
\text{Carrier:}\qquad
\delta_{S,t}^{\mathrm C}
&=
\lambda\sigma_t\sqrt{n}\,
\frac{g_{S,t}}
{\sqrt{
\|g_{S,t}\|_2^2+
\|g_{C,t}\|_2^2+
\|g_{R,t}\|_2^2
}}.
\end{aligned}
\]

The two updates have the same numerator. Because the carrier contributes a
class-related gradient component,
\[
\|g_{C,t}\|_2^2>0,
\]
their denominators satisfy
\[
\sqrt{
\|g_{S,t}\|_2^2+
\|g_{C,t}\|_2^2+
\|g_{R,t}\|_2^2
}
>
\sqrt{
\|g_{S,t}\|_2^2+
\|g_{R,t}\|_2^2
}.
\]
It follows that
\[
\left\|
\delta_{S,t}^{\mathrm C}
\right\|_2
<
\left\|
\delta_{S,t}^{\mathrm{NC}}
\right\|_2,
\]
which proves Theorem~\ref{thm:subject_update}.

Therefore the carrier provides an additional class-related region within the globally normalized update. In the main attack in~\ref{tab:main_attack_analysis}, this additional component
reduces the update applied to the primary-subject region at the same guidance
strength. As the attack is strengthened as~\ref{tab:subject_preservation}, the additional attack signal does not need
to be borne entirely by the primary subject.

This result concerns the applied globally normalized update. It does not state that
introducing the carrier directly reduces the raw subject-gradient magnitude
$\|g_{S,t}\|_2$, which is held fixed in the matched comparison. The corresponding
final-image preservation benefit is evaluated empirically in
Section~\ref{sec:preservation}.

\section{Targeted Transferability Improvement}
\label{app:transferability_improvement}
This appendix provides the complete proof of
Theorem~\ref{thm:transferability}.

\subsection{Carrier Conditions and Target Margins}
Let $\mathcal{D}=\{\mathrm{NC},\mathrm{NT},\mathrm{H},\mathrm{T}\}$ denote the No-Carrier, Non-Target Carrier, Hybrid Carrier, and Target Carrier conditions. For a fixed source, target, seed, and attack configuration, \(x_d^{\mathrm{adv}}\) denotes the final adversarial image obtained under condition \(d\in\mathcal D\). We also introduce the ideal classifier \(p^*(\cdot\mid x)\), and let \(t\) the target class.
As defined in
Section~\ref{sec:carrier_transferability_principle}, its targeted Top-1 margin under
condition $d$ is
\[
\gamma_d^*
:=
p_t^*\!\left(x_d^{\mathrm{adv}}\right)
-
\max_{y\neq t}
p_y^*\!\left(x_d^{\mathrm{adv}}\right).
\]
For a BlackBox classifier $m \sim \mathcal{Q}$, define
\[
\gamma_{m,d}
=
p_{m,t}\!\left(x_d^{\mathrm{adv}}\right)
-
\max_{y\neq t}
p_{m,y}\!\left(x_d^{\mathrm{adv}}\right).
\]
The targeted transfer probability is
\[
T_d
=
\Pr_{m\sim Q}
\left[
\operatorname{Top1}_m
\left(x_d^{\mathrm{adv}}\right)=t
\right].
\]

Under Assumption~\ref{ass:margin_ordering}, increasingly target-related carrier
conditions have non-decreasing ideal target margins:
\[
\gamma_{\mathrm T}^*
\geq
\gamma_{\mathrm H}^*
\geq
\gamma_{\mathrm{NT}}^*
\geq
\gamma_{\mathrm{NC}}^*.
\]

The discrepancy between the ideal classifier and a practical black-box classifier
$m$ under condition $d$ is
\[
D_{m,d}
=
D_{\mathrm{KL}}
\left(
p^*(\cdot\mid x_d^{\mathrm{adv}})
\;\middle\|\;
p_m(\cdot\mid x_d^{\mathrm{adv}})
\right).
\]
Under Assumption~\ref{ass:classifier_discrepancy}, this discrepancy has a uniform
expected upper bound across all construction conditions:
\[
\mathbb E_{m\sim Q}[D_{m,d}]
\leq
\bar{\kappa},
\qquad
\forall d\in\mathcal D.
\]

\subsection{Black-Box Margin Bound}
By Pinsker’s inequality, the probability assigned to any class $y$ by the black-box
classifier differs from that of the ideal classifier by at most
\[
\left|
p_{m,y}\!\left(x_d^{\mathrm{adv}}\right)
-
p_y^*\!\left(x_d^{\mathrm{adv}}\right)
\right|
\leq
\sqrt{\frac{D_{m,d}}{2}},
\qquad
\forall y.
\]

In particular, the black-box target probability satisfies
\[
p_{m,t}\!\left(x_d^{\mathrm{adv}}\right)
\geq
p_t^*\!\left(x_d^{\mathrm{adv}}\right)
-
\sqrt{\frac{D_{m,d}}{2}},
\]
while its largest non-target probability satisfies
\[
\max_{y\neq t}
p_{m,y}\!\left(x_d^{\mathrm{adv}}\right)
\leq
\max_{y\neq t}
p_y^*\!\left(x_d^{\mathrm{adv}}\right)
+
\sqrt{\frac{D_{m,d}}{2}}.
\]

Subtracting the largest non-target probability from the target probability gives
\[
\begin{aligned}
\gamma_{m,d}
&=
p_{m,t}\!\left(x_d^{\mathrm{adv}}\right)
-
\max_{y\neq t}
p_{m,y}\!\left(x_d^{\mathrm{adv}}\right)
\\
&\geq
p_t^*\!\left(x_d^{\mathrm{adv}}\right)
-
\max_{y\neq t}
p_y^*\!\left(x_d^{\mathrm{adv}}\right)
-
2\sqrt{\frac{D_{m,d}}{2}}
\\
&=
\gamma_d^*
-
\sqrt{2D_{m,d}}.
\end{aligned}
\]

For a positive ideal target margin, if
\[
D_{m,d}
<
\frac{(\gamma_d^*)^2}{2},
\]
then
\[
\sqrt{2D_{m,d}}
<
\gamma_d^*,
\]
and therefore
\[
\gamma_{m,d}
>
0.
\]
A positive black-box target margin means that the target probability is larger than
every non-target probability. Hence,
\[
D_{m,d}
<
\frac{(\gamma_d^*)^2}{2}
\quad\Longrightarrow\quad
\operatorname{Top1}_m
\left(x_d^{\mathrm{adv}}\right)=t.
\]

\subsection{Transferability Lower Bound}

\label{sec:transferability_lower_bound}

The sufficient condition derived about transferability margin implies
\[
T_d
\geq
\Pr_{m\sim Q}
\left[
D_{m,d}
<
\frac{(\gamma_d^*)^2}{2}
\right].
\]
Equivalently,
\[
T_d
\geq
1-
\Pr_{m\sim Q}
\left[
D_{m,d}
\geq
\frac{(\gamma_d^*)^2}{2}
\right].
\]

Because $D_{m,d}\geq0$, Markov's inequality gives
\[
\Pr_{m\sim Q}
\left[
D_{m,d}
\geq
\frac{(\gamma_d^*)^2}{2}
\right]
\leq
\frac{
\mathbb E_{m\sim Q}[D_{m,d}]
}{
(\gamma_d^*)^2/2
}.
\]
Under Assumption~\ref{ass:classifier_discrepancy},
\[
\mathbb E_{m\sim Q}[D_{m,d}]
\leq
\bar{\kappa},
\]
and therefore
\[
\Pr_{m\sim Q}
\left[
D_{m,d}
\geq
\frac{(\gamma_d^*)^2}{2}
\right]
\leq
\frac{2\bar{\kappa}}
{(\gamma_d^*)^2}.
\]

For each condition $d$ with a positive ideal target margin, it follows that
\[
T_d
\geq
\left[
1-
\frac{2\bar{\kappa}}
{(\gamma_d^*)^2}
\right]_+.
\]
Defining the certified targeted transferability lower bound as
\[
L_d
:=
\left[
1-
\frac{2\bar{\kappa}}
{(\gamma_d^*)^2}
\right]_+,
\]
we obtain
\[
T_d\geq L_d.
\]

For a fixed $\bar{\kappa}$, the lower bound $L_d$ is non-decreasing with respect to
the positive ideal target margin $\gamma_d^*$. Therefore, within the positive-margin
regime, Assumption~\ref{ass:margin_ordering} yields
\[
L_{\mathrm T}
\geq
L_{\mathrm H}
\geq
L_{\mathrm{NT}}
\geq
L_{\mathrm{NC}}.
\]
This proves Theorem~\ref{thm:transferability}.

Therefore, progressively stronger target-related carrier conditions produce a monotonically non-decreasing guaranteed lower bound on targeted Top-1 transferability. This is also consistent with our transferability results in Section~\ref{transferability}, where the transfer rate improves from the No-Carrier setting to the carrier-based settings. The corresponding realized cross-model
transferability results are reported in Section~\ref{transferability} and
Appendix~\ref{app:BlackBox_Results}.
\section{BlackBox Results}
\label{app:BlackBox_Results}

\subsection{Complete BlackBox Top-5 Results}
All adversarial images are generated by using WhiteBox model ResNet-50 classifier guidance. After generation, each final image is fixed and directly evaluated on ResNet-101, VGG-19, Inception-v3, ConvNeXt-B, and Swin-B. No model-specific regeneration, additional optimization, or gradient access is used during BlackBox evaluation. For each BlackBox model, an attack is counted as targeted Top-5 success when the designated target class appears among its five highest-probability predictions. The macro average assigns equal weight to the five BlackBox architectures.

\begin{table}[htbp]
\centering
\caption{BlackBox Top-5 results averaged over the Non-Target and Target Carrier conditions. Cond Avg. reports the conditional Top-5 ASR evaluated only on samples whose clean images do not contain the target class in the corresponding model's Top-5 predictions. For a consistent cross-route comparison, Hybrid is excluded because its requested target-related attributes are not reliably realized by the inpainting-based routes (Appendix~\ref{hybrid_incomplete}).}
\label{tab:blackbox_transfer}

\small
\setlength{\tabcolsep}{5pt}
\renewcommand{\arraystretch}{1.05}

\begin{tabular}{@{}lccccccc@{}}
\toprule
&
\multicolumn{5}{c}{BlackBox Models} &
\multicolumn{2}{c}{Overall} \\
\cmidrule(lr){2-6}
\cmidrule(l){7-8}

Method
& ResNet-101
& VGG-19
& Inception-v3
& ConvNeXt-B
& Swin-B
& Macro Avg.
& Cond Avg. \\
\midrule

CRA
& 30.17
& 10.42
& 10.33
& 32.67
& 20.08
& 20.73
& 16.71 \\

JIA
& \textbf{37.08}
& \textbf{15.08}
& \textbf{24.00}
& \textbf{41.58}
& \textbf{32.67}
& \textbf{30.08}
& \textbf{26.15} \\

CIRA
& 35.08
& 13.58
& 21.08
& 37.75
& 30.33
& 27.57
& 23.15 \\

\bottomrule
\end{tabular}
\end{table}
JIA obtains the highest macro-average Top-5 transferability at \(30.08\%\), followed by CIRA at \(27.57\%\) and CRA at \(20.73\%\). ConvNeXt-B exhibits the highest transfer rate for all three routes, whereas VGG-19 consistently gives the lowest. Although CRA and CIRA share the same inversion–return attack formulation, CIRA transfers more strongly after clean carrier conditioned inpainting. JIA achieves the strongest overall transfer by introducing classifier guidance directly into the inpainting trajectory.

\subsection{Clean-Image Prediction Audit}
\label{app:clean_black}

We additionally evaluate the clean images before attack using the five BlackBox
classifiers. Subject Top-5 denotes whether the predefined subject category appears
in the clean prediction Top-5, while Target-Absent Top-5 denotes whether the attack
target is absent from the clean Top-5.

\begin{table}[htb]
\centering
\caption{BlackBox prediction audit on clean images, averaged over the three attack routes and five BlackBox classifiers (\%).}
\label{tab:clean_blackbox_audit}
\small
\begin{tabular}{lcc}
\toprule
Condition & Subject Top-5 $\uparrow$ & Target-Absent Top-5 $\uparrow$ \\
\midrule
Non-Target Carrier & 82.19 & 98.67 \\
Hybrid Carrier     & 80.12 & 97.03 \\
Target Carrier     & 83.99 & 64.10 \\
\bottomrule
\end{tabular}
\end{table}

Across all carrier conditions, the clean images retain substantial BlackBox
Top-5 evidence for the subject category. The attack target is absent from the
clean Top-5 in approximately 97-99\% of Non-Target and Hybrid cases. Its lower
absence rate under Target Carrier is expected because the clean carrier directly
instantiates target-related visual content. These results indicate that the clean
construction generally preserves subject-category evidence and that the subsequent
conditional attack results are not primarily explained by pre-existing target
predictions.

\subsection{Comparison with No-Carrier}
The No-Carrier baseline build upon the original concept-based attack, using the same subject-specific LoRA, ResNet-50 target classifier, targeted cross-entropy objective, global RMS normalization, classifier scale \(0.5\), and 15-step finite-return attack.

\begin{table}[htbp]
\centering
\caption{No-Carrier baseline attack performance (\%).}
\label{tab:no_carrier_baseline}

\small
\setlength{\tabcolsep}{9pt}
\renewcommand{\arraystretch}{1.05}

\begin{tabular}{@{}lcccc@{}}
\toprule
&
\multicolumn{2}{c}{White-box Attack} &
\multicolumn{2}{c}{Black-box Transfer} \\
\cmidrule(lr){2-3}
\cmidrule(l){4-5}

Setting
& Top-1 $\uparrow$
& Top-5 $\uparrow$
& Top-1 $\uparrow$
& Top-5 $\uparrow$ \\
\midrule

No-Carrier
& 93.33
& 98.00
& 0.20
& 2.17 \\

\bottomrule
\end{tabular}
\end{table}

No-Carrier already achieves strong WhiteBox attack performance, reaching \(93.33\%\) targeted Top-1 and \(98.00\%\) targeted Top-5 success. In comparison, CRA and CIRA in Table~\ref{tab:main_attack_analysis} achieve WhiteBox Top-1 success rates between \(95.00\%\) and \(99.67\%\), and Top-5 success rates between \(98.50\%\) and \(99.83\%\), further improving upon the No-Carrier baseline. NatADiff, in contrast, achieves only \(31.83\%\) WhiteBox Top-1 success under its standard setting.

Despite its strong WhiteBox performance, No-Carrier exhibits limited cross-model transferability, with five-model macro-average BlackBox success rates of only \(0.20\%\) Top-1 and \(2.17\%\) Top-5. NatADiff improves the macro-average BlackBox Top-5 transfer rate to \(19.27\%\), but this improvement is accompanied by substantially lower WhiteBox attack success. In comparison, CRA and CIRA improve macro-average BlackBox Top-5 transferability while consistently maintaining high WhiteBox attack success. JIA achieves the highest aggregated BlackBox transferability in Table~\ref{tab:blackbox_transfer}, but its WhiteBox success varies substantially across carrier conditions.

Overall, all three carrier-based routes improve cross-model transferability over No-Carrier. CRA and CIRA preserve strong WhiteBox effectiveness, whereas JIA achieves stronger transferability with more condition-dependent WhiteBox performance.
More baseline about No-Carrier and NatADiff adaptation results are discussed in~\ref{app:baseline_realization}

\section{Subject Preservation Evaluation}
\label{app:preservation}
Subject preservation is evaluated between the clean input and attacked output. 

\subsection{DINOv3 Subject Similarity}
We run the SAM3 independently on the clean and attacked images with the same subject prompt. Ideally, only one subject is detected for both the attacked and clean images. Multiple detected instances are merged only during evaluation to obtain the clean and attacked evaluation masks \(M_{\mathrm{clean}}\) and \(M_{\mathrm{adv}}\). A shared crop is defined by the union bounding box of the two masks with \(10\%\) padding. Within this crop, each image uses its independently predicted mask, while pixels outside the mask are replaced by RGB gray \((128,128,128)\). We extract the FP32 CLS representation from DINOv3 ViT-L/16 by ~\citet{dinov3}, apply \(L_2\) normalization, and compute
\[
\operatorname{DINO}(x_{\mathrm{clean}}, x_{\mathrm{adv}})
=
\frac{
\phi(x_{\mathrm{clean}})^{\top}\phi(x_{\mathrm{adv}})
}{
\left\|\phi(x_{\mathrm{clean}})\right\|_2
\left\|\phi(x_{\mathrm{adv}})\right\|_2
}.
\]

This metric measures only the clean-to-attacked similarity of the segmented subject
representation; it does not independently establish whether the subject was
successfully generated in either image.

\subsection{SAM3 Mask IoU}
Spatial consistency is also measured with the predicted clean and attacked masks:
\[
\operatorname{IoU}(M_{\mathrm{clean}}, M_{\mathrm{adv}})
=
\frac{
\left|M_{\mathrm{clean}} \cap M_{\mathrm{adv}}\right|
}{
\left|M_{\mathrm{clean}} \cup M_{\mathrm{adv}}\right|
}.
\]

Only when the subject is both detected in clean and attack images, the spatial consistency is computed. 
Pairs in which the designated subject is not detected in either the clean or attacked
image are excluded from the DINO and IoU averages, and their missing-detection rate
is reported separately.

\subsection{Preservation Analysis of Main Experiment}
As reported in Table~\ref{tab:main_attack_analysis}, all nine combinations of attack route and carrier condition retain high subject consistency. DINOv3 similarity ranges from \(0.9799\) to \(0.9960\), while SAM3 mask IoU ranges from \(0.9898\) to \(0.9935\). JIA obtains the highest DINOv3 similarities, ranging from \(0.9958\) to \(0.9960\). CRA and CIRA obtain DINOv3 similarities of \(0.9799\)–\(0.9857\), while preserving similarly high spatial overlap. These results indicate that the carrier-based attacks consistently preserve the subject in both feature similarity and spatial structure across different attack routes and carrier conditions.
For NatADiff, preservation metrics are available for 409 of 600 pairs (68.17\%).
Because the remaining 31.83\% contain at least one image without a detected
personalized subject, the reported NatADiff DINO and IoU averages characterize only
the valid detected subset. Detailed missing-pair analysis is provided in
Appendix~\ref{app:baseline_realization}.

\subsection{Preservation Analysis of Strong Targeted Attacks}

We compare Target-Carrier CRA, Target-Carrier CIRA, and No-Carrier under the matched
15-step strong-attack setting. According to Table~\ref{tab:subject_preservation}, these three conditions reach \(100\%\) targeted Top-1 success. Target-Carrier CRA and CIRA retain DINO similarities of 0.9345 and 0.9809,
respectively, compared with 0.7928 for No-Carrier. Their SAM3 mask IoU values remain
comparable, indicating that the principal difference lies in appearance-level subject
consistency rather than coarse spatial extent.

We also evaluate the strong NatADiff setting, which reaches 98.83\% targeted
Top-1 success but obtains a DINO similarity of 0.7461 and a SAM3 IoU of 0.9149.
Moreover, its preservation metrics are available for only 51.00\% of total. The reported averages therefore characterize a selected valid subset and should not
be interpreted as full-benchmark preservation scores. NatADiff retains its
method-specific time-travel schedule and consequently receives a larger optimization
budget than the matched 15-step finite-return attacks. Detailed about baseline setting and results are in~\ref{app:baseline_realization}

\subsection{Qualitative Analysis of JIA under Strong Guidance}
JIA may introduce more visible changes or artifacts in the
editable background, while the primary subject generally remains stable because
native inpainting blending repeatedly restores the source-conditioned foreground
region. Comparable background changes are less frequent in CRA and CIRA, which apply
finite-return attacks to already completed clean images. These observations are
qualitative and are not included in the matched reconstruction comparison in
Table~\ref{tab:subject_preservation}.
\section{Regional Gradient Diagnostics and Grad-CAM}

This appendix provides qualitative analyses of where classifier-sensitive signals are
spatially expressed. These measurements are computed after attack generation and do
not modify the globally guided attack procedures defined in Appendix~\ref{app:algorithm}.

\subsection{Regional Gradient Allocation}
Let \(g_k=\nabla_{z^{(k)}}\mathcal L_t\) denote the classifier gradient at return step \(k\). Using the spatially resized subject, carrier, and remaining-region masks \(M_S\), \(M_C\), and \(M_R\), we define
\[
G_{S,k}=\|M_S\odot g_k\|_2,\qquad
G_{C,k}=\|M_C\odot g_k\|_2,\qquad
G_{R,k}=\|M_R\odot g_k\|_2.
\]
and the normalized regional gradient magnitudes are

\[
\rho_{S,k}=\frac{G_{S,k}}{\|g_k\|_2},\qquad
\rho_{C,k}=\frac{G_{C,k}}{\|g_k\|_2},\qquad
\rho_{R,k}=\frac{G_{R,k}}{\|g_k\|_2}.
\]

These measurement only describe the regional allocation of the classifier gradient before the global RMS normalization. Detailed gradient decomposition can be found in Appendix~\ref{app:attack_update_allocation}. Our attacks still remain apply the complete global gradient without spatial masking.

\subsection{Grad-CAM}

We use Grad-CAM to visualize spatial evidence associated with the designated target class \(t\). We use the final convolutional block, \(\mathrm{layer4}[-1]\), and compute

For the designated target class $t$, we compute
\[
H^{t}
=
\operatorname{ReLU}
\left(
\sum_q \alpha_q^{t}A^q
\right),
\qquad
\alpha_q^{t}
=
\frac{1}{|\Omega|}
\sum_{u\in\Omega}
\frac{\partial z_t}{\partial A_u^q},
\]
where $A^q$ is feature channel $q$ and $z_t$ is the target-class logit. The resulting
map is bilinearly resized to the classifier input resolution and normalized for
visualization.

For successful attacks, this map localizes spatial evidence associated with the final
target prediction. For unsuccessful attacks, it shows regions that support the target
logit even though this evidence is insufficient to make the target the Top-1 class.
Grad-CAM is used only as a qualitative localization tool and does not measure the raw
attack-gradient magnitude or establish causal contribution.

\subsection{Qualitative Comparison across Attack Routes}
\label{qualitative_analysis}

\subsubsection{Target-Carrier Case Analysis}

\begin{figure*}[ht]
    \centering
    \includegraphics[width=0.95\textwidth]{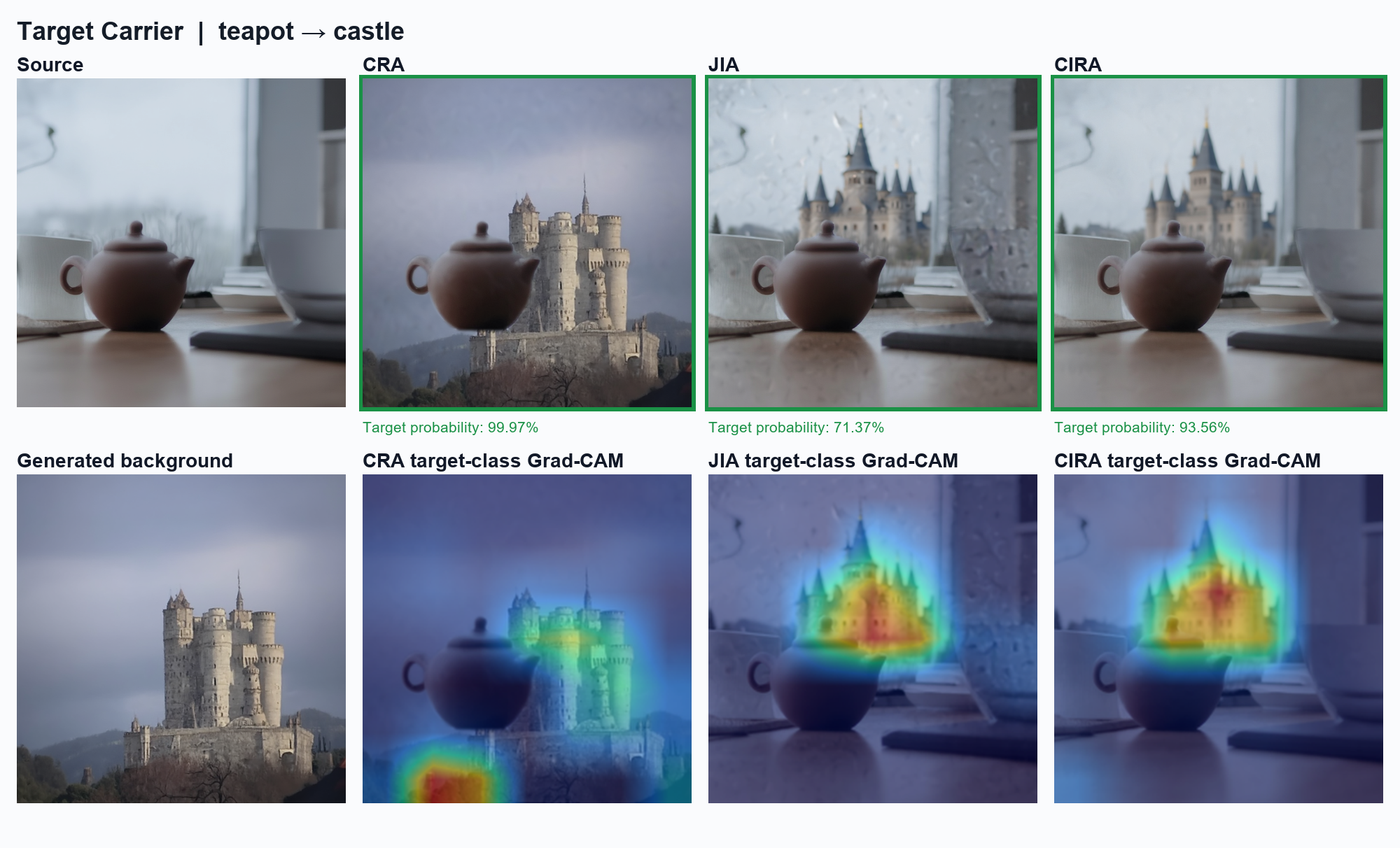}
    \caption{
    Qualitative comparison of CRA, JIA, and CIRA under the Target-Carrier condition for
    teapot$\rightarrow$castle. The bottom row shows the generated carrier background and
    target-class Grad-CAM maps for the three attacked outputs.
    }
    \label{fig:gradcam_comparison}
\end{figure*}

Figure~\ref{fig:gradcam_comparison} presents matched Target-Carrier example in which a personalized teapot is attacked toward the target class castle. The top row shows the source image and the final outputs of CRA, JIA, and CIRA, while the bottom row shows the independently generated carrier background and the corresponding Grad-CAM maps. All three routes preserve the teapot as the primary foreground object while introducing castle-related visual evidence into the surrounding scene.

For CRA the strongest Grad-CAM response lies mainly on the visible castle structure composited into the background, indicating that the final target prediction is associated with the explicitly constructed carrier region. JIA produces a more centrally distributed response over the castle-like structure generated during the jointly guided inpainting trajectory. CIRA shows a similarly background-oriented but more spatially distributed response after attacking the completed clean inpainting result.

\begin{figure*}[ht]
    \centering
    \includegraphics[width=0.95\textwidth]
    {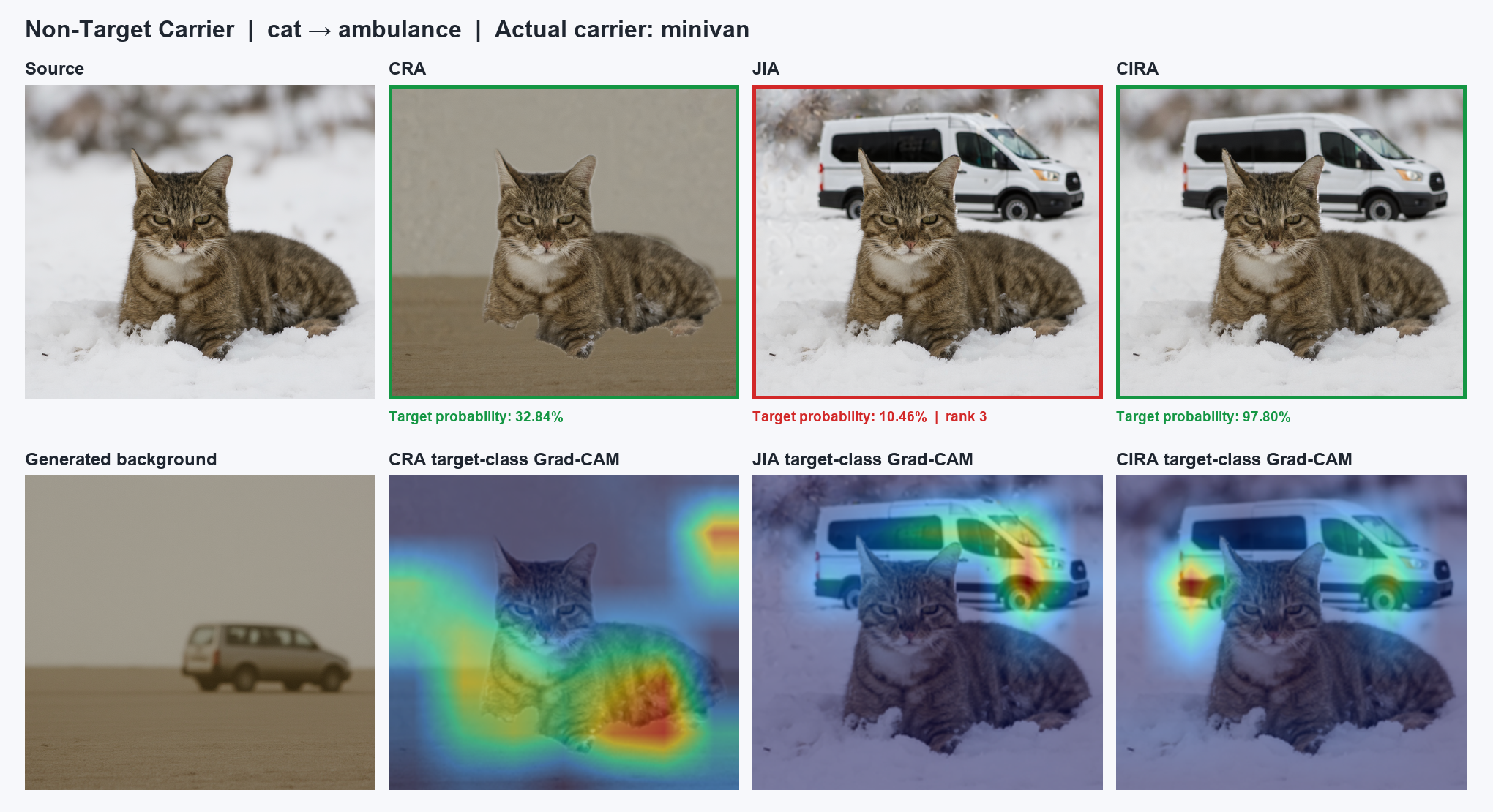}
    \caption{
    Non-Target-Carrier example for cat $\rightarrow$ ambulance. CRA largely
occludes the generated carrier; JIA retains the carrier but fails to reach targeted
Top-1; CIRA preserves both regions and succeeds. Green and red indicate targeted
Top-1 success and failure, respectively.
    }
    \label{fig:nontarget_carrier_gradcam}
\end{figure*}

\subsection{Non Target Carrier Case Analysis}

Figure~\ref{fig:nontarget_carrier_gradcam} resents a Non-Target-Carrier example in
which the primary subject, cat, is attacked toward the target class ambulance, while the constructed carrier is a semantically distinct minivan. These three routes exhibit
different outcomes that reflect how carrier construction interacts with adversarial
optimization.  In CRA, the independently generated carrier
is largely occluded after subject compositing, illustrating a limitation of direct
composition. In JIA, the carrier remains visible and receives target-class Grad-CAM
activation, but the target reaches only rank 3, indicating insufficient target evidence
under the current guidance. CIRA preserves both the subject and carrier while achieving
97.80\% target probability, with the strongest target-related response concentrated on
the vehicle region. For both CIRA and JIA, the Grad-CAM responses are primarily concentrated on the carrier region. Notably, although JIA does not achieve the ambulance target as its Top-1 prediction, the target-class Grad-CAM map for ambulance still localizes predominantly on the visible minivan carrier. This qualitatively shows that the target-class activation is concentrated on the visible carrier, even though it is insufficient for targeted Top-1 success.

\section{whitening}
\label{app:whitening}
\subsection{Intervention Protocol}

The whitening intervention is only apply to CRA outputs because CRA constructs the carrier in an independently generated background before compositing it with the personalized subject. For each composite, we manually localize the visible carrier in the generated background and map its bounding box to the attacked image.
To avoid modifying the personalized subject, we remove the overlap between the
carrier bounding box and the final subject mask.
\[
B_{\mathrm{vis}}
=
B_{\mathrm{carrier}}
\cap
(1-m_S),
\]
where \(B_{\mathrm{carrier}}\) is the mapped carrier bounding box and \(n_S\) is the Subject mask. Pixels within \(B_{\mathrm{vis}}\) are replaced with pure white, while all other pixels remain unchanged:
\[
x_{\mathrm{white}}(u)
=
\begin{cases}
(255,255,255), & u \in B_{\mathrm{vis}}, \\
x_{\mathrm{adv}}(u), & u \notin B_{\mathrm{vis}}.
\end{cases}
\]

\begin{figure*}[htbp]
    \centering
    \includegraphics[width=\textwidth]{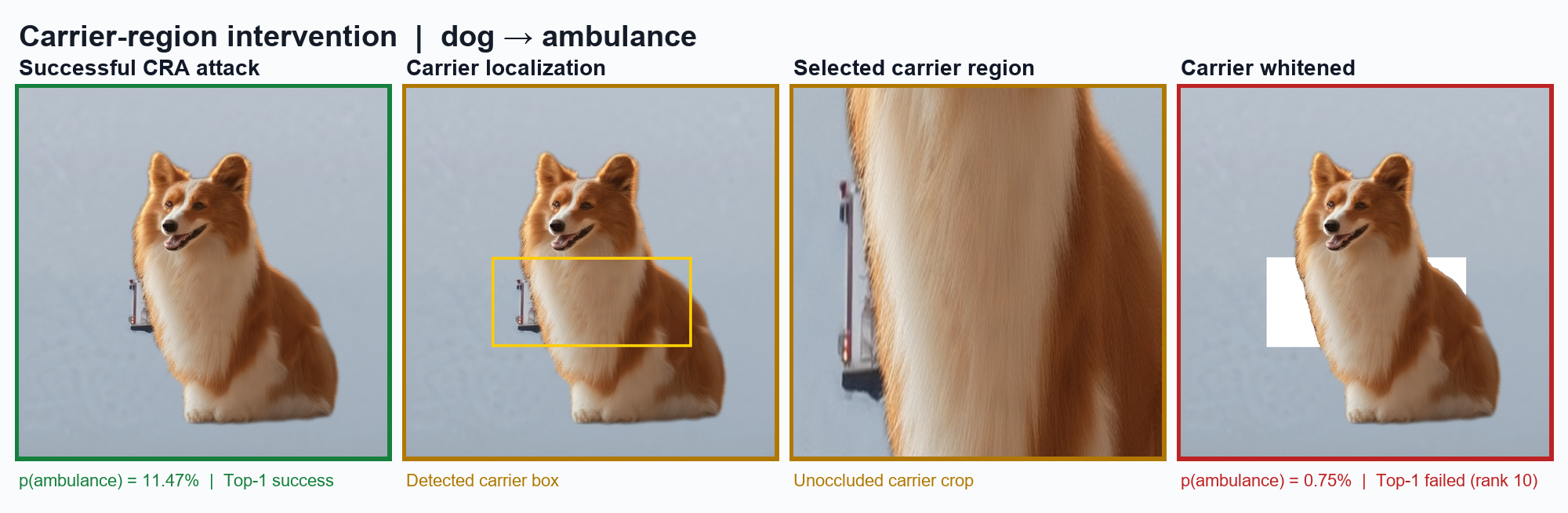}
    \caption{
    Example of the carrier-region whitening intervention for a successful CRA attack on
    dog$\rightarrow$ambulance. The visible carrier is localized, its unoccluded region is
    selected, and the region is whitened. After intervention, the target probability drops from
    \(11.47\%\) to \(0.75\%\), causing the target class to fall from Top-1 to rank 10.
    }
    \label{fig:carrier_whitening}
\end{figure*}

\subsection{Target-Logit Change}

Let \(z_t(x)\) denote the ResNet-50 logit of the attack target. We measure the change caused by whitening \(\Delta z_t = z_t(x_{\mathrm{adv}}) - z_t(x_{\mathrm{white}})\).

Among all 1,200 CRA composites, our intervention is feasible for 1,163. In the remaining \(37\) images, it was fully occluded by the subject or could not be reliably localized. By whitening the visible carrier region, target logit decreases in \(1{,}143\) of the \(1{,}163\) CRA images, corresponding to \(98.3\%\).
Removing a larger portion of the visible carrier is associated with a greater decrease in the target logit:
\[
\rho_{\mathrm{Spearman}} = 0.774,
\qquad
p < 0.001.
\]

\subsection{Intervention-caused Loss of Target Top-1 Prediction}

We further examine attacks that initially predict the designated target as Top-1. Among \(1{,}130\) initially successful CRA attacks, \(597\) lose the target Top-1 prediction after whitening, giving an overall failure rate of \(52.8\%\). The intervention produces comparable failure rates under the Target-Carrier and
Non-Target Carrier conditions, at 54.6\% and 51.0\%, respectively.

The overall 1,130 initial success refers to the attack success image before whitening.
Together, the logit reduction and Top-1 failure results show that the visible carrier region contains prediction-relevant evidence in a substantial fraction of successful attacks.

\begin{table}[htbp]
\centering
\caption{CRA attack failures after whitening the visible carrier region.}
\label{tab:carrier_intervention}

\small
\setlength{\tabcolsep}{9pt}
\renewcommand{\arraystretch}{1.05}

\begin{tabular}{@{}lccc@{}}
\toprule

Condition
& Initial Success
& Failed
& Failure Rate \\

\midrule

Target Carrier
& 575
& 314
& 54.6\% \\

Non-Target Carrier
& 555
& 283
& 51.0\% \\

\midrule

All
& 1130
& 597
& 52.8\% \\

\bottomrule
\end{tabular}
\end{table}

Our ablation design is motivated by two related ideas. \citet{objectcompose}
motivate intervening on non-subject regions to examine prediction dependence,
while \citet{doublemainbody} motivate separating the predictive roles of
multiple visible objects. Accordingly, our whitening ablation removes only the
visible, non-overlapping carrier region and measures the resulting changes in
the target logit and Top-1 prediction while leaving the primary subject unchanged.
\section{related work realizations}
\label{app:baseline_realization}

\subsection{No-Carrier realization and measurement}
No-Carrier build upon the original concept-based attack by~\citet{conceptbased} on FLUX. It uses the same selected personalized source, subject-specific LoRA, ResNet-50
classifier, targeted cross-entropy objective, global RMS normalization, and
finite-return inversion as CRA and CIRA, but does not introduce a carrier, carrier
prompt, or carrier-construction mask. Classifier guidance is also applied globally to the complete latent state.

No-Carrier achieves strong WhiteBox attack success, but its BlackBox transferability remains substantially lower than that of our carrier-based methods according to Table~\ref{tab:main_attack_analysis}. This suggests that directly applying classifier guidance without an explicit carrier is sufficient for attacking the victim model, but provides limited cross-model transferability compared with introducing a dedicated carrier region.

\subsection{NatADiff realization}
NatADiff by~\citet{natadiff} was originally designed for class-level attack rather than preservation of a specific personalized subject. Because NatADiff shares the similar objective as our method: strengthening the attack by introducing target-related features, even with a different implementation, we adapt it to FLUX by keeping the subject-specific LoRA active during sampling and using the same personalized prompts, target classes, and ResNet-50 classifier as the other methods.

For each source-target pair, the clean and attacked outputs are generated from the
same initial noise using matched seeds, prompts, LoRA, and sampling configurations.
Adversarial guidance is disabled for the clean image generation and enabled for the attacked
output. NatADiff does not invert and reconstruct clean images; both members of its evaluation pair are newly generated from matched noise.

To maintain same adversarial configuration with us, we keep NatADiff adaptation with approximately 65 active classifier-gradient updates. The main setting uses classifier
scale $0.5$ and maximum update norm $10$, whereas the strong setting uses scale $5.0$
and maximum update norm $100$.

\subsection{Subject-Generation Reliability of NatADiff and measurement}

NatADiff adaptation does not consistently generate a recognizable
instance of the designated subject. There are only (68.17\%) of total pairs are valid pairs for SAM3 that both clean and attacked images containing designated subject under main attack setting. (409 in 600) Among the remaining 191 invalid pairs, 157 contain no detected subject in the clean image,
corresponding to 26.17\% of the complete benchmark and 82.20\% of all invalid pairs. Specifically, 149 pairs fail detection in both images, 8 pairs fail in the clean
image, and 34 fail only in the attacked image.

\begin{figure*}[htbp]
    \centering
    \includegraphics[width=0.96\textwidth]{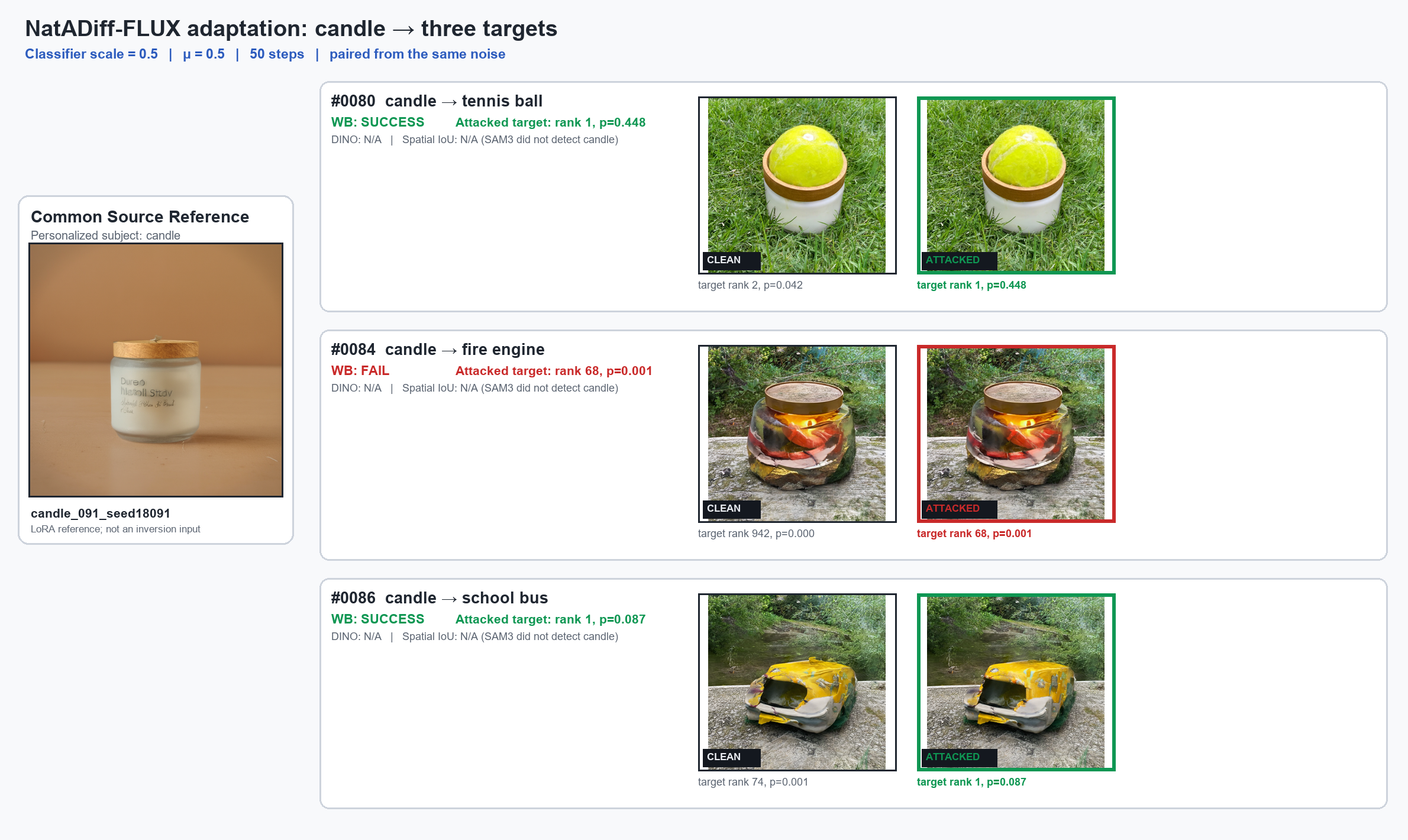}
    \caption{
    Personalized-subject generation failures of the NatADiff-FLUX adaptation
    under the main setting. The left panel shows the common candle reference used for
    personalization; it is not an inversion input. Each row shows matched-noise clean and
    attacked outputs for a different target. Although adversarial guidance is disabled
    for clean generation, the outputs are dominated by target-related object
    characteristics and SAM3 cannot detect the designated candle. Consequently, DINO
    similarity and mask IoU are unavailable. Green and red borders indicate targeted
    Top-1 success and failure, respectively.
    }
    \label{fig:natadiff_candle_failures}
\end{figure*}

\begin{figure*}[htbp]
    \centering
    \includegraphics[width=0.96\textwidth]{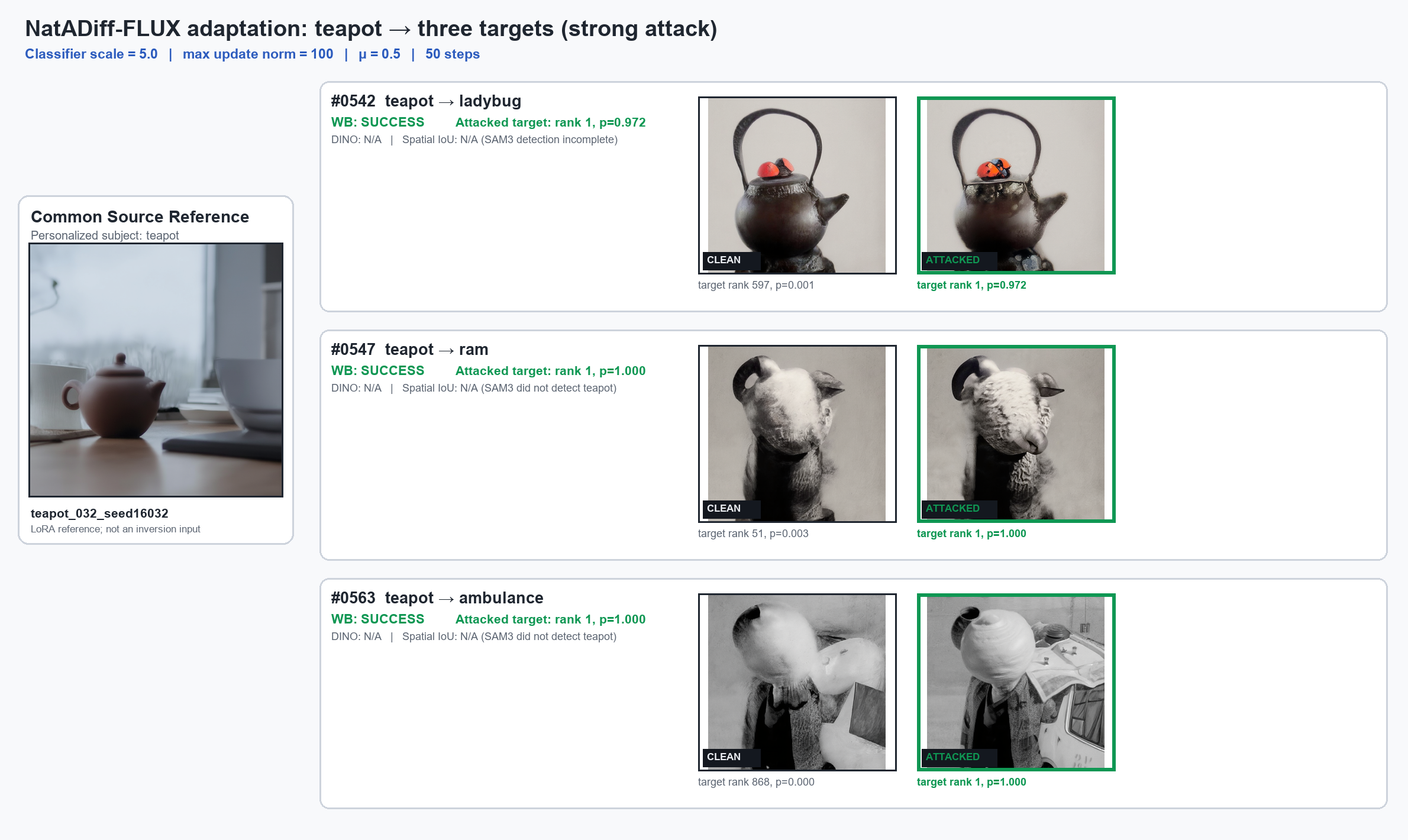}
    \caption{
    NatADiff adaptation generation under the strong attack setting. The left panel
    shows the common personalized teapot reference, while each row presents matched-noise
    clean and attacked outputs for one target. All attacked images achieve targeted
    Top-1 success, but the teapot characteristics are severely altered or replaced by
    target-related content. Because SAM3 detection is missing or incomplete, DINO
    similarity and mask IoU cannot be computed for these pairs.
    }
    \label{fig:natadiff_teapot_strong}
\end{figure*}

The clean images detection failures are particularly important because adversarial guidance is disabled during clean generation. Figure~\ref{fig:natadiff_candle_failures} shows three
candle cases under the main setting. This can indicate the instability in NatADiff adaptation's personalized subject generation rather than merely attack-induced degradation. Qualitatively, target-related shapes, textures, or parts can be fused directly into the personalized subject even during clean generation, substantially altering its original semantics concept despite the active subject-specific LoRA. Under stronger guidance, this effect becomes more severe as target-related features increasingly dominate the subject itself.Figure~\ref{fig:natadiff_teapot_strong} shows three teapot cases for which all attacked
outputs achieve targeted Top-1 success. However, the teapot identity is substantially
altered or replaced by ladybug-, ram-, and ambulance-related characteristics. SAM3
cannot reliably localize the designated teapot in the corresponding clean--attacked
pairs, leaving DINO similarity and mask IoU unavailable. These examples demonstrate
that high attack success does not imply preservation of the personalized subject.

In our setting, NatADiff adaptation cannot control the region or strength of target-related feature during generation. SAM3 is an independent vision-based segmentation model rather than the victim model ResNet-50. Therefore, a failure of SAM3 to detect the subject cannot be directly interpreted as a consequence of attack success.

\begin{table}[ht]
\centering
\caption{NatADiff adaptation attack strength and subject-generation reliability over 600
source--target pairs. Preservation metrics are computed only on valid pairs with
successful subject detection in both images.}
\label{tab:natadiff_settings}
\small
\begin{tabular}{lcccc}
\toprule
Setting & Top-1 ASR $\uparrow$ & Valid Pairs & DINO $\uparrow$ & IoU $\uparrow$ \\
\midrule
Main & 31.83 & 409/600 & 0.9411 & 0.9649 \\
Strong & 98.83 & 306/600 & 0.7461 & 0.9149 \\
\bottomrule
\end{tabular}
\end{table}

As a result, NatADiff adaptation already shows smaller DINO similarity and SAM3 IoU than No-Carrier and carrier-based methods in Table~\ref{tab:main_attack_analysis}. Moreover, these values likely overestimate its subject preservation, because DINO and IoU are computed only when SAM3 successfully detects the designated subject in both clean and attacked images. Many NatADiff adaptation clean outputs already fail this requirement, indicating that the reported scores reflect only the successfully detected pairs.

Under our setting, NatADiff adaptation also exhibits relatively low attack performance under the main setting. It BlackBox transferability is relatively higher that  of our Hybrid-Carrier condition, but it remains substantially lower than the transferability achieved by Target-Carrier.  Under stronger guidance, the number of valid preservation pairs further decreases to 306 out of 600 (51.00\%). Although the white-box Top-1 success rate increases from 31.83\% to 98.83\%, DINO similarity drops sharply to 0.7461, substantially below that of our carrier-based methods, while SAM3 IoU also decreases to 0.9149. Moreover, as discussed above, the reported DINO score is computed only on the successfully detected valid pairs and therefore likely overestimated.

\subsection{LoRA influence on NatADiff}
Subject-specific LoRA partially restricts NatADiff adaptation's generation freedom by biasing the sampling process toward the personalized concept, but it should also help preserve the subject.
However, NatADiff does not explicitly control where target-related characteristics are introduced during generation. As a result, these features tend to expressed directly on the personalized subject rather than in a separate region, altering its shape, texture, or semantic identity. Notably, such instability persists even with the subject-specific LoRA active, indicating that LoRA conditioning alone is insufficient to reliably preserve the personalized subject under NatADiff adaptation's unconstrained target-feature generation.

\subsection{Relation to Other Diffusion-Based Unrestricted Attacks}

Several recent diffusion-based unrestricted attacks are related to our setting but address different attack and preservation objectives. ObjectAdv~\citep{objectadv} localizes adversarial modification to the primary object while preserving the surrounding background, primarily to reduce unnecessary global distortion. Our setting considers a complementary spatial objective: the personalized primary subject is the region that should be preserved, while a spatially separate non-subject carrier is introduced to provide an alternative region for targeted adversarial evidence. Thus, although both approaches impose spatial control on diffusion-based attacks, the protected and adversarially exploited regions play fundamentally different roles.

Dual-label guided unrestricted attack~\citep{dual} combines source- and target-label guidance during diffusion to balance targeted attack effectiveness with preservation of source-category semantics. This form of preservation remains class-level, since maintaining characteristics associated with the source category does not necessarily preserve the identity or appearance of a particular personalized instance. Our setting instead requires instance-level preservation of a specific personalized subject while simultaneously achieving a designated target prediction. To separate these objectives, we preserve the personalized subject and introduce a spatially separate non-subject carrier that provides additional adversarial degrees of freedom outside the subject.

We therefore regard ObjectAdv and dual-label guided attacks as related unrestricted diffusion attacks, but not as direct baselines for personalized-subject preservation.
\section{Human Evaluation}
\label{app:human_test}

We conduct a blinded human evaluation on all 5,400 carrier-based attacked images,
covering 600 source--target pairs, three carrier conditions, and three attack routes.
Each image is evaluated independently by three annotators. Annotators are not shown
the attack route, carrier condition, expected subject label, or model predictions.
To avoid priming annotators with the intended subject-carrier hierarchy, the interface uses a neutral category-selection question.

For each image, annotators answer: ``Most relevant option describing the primary subject in the image'' They select from four randomly ordered options: the personalized-subject category, the attack target, and two distractor categories. A response is counted as a subject-category selection when the chosen label matches the predefined personalized-subject category. The final image-level judgment is determined by majority vote.

According to Table~\ref{tab:human_study}, individual subject-selection rates range from 99.74\% to 99.93\%, which is quite a high number. These results show that image-level category judgments remain overwhelmingly aligned with the personalized-subject category. Together with the DINO similarity and SAM3 IoU results~\ref{tab:subject_preservation} and clean image prediction by BlackBox and WhiteBox classifier in~\ref{app:BlackBox_Results} and conditional ASR , this provides complementary evidence that the carrier does not displace the original subject as the dominant image content.

\begin{table}[htbp]
\centering
\caption{Human-evaluation interface. Annotators are shown an attacked image
and asked to select its primary category from four randomly ordered options:
the personalized-subject category, the attack target, and two distractor
categories. The reported selection rate is the proportion of responses
matching the personalized-subject category.}
\label{tab:human_study}
\small
\begin{tabular}{lcc}
\toprule
Evaluation & Subject-Category Selections & Rate (\%) \\
\midrule
Annotator 1 & 5,391 / 5,400 & 99.83 \\
Annotator 2 & 5,386 / 5,400 & 99.74 \\
Annotator 3 & 5,396 / 5,400 & 99.93 \\
Majority vote & 5,398 / 5,400 & 99.96 \\
\bottomrule
\end{tabular}
\end{table}

\begin{figure}[t]
    \centering
    \includegraphics[width=\linewidth]{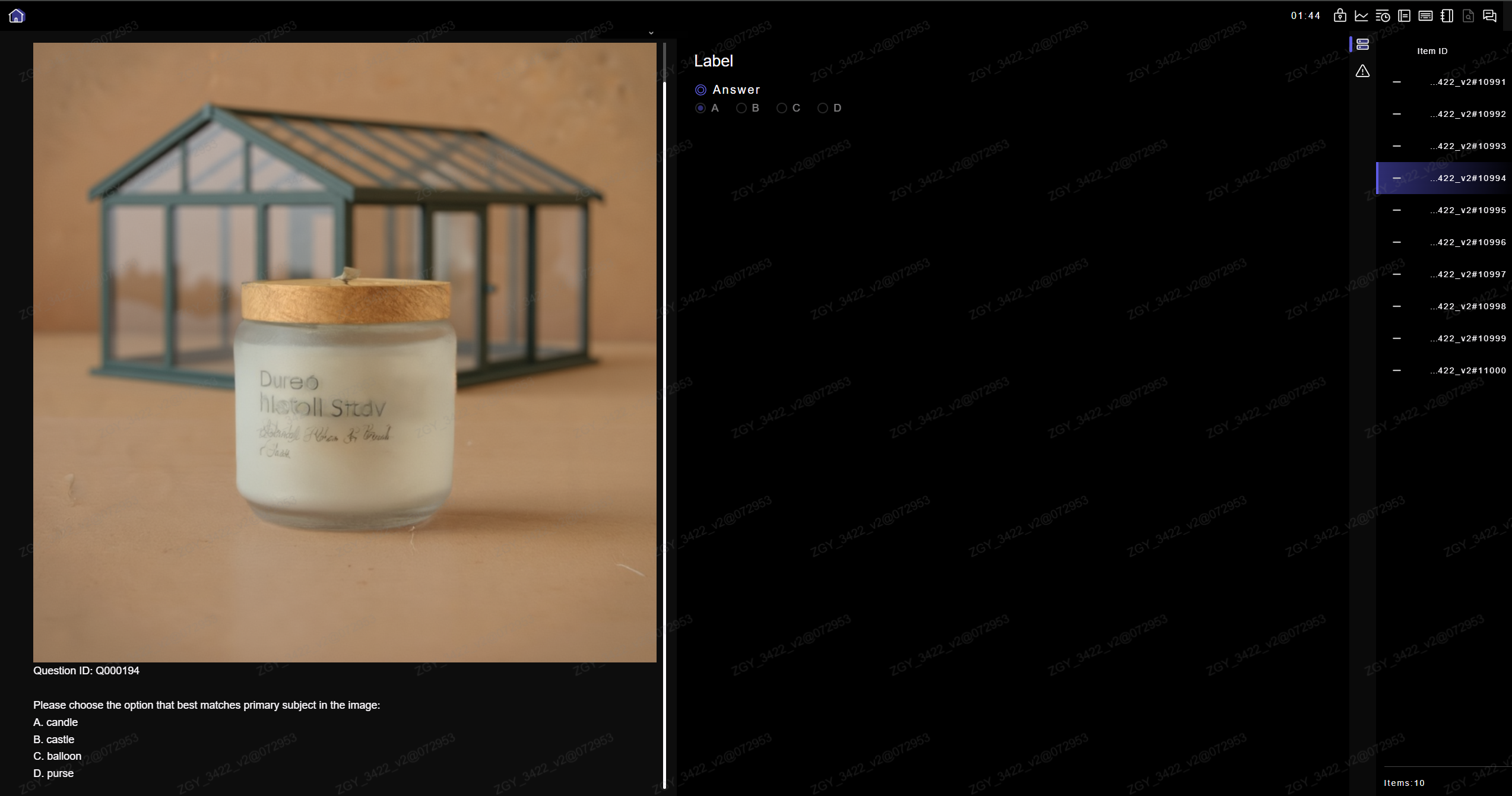}
    \caption{
    Screenshot of the human-evaluation interface. Annotators are shown an attacked
    image and asked to identify its primary subject from four randomly ordered
    candidate categories.
    }
    \label{fig:human_study_interface}
\end{figure}
\section{Limitations and Scope}
\label{Limitations and Scope}

Our current carrier realizations relies on visible object instances generated through direct compositing or mask-guided inpainting. However, the carrier construction is not always faithful only based on prompt generation by FLUX. Specifically, Hybrid attributes are not
consistently realized by CIRA and JIA, while CRA may partially occlude a valid carrier after subject compositing. Thus, we use the gate check discussed in Appendix~\ref{app:SAM3_gates} to ease the gap but when all candidates fail, the construction still retain the forced selection.

Our transferability analysis provides a conditional lower bound under the ideal-margin ordering and common classifier-discrepancy assumptions. It explains how a larger ideal target margin can yield a higher transferability guarantee, while the margin ordering and the corresponding increase in realized cross-model transferability remain empirically evaluated.

\end{document}